\documentclass[conference]{IEEEtran}
\usepackage{times}

\usepackage[numbers]{natbib}
\usepackage{multicol}
\usepackage{amsthm}
\usepackage[bookmarks=true]{hyperref}

\usepackage{comment}
\usepackage{siunitx}
\usepackage{relsize}
\usepackage{ifthen}
\usepackage[colorinlistoftodos]{todonotes}

\usepackage[caption=false]{subfig}

\usepackage[vlined,ruled,linesnumbered]{algorithm2e}
\usepackage{graphics} 
\usepackage{rotating}
\usepackage{color}
\usepackage{enumerate}
\usepackage[T1]{fontenc}
\usepackage{psfrag}
\usepackage{epsfig} 
\usepackage{booktabs}
\usepackage{graphicx,url}
\usepackage{multirow}
\usepackage{array}
\usepackage{latexsym}
\usepackage{amsfonts}
\usepackage{amsmath}
\usepackage{amssymb}
\usepackage{xstring}
\usepackage[noend]{algorithmic}
\usepackage{multirow}
\usepackage{xcolor}
\usepackage{prettyref}
\usepackage{flexisym}
\usepackage{bigdelim}
\usepackage{breqn} 
\usepackage{listings}
\let\labelindent\relax
\usepackage{enumitem}
\usepackage{xspace}
\usepackage{bm}
\graphicspath{{./figures/}}
\usepackage{tikz}
\usetikzlibrary{matrix,calc}

\usepackage{mdwlist}

\let\itemize\undefined
\makecompactlist{itemize}{stditemize}

\newrefformat{prob}{Problem\,\ref{#1}}
\newrefformat{def}{Definition\,\ref{#1}}
\newrefformat{sec}{Section\,\ref{#1}}
\newrefformat{sub}{Section\,\ref{#1}}
\newrefformat{prop}{Proposition\,\ref{#1}}
\newrefformat{app}{Appendix\,\ref{#1}}
\newrefformat{alg}{Algorithm\,\ref{#1}}
\newrefformat{cor}{Corollary\,\ref{#1}}
\newrefformat{thm}{Theorem\,\ref{#1}}
\newrefformat{lem}{Lemma\,\ref{#1}}
\newrefformat{fig}{Fig.\,\ref{#1}}
\newrefformat{tab}{Table\,\ref{#1}}

\newtheorem{theorem}{Theorem}

\newtheorem{lemma}[theorem]{Lemma}

\newtheorem{proposition}[theorem]{Proposition}
\newtheorem{remark}[theorem]{Remark}
\newtheorem{example}[theorem]{Example}

\newcommand{\cf}{\emph{cf.}\xspace}

\newcommand{\bdmath}{\begin{dmath}}
\newcommand{\edmath}{\end{dmath}}
\newcommand{\beq}{\begin{equation}}
\newcommand{\eeq}{\end{equation}}
\newcommand{\bdm}{\begin{displaymath}}
\newcommand{\edm}{\end{displaymath}}
\newcommand{\bea}{\begin{eqnarray}}
\newcommand{\eea}{\end{eqnarray}}
\newcommand{\beal}{\beq \begin{array}{ll}}
\newcommand{\eeal}{\end{array} \eeq}
\newcommand{\beas}{\begin{eqnarray*}}
\newcommand{\eeas}{\end{eqnarray*}}
\newcommand{\ba}{\begin{array}}
\newcommand{\ea}{\end{array}}
\newcommand{\bit}{\begin{itemize}}
\newcommand{\eit}{\end{itemize}}
\newcommand{\ben}{\begin{enumerate}}
\newcommand{\een}{\end{enumerate}}
\newcommand{\expl}[1]{&&\qquad\text{\color{gray}(#1)}\nonumber}

\newcommand{\calA}{{\cal A}}
\newcommand{\calB}{{\cal B}}
\newcommand{\calC}{{\cal C}}

\newcommand{\calE}{{\cal E}}

\newcommand{\calG}{{\cal G}}

\newcommand{\calN}{{\cal N}}

\newcommand{\calS}{{\cal S}}

\newcommand{\calV}{{\cal V}}

\newcommand{\setal}{~\emph{et~al.}\xspace}

\newcommand{\M}[1]{{\bm #1}} 
\renewcommand{\boldsymbol}[1]{{\bm #1}}

\newcommand{\hide}[1]{}

\newcommand{\hiddenText}{{\color{gray} hidden text.}}
\newcommand{\hideWithText}[1]{\hiddenText}

\newcommand{\kron}{\otimes}

\newcommand{\subject}{\text{ subject to }}
\DeclareMathOperator*{\argmax}{arg\,max}
\DeclareMathOperator*{\argmin}{arg\,min}

\newcommand{\normsq}[2]{\left\|#1\right\|^2_{#2}}

\newcommand{\tran}{^{\mathsf{T}}}

\newcommand{\trace}[1]{\mathrm{tr}\left(#1\right)}

\newcommand{\rank}[1]{\mathrm{rank}\left(#1\right)}

\newcommand{\inv}{^{-1}}

\newcommand{\ones}{{\mathbf 1}}
\newcommand{\zero}{{\mathbf 0}}
\newcommand{\eye}{{\mathbf I}}

\newcommand{\Real}[1]{ { {\mathbb R}^{#1} } }

\newcommand{\setdef}[2]{ \{#1 \; {:} \; #2 \} }

\newcommand{\SOthree}{\ensuremath{\mathrm{SO}(3)}\xspace}
\newcommand{\Othree}{\ensuremath{\mathrm{O}(3)}\xspace}

\newcommand{\MA}{\M{A}}

\newcommand{\MM}{\M{M}}

\newcommand{\MQ}{\M{Q}}

\newcommand{\MR}{\M{R}}

\newcommand{\MI}{\M{I}}

\newcommand{\MX}{\M{X}}

\newcommand{\MZ}{\M{Z}}

\newcommand{\va}{\boldsymbol{a}} 
 
\newcommand{\vb}{\boldsymbol{b}}

\newcommand{\ve}{\boldsymbol{e}}

\newcommand{\vo}{\boldsymbol{o}}

\newcommand{\vv}{\boldsymbol{v}}
\newcommand{\vt}{\boldsymbol{t}}
\newcommand{\vxx}{\boldsymbol{x}}

\newcommand{\vepsilon}{\boldsymbol{\epsilon}}

\newcommand{\scenario}[1]{{\smaller \sf#1}\xspace}

\newcommand{\cvx}{{\sf cvx}\xspace}

\newcommand{\blue}[1]{{\color{blue}#1}}

\newcommand{\linkToPdf}[1]{\href{#1}{\blue{(pdf)}}}
\newcommand{\linkToPpt}[1]{\href{#1}{\blue{(ppt)}}}
\newcommand{\linkToCode}[1]{\href{#1}{\blue{(code)}}}
\newcommand{\linkToWeb}[1]{\href{#1}{\blue{(web)}}}
\newcommand{\linkToVideo}[1]{\href{#1}{\blue{(video)}}}
\newcommand{\award}[1]{\xspace} 

\newcommand{\moderateOutliers}{{60\%}\xspace}
\newcommand{\maxOutliers}{{99\%}\xspace}
\newcommand{\maxOutliersRot}{{90\%}\xspace}

\renewcommand{\qed}{{\hfill $\square$}}

\newcommand{\tlsProblem}{TR\xspace}
\newcommand{\tlsProblemLong}{Truncated Least Squared Registration\xspace}
\newcommand{\TLS}{\scenario{TLS}}
\newcommand{\TIM}{\scenario{TIM}}
\newcommand{\TIMs}{\scenario{TIMs}}
\newcommand{\TRIM}{\scenario{TRIM}}
\newcommand{\TRIMs}{\scenario{TRIMs}}

\newcommand{\bnb}{BnB\xspace}

\newcommand{\SPC}{SPC\xspace}

\newcommand{\TIMa}{\bar{\va}}
\newcommand{\TIMb}{\bar{\vb}}

\newcommand{\hats}{\hat{s}}
\newcommand{\hatMR}{\hat{\MR}}
\newcommand{\hatvt}{\hat{\vt}}

\newcommand{\barMQ}{\bar{\MQ}}
\newcommand{\eps}{{\epsilon}}
\newcommand{\veps}{\M{\epsilon}}

\newcommand{\nrPoints}{N}
\newcommand{\nrTIM}{K}
\newcommand{\sumAllPointsi}{\sum_{i=1}^{N}}

\newcommand{\sumAllIM}{\sum_{k=1}^{K}}
\newcommand{\barc}{\bar{c}}
\newcommand{\barcsq}{\barc^2}

\newcommand{\myParagraph}[1]{{\bf #1.}}

\newcommand{\supp}{Supplementary Material\xspace}

\newcommand{\name}{\scenario{TEASER}}
\newcommand{\nameLong}{Truncated least squares Estimation And SEmidefinite Relaxation\xspace}

\newcommand{\nameCloning}{Binary cloning}
\newcommand{\mcis}{\scenario{MCIS}}

\newcommand{\FGR}{\scenario{FGR}}
\newcommand{\GORE}{\scenario{GORE}} 
\newcommand{\goICP}{\scenario{Go-ICP}} 
\newcommand{\ransac}{\scenario{RANSAC}}
\newcommand{\ransaconek}{\scenario{RANSAC (1K)}}
\newcommand{\ransactenk}{\scenario{RANSAC (10K)}}
\newcommand{\ICP}{\scenario{ICP}} 

\newcommand{\SIFT}{\scenario{SIFT}} 
\newcommand{\ORB}{\scenario{ORB}} 

\newcommand{\bunny}{\scenario{Bunny}} 
\newcommand{\armadillo}{\scenario{Armadillo}} 
\newcommand{\dragon}{\scenario{Dragon}}
\newcommand{\buddha}{\scenario{Buddha}}

\newcommand{\bmat}{\left[ \begin{array}}
\newcommand{\emat}{\end{array} \right]}

\newcommand{\barva}{\bar{\va}}
\newcommand{\barvb}{\bar{\vb}}

\newcommand{\ie}{\emph{i.e.}}

\newcommand{\edit}[1]{ #1 }

\newcommand\mydots{\makebox[0.7em][c]{\!.\,.}}

\renewcommand{\expl}[1]{\text{\small\color{gray}(#1)}\nonumber\\}

\begin{document}

\title{A Polynomial-time Solution for Robust \\  Registration with Extreme Outlier Rates}

\author{Heng Yang and Luca Carlone}

\author{\authorblockN{Heng Yang and Luca Carlone} 
\authorblockA{Laboratory for 
Information \& Decision Systems (LIDS)\\
Massachusetts Institute of Technology,
Cambridge, Massachusetts 02139\\
Email: \{hankyang, lcarlone\}@mit.edu}
}


%

\maketitle

\begin{tikzpicture}[overlay, remember picture]
\path (current page.north east) ++(-4.3,-0.2) node[below left] {
This paper has been accepted for publication in Robotics: Science and Systems, 2019.
};
\end{tikzpicture}
\begin{tikzpicture}[overlay, remember picture]
\path (current page.north east) ++(-7.0,-0.6) node[below left] {
Please cite the paper as: H. Yang and L. Carlone,
};
\end{tikzpicture}
\begin{tikzpicture}[overlay, remember picture]
\path (current page.north east) ++(-4.6,-1) node[below left] {
``A Polynomial-time Solution for Robust Registration with Extreme Outlier Rates'',
};
\end{tikzpicture}
\begin{tikzpicture}[overlay, remember picture]
\path (current page.north east) ++(-7.5,-1.4) node[below left] {
 Robotics: Science and Systems (RSS), 2019.
};
\end{tikzpicture}


\begin{abstract}
We propose a robust approach for the registration of two sets of 3D points in the presence of a large amount of outliers. 
%
%
Our first contribution is to reformulate the registration problem using a \emph{Truncated Least Squares} (\TLS) cost that makes the estimation insensitive to a large fraction of spurious point-to-point correspondences. 
The second contribution is a general framework to decouple rotation, translation, and scale estimation, which 
allows solving in cascade for the three transformations. 
%
Since each subproblem (scale, rotation, and translation estimation) is still non-convex and combinatorial in nature, 
out third contribution is to show that (i) \TLS scale and (component-wise) translation estimation can be solved exactly and in polynomial time via an \emph{adaptive voting} scheme, (ii) \TLS rotation estimation can be relaxed to a semidefinite program and the relaxation is tight in practice, even in \edit{the} presence of an extreme amount of outliers. 
%
We validate the proposed algorithm, named \name (\emph{\nameLong}),
 in standard registration benchmarks showing that the algorithm outperforms \ransac and robust local optimization techniques, and  
 favorably compares with Branch-and-Bound methods, while being a polynomial-time algorithm. \name can tolerate 
up to $\maxOutliers$ outliers and returns highly-accurate solutions.
\end{abstract}




\section{Introduction}
\label{sec:intro}

\emph{Point cloud registration}
is a fundamental problem in robotics and computer vision and consists in finding 
the best transformation (rotation, translation, and potentially scale) that aligns two point clouds.
It finds applications in motion estimation and 
%
%
3D reconstruction~\cite{Henry12ijrr-rgbdMapping,Blais95pami-registration,Choi15cvpr-robustReconstruction,Zhang15icra-vloam},
object recognition and localization~\cite{Drost10cvpr,Wong17iros-segicp,Zeng17icra-amazonChallenge,Marion18icra-labelFusion}, 
panorama stitching~\cite{Bazin14eccv-robustRelRot},  
and medical imaging~\cite{Audette00mia-surveyMedical,Tam13tvcg-registrationSurvey}, to name a few.

When the ground-truth correspondences between the point clouds are known and the noise follows a zero-mean Gaussian distribution, 
the registration problem can be readily solved, since elegant closed-form solutions~\cite{Horn87josa,Arun87pami} 
exist for the case of isotropic noise, 
and recently proposed convex relaxations~\cite{Briales17cvpr-registration} 
 are empirically tight even in \edit{the} presence of large anisotropic noise.
In practice, however, the correspondences are either unknown, or contain a high ratio of outliers. 
Large outlier rates are typical of 3D keypoint detection and matching techniques~\cite{Tombari13ijcv-3DkeypointEvaluation,Rusu08iros-3Dkeypoints}.  
Therefore, it is common to use the aforementioned methods within a \ransac scheme~\cite{Fischler81}.

While \ransac is \edit{a} popular approach for several robust vision and robotics problems, 
its runtime grows exponentially with the outlier ratio~\cite{Bustos18pami-GORE} and it
can perform poorly with extreme outlier rates.
 The capability of tolerating a large amount of outliers is of paramount importance in applications where 
 the correspondences are unknown and when operating in the 
 clutter (e.g., object pose estimation in the wild). Moreover, even when the correspondences are known but uncertain, 
 it is desirable to develop registration techniques that can afford stronger performance guarantees compared to \ransac.

This paper is motivated by the goal of designing an approach that 
(i) can solve registration globally (without relying on an initial guess), 
(ii) can tolerate extreme amounts of outliers 
(e.g., when $\maxOutliers$ of the measurements are outliers),  
(iii) runs in polynomial time, 
and (iv) provides formal performance guarantees. 
The related literature, reviewed in Section~\ref{sec:relatedWork}, 
fails to simultaneously address all these aspects, 
and only includes techniques that are robust to moderate  amounts (e.g., $\moderateOutliers$) of outliers 
and lack optimality guarantees~(e.g., \FGR~\cite{Zhou16eccv-fastGlobalRegistration}), 
or are globally optimal but run in exponential time in the worst case, such as branch-and-bound (\bnb) methods
 (e.g., \goICP~\cite{Yang16pami-goicp}). 


\newcommand{\myhspace}{\hspace{-3mm}}

\newcommand{\mpw}{4.5cm}
\begin{figure}[t]
	\begin{center}
	\begin{minipage}{\textwidth}
	\hspace{-0.2cm}
	\begin{tabular}{cc}%
		\myhspace
			\begin{minipage}{\mpw}%
			\centering%
			\includegraphics[width=1.0\columnwidth]{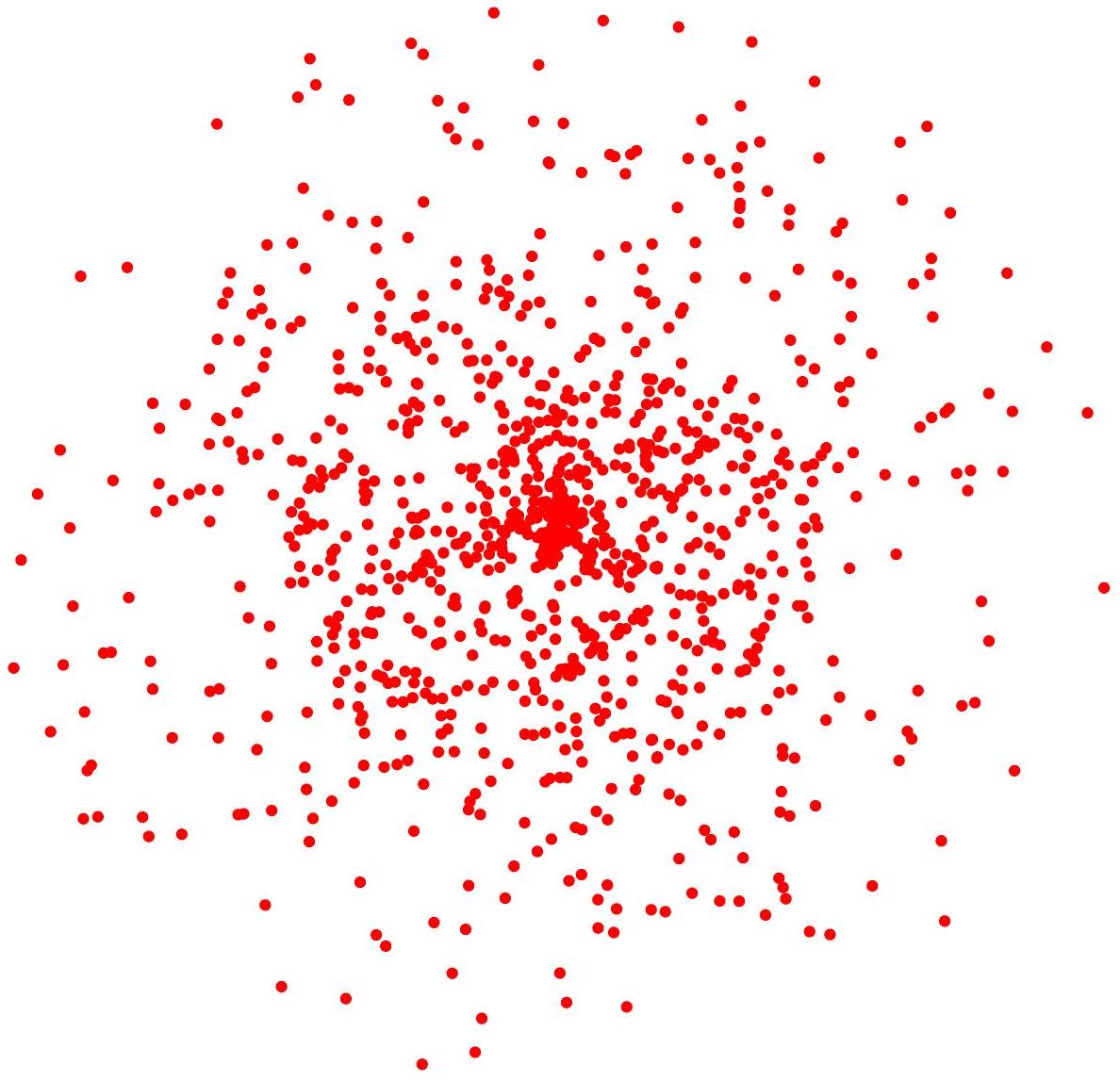} \vspace{-7mm}\\
			(a) Cluttered scene 
			\end{minipage}
		& \myhspace
			\begin{minipage}{\mpw}%
			\centering%
			\includegraphics[width=1.0\columnwidth]{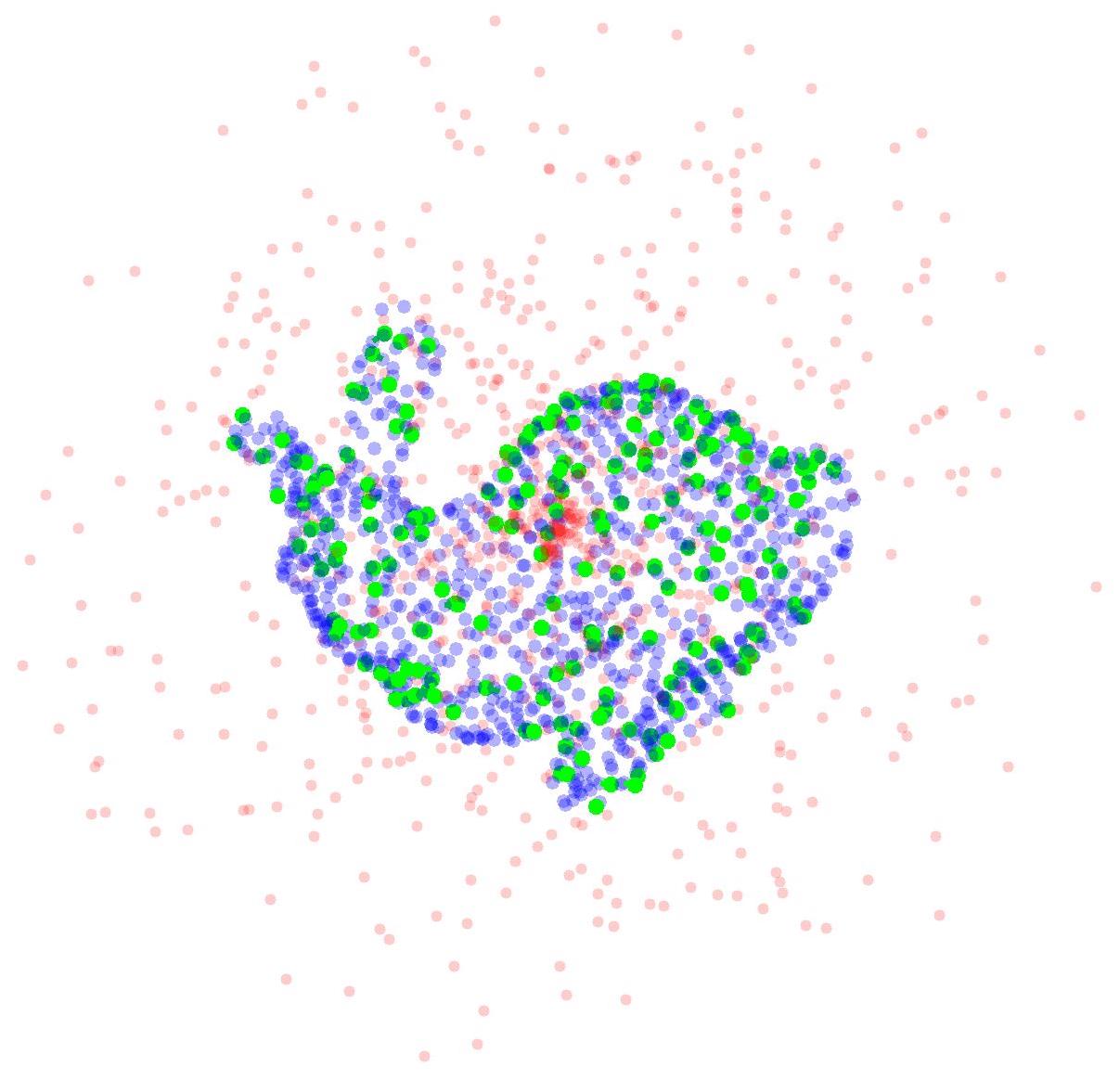} \vspace{-7mm}\\
			(b) Registration result
			\end{minipage}
		\end{tabular}
	\end{minipage}
	\begin{minipage}{\textwidth}
	\end{minipage}
	\vspace{-3mm} 
	\caption{We address 3D point cloud registration with extreme outlier rates. 
	(a) \bunny dataset spoiled with 80\% outlier correspondences: finding/localizing the bunny is challenging even for a human;
	(b) The proposed \emph{\nameLong} (\name) is able to find the correct inlier correspondences (green) and compute 
	the correct registration result in polynomial time.
	\name is robust to \maxOutliers outlier correspondences.
	 \label{fig:overview}}
	\vspace{-12mm} 
	\end{center}
\end{figure}

\myParagraph{Contribution}
%
Our first contribution (presented in \prettyref{sec:TLSregistration}) 
is to reformulate the registration problem using a \emph{Truncated Least Squares} (\TLS) cost that is insensitive to a large fraction of spurious data. We name the resulting problem the \emph{\tlsProblemLong} (\tlsProblem) problem.

The second contribution (\prettyref{sec:decoupling}) 
 is a general framework to decouple scale, rotation, and translation estimation. 
 The idea of decoupling rotation and translation has appeared in related work, \edit{e.g.,~\cite{Makadia06cvpr-registration,Liu18eccv-registration,Bustos18pami-GORE}.}
 The novelty of our proposal is threefold: 
 (i) we develop invariant measurements to estimate the scale (\cite{Liu18eccv-registration,Bustos18pami-GORE} 
 assume the scale is given), 
 (ii) we make the decoupling formal within the framework of 
 \emph{unknown-but-bounded} noise~\cite{Milanese89chapter-ubb}, and 
 (iii) we provide a general graph-theoretic framework to derive these invariant measurements.
 
The decoupling allows solving in cascade for scale, rotation, and translation.
However, each subproblem is still combinatorial in nature. 
Our third contribution is to show that (i) in the scalar case \TLS estimation 
can be solved exactly in polynomial time 
using an \emph{adaptive voting} scheme, and this enables efficient estimation of 
the scale and the (component-wise) translation; 
(ii) we can prune a large amount of outliers by finding a \emph{maximal clique} of the graph defined by the
invariant measurements; 
(iii) we  can formulate a tight \emph{semidefinite programming (SDP) relaxation} to estimate the rotation, 
(iv) we can provide per-instance bounds on the performance of the SDP relaxation.
 To the best of our knowledge, this is the first polynomial-time algorithm for outlier-robust registration 
 with computable performance guarantees.

We validate the proposed algorithm, named \emph{\nameLong} (\name),
in standard registration benchmarks as well as robotics datasets, 
showing that the algorithm outperforms \ransac and robust local optimization techniques, and  
 favorably compares with Branch-and-Bound methods, while being a polynomial-time algorithm. \name can tolerate 
up to $\maxOutliers$ outliers (Fig.~\ref{fig:overview})  and returns highly-accurate solutions.


\section{Related Work}
\label{sec:relatedWork}
There are two established paradigms for the registration of 3D point clouds: 
\emph{Correspondence-based} and 
\emph{Simultaneous Pose and Correspondence} methods. 

\subsection{Correspondence-based Methods}
Correspondence-based methods first detect and match 3D keypoints between point clouds using local~\cite{Tombari13ijcv-3DkeypointEvaluation,Guo14pami-3Dkeypoints,Rusu08iros-3Dkeypoints,Rusu09icra-fast3Dkeypoints} or global~\cite{Drost10cvpr,Koppula11nips} descriptors to establish putative correspondences, and then either use closed-form solutions~\cite{Horn87josa,Arun87pami} in a \ransac~\cite{Fischler81} scheme, or apply robust optimization methods~\cite{Zhou16eccv-fastGlobalRegistration,Bustos18pami-GORE} to gain robustness against outliers. 3D keypoint matching is 
known to be less accurate compared to 2D counterparts like \SIFT and \ORB,
thus causing much higher outlier rates, e.g., having 95\% spurious correspondences is considered common~\cite{Bustos18pami-GORE}. Therefore, a robust backend that can deal with extreme outlier rates is highly desirable.

\myParagraph{Registration without outliers} Horn~\cite{Horn87josa} and Arun~\cite{Arun87pami} show that optimal solutions 
(in the maximum likelihood sense) for scale, rotation, and translation can be computed in closed form 
when the points are affected by isotropic zero-mean Gaussian noise.
 Olsson\setal~\cite{olsson2009pami-branch} propose a method based on \bnb that is globally optimal and allows point-to-point, point-to-line, and point-to-plane correspondences. Recently, Briales and Gonzalez-Jimenez~\cite{Briales17cvpr-registration} propose a semidefinite relaxation that can deal with anisotropic Gaussian noise, and has per-instance optimality guarantees. 

\myParagraph{Robust registration} Probably the most widely used robust registration approach is based on \ransac~\cite{Fischler81,Chen99pami-ransac}, which has enabled several early applications in vision and robotics~\cite{Hartley00,Meer91ijcv-robustVision}. Despite its efficiency in the low-noise and low-outlier regime, \ransac exhibits slow convergence and low accuracy 
 with large outlier rates~\cite{Bustos18pami-GORE}, where it becomes harder to sample a ``good'' consensus set.
 Other approaches resort to \emph{M-estimation}, which replaces the least squares objective function with robust costs that are less sensitive to outliers~\cite{MacTavish15crv-robustEstimation,Black96ijcv-unification,Lajoie19ral-DCGM}. Zhou\setal~\cite{Zhou16eccv-fastGlobalRegistration} propose \emph{Fast Global Registration} (\FGR) that uses the Geman-McClure cost function and leverages 
 graduated non-convexity to solve the resulting non-convex optimization.
 Since graduated non-convexity has to be solved in discrete steps, \edit{\FGR} does not guarantee global optimality~\cite{Bustos18pami-GORE}. Indeed, \FGR tends to fail when the outlier ratio is high (>80\%), as we show in~\prettyref{sec:experiments}.
 Bustos and Chin~\cite{Bustos18pami-GORE} propose a \emph{Guaranteed Outlier REmoval} (\GORE) technique, that uses 
 geometric operations to significantly reduce the amount of outlier correspondences before passing them to the optimization backend. \GORE has been shown to be robust to 95\% spurious correspondences~\cite{Bustos18pami-GORE}. 
 However, \GORE does not estimate the \emph{scale} of the registration and has exponential worst-case time complexity due to the possible usage of \bnb (see Algorithm 2 in~\cite{Bustos18pami-GORE}).

\subsection{Simultaneous Pose and Correspondence Methods}
\emph{Simultaneous Pose and Correspondence} (\SPC) methods 
 alternate between finding the correspondences and computing the best transformation given the correspondences.

\myParagraph{Local methods} The \emph{Iterative Closest Point} (\ICP) algorithm~\cite{Besl92pami} is considered a milestone in point cloud registration. 
However, \ICP is \edit{prone to converge}
 to \emph{local minima} and it only performs well given a good initial guess. 
Multiple variants of \ICP~\cite{Granger02eccv,Sandhu10pami-PFregistration,Zhang16sp-robustICP,
Maier12pami-convergentICP,Chetverikov05ivc-trimmedICP,Kaneko03pr-robustICP,
Bouaziz13acmsig-sparseICP} \edit{have proposed to use}
 robust cost functions to improve convergence.
Probabilistic interpretations have also been proposed to improve \ICP convergence, for instance interpreting the registration problem as a minimization of the Kullback-Leibler divergence between two Gaussian Mixture Models~\cite{Jian05iccv-registrationGMM,Myronenko10pami-coherentPointDrift,Jian11pami-registrationGMM,Campbell15iccv-adaptiveRegistration}. 
All these methods rely on iterative local search, do not provide global optimality guarantees, and typically fail 
without a good initial guess.

\myParagraph{Global methods} Global \SPC approaches compute a globally optimal solution without initial guesses,
and are usually based on \bnb, which at each iteration divides the parameter space into multiple sub-domains (branch) and computes the bounds of the objective function for each sub-domain (bound). A series of geometric techniques have been proposed to improve the bounding tightness~\cite{Hartley09ijcv-globalRotationRegistration,
Breuel03cviu-BnBimplementation,Yang16pami-goicp,
Campbell16cvpr-gogma,Parra14cvpr-fastRotationRegistration} and increase the search speed~\cite{Yang16pami-goicp,Li19arxiv-fastRegistration}. However, the runtime of \bnb increases exponentially with the size of the point cloud and it can be made worse by the explosion of the number of local minima resulting from high outlier ratios~\cite{Bustos18pami-GORE}. \edit{Global SPC registration can  be also formulated as a mixed-integer program~\cite{Izatt17isrr-MIPregistration}, though the runtime remains exponential.}

\section{Robust Registration with \\ Truncated Least Squares Cost}
\label{sec:TLSregistration}

In the robust registration problem, we are given two 3D point sets 
$\calA = \{\va_i\}_{i=1}^N$ 
and 
$\calB=\{\vb_i\}_{i=1}^N$, 
with $\va_i, \vb_i \in \Real{3}$, 
such that:
 \bea
 \label{eq:robustGenModel}
 \vb_i = s \MR \va_i + \vt + \vo_i + \veps_i
 \eea
 where \edit{$s > 0$,}
$\MR \!\in\! \SOthree$, and $\vt \!\in\!\Real{3}$ are an unknown scale, rotation, and translation,
 $\veps_i$ models measurement noise, and $\vo_i$ is a vector of zeros for \emph{inliers}, or a vector of arbitrary numbers for \emph{outliers}. 
 In words, if the $i$-th correspondence $(\va_i,\vb_i)$ is an inlier, $\vb_i$ corresponds to a 3D transformation of $\va_i$ (plus noise), while if 
 $(\va_i,\vb_i)$ is an outlier correspondence,  $\vb_i$ is just an arbitrary vector.
 $\SOthree \doteq \setdef{\MR\in\Real{3 \times 3}}{\MR\tran \MR = \eye_3, \det(\MR)=+1}$ is the set of proper rotation matrices (where $\eye_d$ is the identity matrix of size $d$).
 We consider a correspondence-based setup, where we need to 
 compute $(s,\MR,\vt)$ given putative correspondences $(\va_i, \vb_i), i=1,\ldots,\nrPoints$. 

 \myParagraph{Registration without outliers} 
 When $\veps_i$ is a zero-mean Gaussian noise with isotropic covariance $\sigma_i^2 \eye_3$, and all the correspondences are correct (i.e., $\vo_i=\bm{0},\forall i$), the Maximum Likelihood estimator of $(s,\MR,\vt)$ can be computed 
 in closed form by decoupling the estimation of the scale, translation, and rotation, using Horn's~\cite{Horn87josa} or Arun's method~\cite{Arun87pami}.

\myParagraph{Robust registration} In practice, a large fraction of the correspondences are \emph{outliers}, due to 
incorrect keypoint matching.
Despite the elegance of the closed-form solutions~\cite{Horn87josa,Arun87pami}, they 
are not robust to outliers, and a single ``bad'' outlier can 
 compromise the correctness of the resulting estimate. 
 Hence, we propose a \emph{truncated least squares registration} formulation that can tolerate extreme amounts of spurious data.

\myParagraph{Truncated Least Squares Registration}  
We depart from the Gaussian noise model and assume the noise is \emph{unknown but bounded}~\cite{Milanese89chapter-ubb}.
Formally, we assume
the noise $\veps_i$ in~\eqref{eq:robustGenModel} is such that $\| \veps_i \| \leq \beta_i$, where 
$\beta_i$ is a given bound. 

Then we adopt the following \edit{\emph{Truncated Least Squares Registration (TR)} formulation}:
%
 \bea
 \label{eq:TLSRegistration}
 \min_{ \substack{ s > 0, \vt \in \Real{3}, \MR \in \SOthree} } 
 \sumAllPointsi \min \left( \frac{1}{\beta_i^2} \normsq{  \vb_i - s \MR \va_i - \vt }{} \; , \; \barcsq \right) 
 \eea
 which computes a least squares solution of measurements with small residual ($\frac{1}{\beta_i^2} \normsq{  \vb_i - s \MR \va_i - \vt }{} \leq \barcsq$), while discarding measurements with large residuals (when $\frac{1}{\beta_i^2} \normsq{  \vb_i - s \MR \va_i - \vt }{} > \barcsq$ the $i$-th summand becomes constant and does not influence the optimization). 
 The constant $\barcsq$ is typically chosen to be $1$, 
  while one may use a different $\barcsq$ 
 to be stricter or more lenient towards potential outliers.

\section{Decoupling Scale, Rotation, \\ and Translation Estimation}
\label{sec:decoupling}

We propose a polynomial-time algorithm that 
decouples the estimation of scale, translation, and rotation in problem~\eqref{eq:TLSRegistration}. The key insight is that we can reformulate 
the measurements~\eqref{eq:robustGenModel} to obtain quantities that are invariant to a subset of the 
transformations (scaling, rotation, translation). 

\subsection{Translation Invariant Measurements (\TIM)}

While the absolute positions of the points in $\calB$ depend on the translation $\vt$, 
the relative positions are invariant to $\vt$. 
Mathematically, given two points $\vb_i$ and $\vb_j$ from~\eqref{eq:robustGenModel},
the relative position of these two points is:  
\bea
 \vb_j - \vb_i = s \MR (\va_j - \va_i) + (\vo_j - \vo_i) + (\veps_j - \veps_i)
 \eea
 where the translation $\vt$ cancels out in the subtraction.
 Therefore, we can obtain a \emph{Translation Invariant Measurement} (\TIM) by computing $\TIMa_{ij} \doteq \va_j - \va_i$
 and  $\TIMb_{ij} \doteq \vb_j - \vb_i$, and the \TIM satisfies the following generative model:
 \beq
 \label{eq:TIM}
 \tag{TIM}
 \TIMb_{ij} = s \MR \TIMa_{ij} + \vo_{ij} + \veps_{ij} 
 \eeq
 where $\vo_{ij} \doteq \vo_j - \vo_i$ is zero if \emph{both} the $i$-th \emph{and} the $j$-th measurements are inliers (or it is 
 an arbitrary vector otherwise), while $\veps_{ij} \doteq \veps_j - \veps_i$ is the measurement noise. 
 It is easy to see that if $\|\veps_i\| \leq \beta_i$ and $\|\veps_j\| \leq \beta_j$ 
 then $\|\veps_{ij}\| \leq \beta_i + \beta_j \doteq \beta_{ij}$.

 The advantage of the \TIMs in eq.~\eqref{eq:TIM} is that their generative model \edit{only depends on} two unknowns, $s$ and $\MR$. 
 \edit{The number of \TIMs is upper-bounded by $\left( \substack{\nrPoints \\ 2} \right) = \nrPoints (\nrPoints-1) / 2$, where 
 pairwise relative measurements between all pairs of points are computed. For computational reasons, one might want to downsample the \TIMs.
 \prettyref{thm:TIM} below provides a graph-theoretic way to create the \TIMs.}


 \begin{theorem}[Translation Invariant Measurements]\label{thm:TIM}
 Define the vectors $\va \in \Real{3\nrPoints}$ (resp. $\vb \in \Real{3\nrPoints}$), obtained by concatenating all vectors $\va_i$ (resp. $\vb_i$) in a single column vector.
 Moreover, define an arbitrary graph $\calG$ with nodes $\{1,\ldots,\nrPoints\}$ and an arbitrary set of edges $\calE$. 
 Then, the vectors $\TIMa = (\MA \kron \eye_3) \va$ and $\TIMb = (\MA \kron \eye_3) \vb$ are \TIMs, where $\MA \in \Real{\vert \calE \vert \times \nrPoints}$ is the incidence matrix of $\calG$, and $\kron$ is the Kronecker product. 
 \end{theorem}

\begin{figure}[t]
	\begin{center}
			\includegraphics[width=0.7\columnwidth]{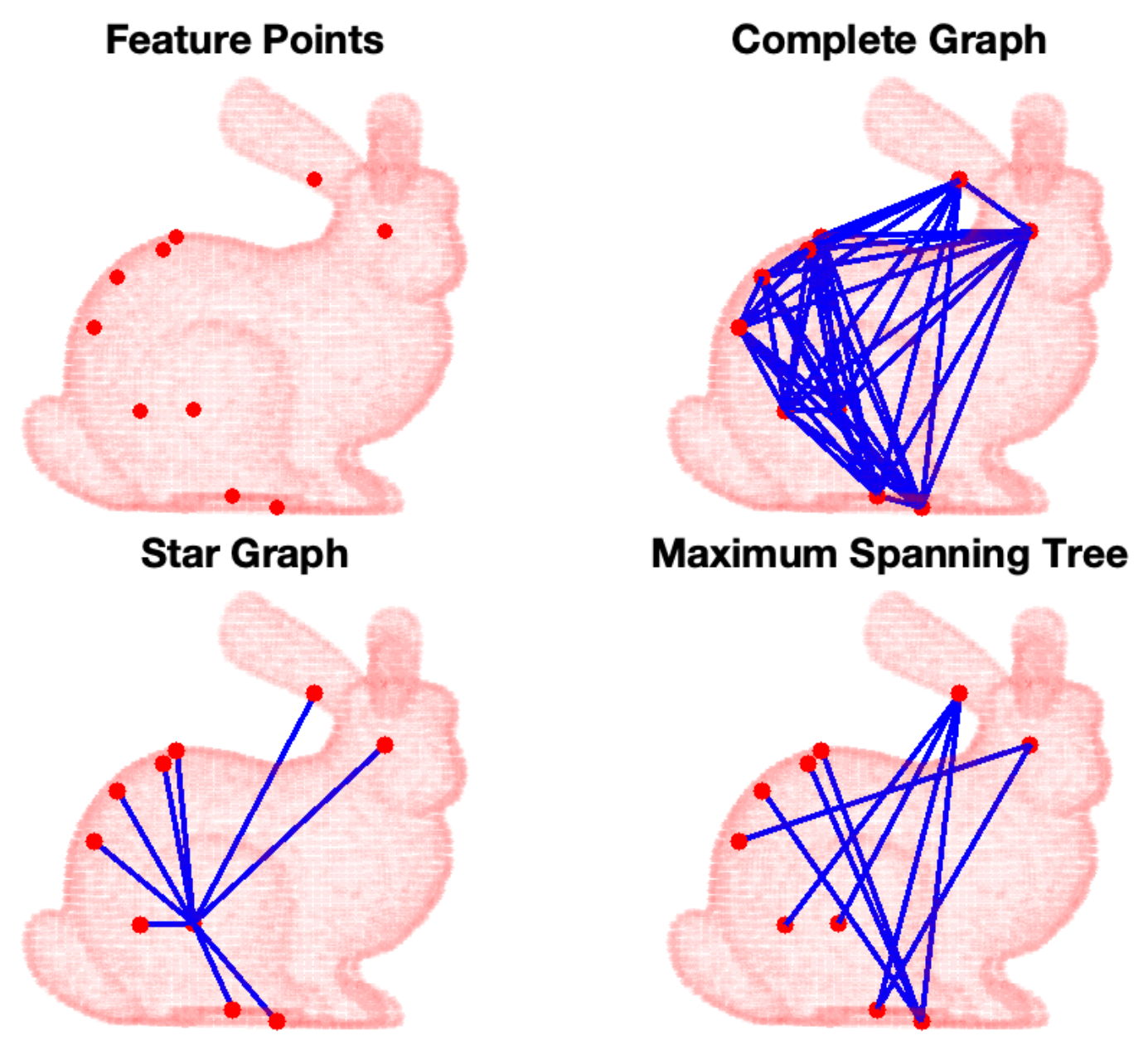}
	\caption{Graph topologies for generating \TIMs in the \bunny dataset.
	 \label{fig:TIMGraph}}
	 	\vspace{-8mm} 
	\end{center}
\end{figure}

A proof of the theorem is given in the \supp. \edit{Three potential graph topologies for generating \TIMs are illustrated in Fig.~\ref{fig:TIMGraph}.}



 \subsection{Translation and Rotation Invariant Measurements (\TRIM)}

While the relative locations of pairs of points (\TIMs) still depends on the rotation $\MR$, their distances 
are invariant to both $\MR$ and $\vt$. Therefore, to build 
rotation invariant measurements, we compute  
the norm of each \TIM vector:
\bea
\label{eq:TRIM01}
 \| \TIMb_{ij} \|  = \| s \MR \TIMa_{ij} + \vo_{ij} + \veps_{ij} \|
 \eea
 We now note that for the inliers ($\vo_{ij} = \bm{0}$) it holds (using $\| \veps_{ij} \| \leq \beta_{ij}$ and the triangle inequality):
 \bea
  \| s \MR \TIMa_{ij} \| - \beta_{ij}
 \leq 
 \| s \MR \TIMa_{ij} + \veps_{ij} \| 
 \leq 
 \| s \MR \TIMa_{ij} \| + \beta_{ij}
 \eea
 hence we can write~\eqref{eq:TRIM01} equivalently as:
 \bea
\label{eq:TRIM02}
 \| \TIMb_{ij} \|  = \| s \MR \TIMa_{ij} \| + \tilde{o}_{ij} + \tilde{\eps}_{ij}
 \eea
 with $|\tilde{\eps}_{ij}| \leq \beta_{ij}$, and $\tilde{o}_{ij} = 0$ if both $i$ and $j$ are inliers 
 or is an arbitrary scalar otherwise. Recalling that the norm is rotation invariant and that $s>0$, and dividing both members of~\eqref{eq:TRIM02}
 by $\|\TIMa_{ij}\|$, we obtain new measurements $s_{ij} \doteq \frac{\| \TIMb_{ij} \|}{\| \TIMa_{ij} \|}$:
 \beq
 \label{eq:TRIM}
 \tag{TRIM}
 s_{ij} = s + o^s_{ij} + \eps^s_{ij} 
 \eeq
 where $\eps^s_{ij} \doteq \frac{\tilde{\eps}_{ij}}{ \| \TIMa_{ij} \| }$, 
 and $o^s_{ij} \doteq \frac{\tilde{o}_{ij}}{ \| \TIMa_{ij} \| }$. 
 It is easy to see that  $|\eps^s_{ij} | \leq \beta_{ij} / \| \TIMa_{ij} \|$ since $|\tilde{\eps}_{ij}| \leq \beta_{ij}$. 
 We define $\alpha_{ij} \doteq \beta_{ij} / \| \TIMa_{ij} \|$.

 Eq.~\eqref{eq:TRIM} describes a \emph{translation and rotation invariant measurement} (\TRIM) whose generative model is only function of the unknown scale $s$. \edit{A summary table of the invariant measurements and a remark on the novelty of creating \TIMs and \TRIMs is presented in the \supp}. 

 \subsection{Our Registration Algorithm: \nameLong (\name)}

 We propose a decoupled approach 
 to solve in cascade for the scale, the rotation, and the translation in~\eqref{eq:TLSRegistration}.
 The approach, named \emph{\nameLong} (\name),
  works as follows:
 \begin{enumerate}
 \item we use the \TRIMs to estimate the scale $\hats$
 \item we use $\hats$ and the \TIMs to estimate the rotation $\hat\MR$
 \item we use $\hats$ and $\hat\MR$ to estimate the translation $\hat\vt$ from the original 
 \TLS problem~\eqref{eq:TLSRegistration}.
\end{enumerate}
The pseudocode is also summarized in Algorithm~\ref{alg:ITR}.

 \begin{algorithm}[h]
\SetAlgoLined
\textbf{Input:} \ points $(\va_i,\vb_i)$ and bounds $\beta_i$ ($i=1,\ldots,\nrPoints$), threshold $\barcsq$ (default: $\barcsq=1$), graph edges $\calE$ (default: $\calE$ describes the complete graph)\;  
\textbf{Output:} \  $s, \MR, \vt$\;
\% Compute \TIM and \TRIM \\
$\TIMb_{ij} = \vb_j-\vb_i \;,\; \TIMa_{ij} = \va_j-\va_i \;,\; \beta_{ij} = \beta_i + \beta_j \;\; \forall (i,j) \in \calE$ \\
$s_{ij} = \frac{ \| \TIMb_{ij} \| }{ \| \TIMa_{ij} \| }\;,\; \alpha_{ij} = \frac{\beta_{ij}}{\| \TIMa_{ij} \|}  \;\; \forall (i,j) \in \calE$ \\
\% Decoupled estimation of $s, \MR, \vt$\\
$\hats = {\tt estimate\_s}( \{ s_{ij},\alpha_{ij} : \forall (i,j) \in \calE\}, \barcsq )$\label{line:est_s}\\
$\hatMR = {\tt estimate\_R}( \{ \TIMa_{ij},\TIMb_{ij},\beta_{ij} : \forall (i,j) \in \calE\}, \barcsq, \hats )$\\
$\hatvt = {\tt estimate\_t}( \{ \va_{i},\vb_{i},\beta_{i} : i =1\ldots,\nrPoints \}, \barcsq, \hats, \hatMR )$\label{line:est_t}\\
 \textbf{return:} $\hats, \hatMR, \hatvt$
 \caption{\emph{\nameLong} (\name).\label{alg:ITR}}
\end{algorithm}

The following section describes how to implement the functions in 
lines~\ref{line:est_s}-\ref{line:est_t} of Algorithm~\ref{alg:ITR}.  In particular, we show how to
obtain global and robust estimates of scale (${\tt estimate\_s}$) in Section~\ref{sec:scaleEstimation},
rotation (${\tt estimate\_R}$) in Section~\ref{sec:rotationEstimation}, and translation (${\tt estimate\_t}$) in Section~\ref{sec:translationEstimation}. 


\section{Solving the Registration Subproblems}
\label{sec:subproblems}


\subsection{Robust Scale Estimation}
\label{sec:scaleEstimation}

The generative model~\eqref{eq:TRIM} describes linear measurements $s_{ij}$ of the unknown scale $s$, affected 
by bounded noise $|\eps^s_{ij}| \leq \alpha_{ij}$ including potential outliers (when $o^s_{ij} \neq 0$).
Again, we estimate the scale given the measurements $s_{ij}$ and the bounds $\alpha_{ij}$ using a 
\TLS estimator:

\vspace{-5mm}
\bea
\label{eq:TLSscale}
\hats = \argmin_{s} \sumAllIM \min\left( \frac{  (  s-s_k )^2 }{  \alpha^2_k }  \;,\; \barcsq \right)
\eea
where for simplicity we numbered the measurements from $1$ to $K = |\calE|$ and adopted the notation $s_k$ instead of $s_{ij}$.

The following theorem shows that one can solve~\eqref{eq:TLSscale} in polynomial time by a simple enumeration. 

\begin{theorem}[Optimal \TLS Scale Estimation]\label{thm:scalarTLS}
For a given $s \in \Real{}$, define the \emph{consensus set of $s$} as $\calC(s) = \setdef{k}{ \frac{  (  s-s_k )^2 }{  \alpha^2_k } \leq \barcsq }$. 
Then, for any $s \in \Real{}$, there are at most $2\nrTIM-1$ different non-empty consensus sets. If we name these sets $\calC_1, \ldots, \calC_{2K-1}$, then the solution of~\eqref{eq:TLSscale} can be computed by enumeration as:
\bea
\hats = \argmin 
\left\{ 
f_s(\hats_i) : \hats_i = 
\left(\sum_{k\in\calC_i} \frac{1}{\alpha_k^2} \right) \inv \sum_{k\in\calC_i} \frac{s_k}{\alpha_k^2}, \forall i 
\right\}
\eea   
where $f_s(\cdot)$ is the objective function of~\eqref{eq:TLSscale}. 
\end{theorem}

\begin{figure}[t]
	\begin{center}
	\begin{minipage}{\columnwidth}
	\includegraphics[width=1.0\columnwidth]{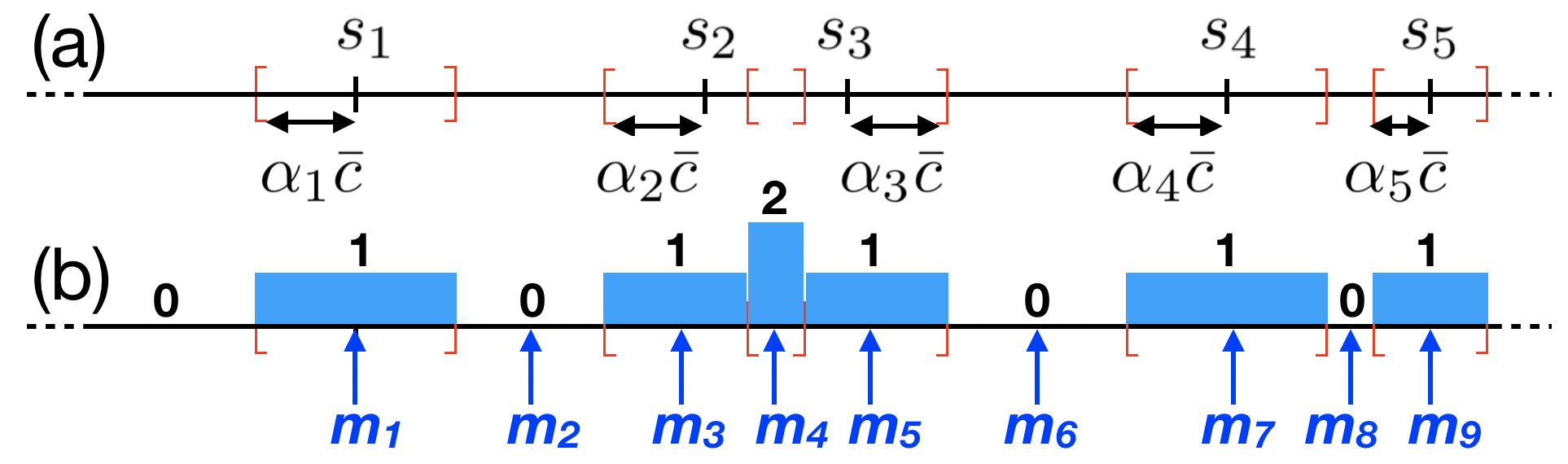}
	\end{minipage}
	\vspace{-3mm} 
	\caption{(a) confidence interval for each measurement $s_k$ (every $s$ in the $k$-th interval satisfies 
	$\frac{\|s-s_k\|^2}{\alpha_k^2} \leq \barcsq$;
	(b) \edit{cardinality} of the consensus set for every $s$ and middle-points $m_j$ for each interval with constant consensus set. 
	 \label{fig:consensusMax}}
	\vspace{-5mm} 
	\end{center}
\end{figure}

\prettyref{thm:scalarTLS}, whose proof is given in the \supp, is based on the insight 
that the consensus set can only change at the boundaries of the intervals $[s_k - \alpha_k\barc, s_k + \alpha_k\barc]$ (Fig.~\ref{fig:consensusMax}(a)) and there are at most $2\nrTIM$ such boundaries. The theorem
also suggests a straightforward \emph{adaptive voting} algorithm to solve~\eqref{eq:TLSscale}, with pseudocode given in Algorithm~\ref{alg:adaptiveVoting}. The algorithm first builds the boundaries of the 
intervals shown in Fig.~\ref{fig:consensusMax}(a) (line~\ref{line:boundaries}). Then, for each interval, it evaluates the consensus set (line~\ref{line:consensusSet}, see also Fig.~\ref{fig:consensusMax}(b)). 
Since the consensus set does not change within an interval, we compute it at the interval centers (line~\ref{line:midPoints}, see also Fig.~\ref{fig:consensusMax}(b)). Finally, the cost of each consensus set is computed and the smallest cost is returned as optimal solution (line~\ref{line:enumeration}).  

\begin{algorithm}[t]
\SetAlgoLined
\textbf{Input:} \ $s_k, \alpha_k, \barc$\; 
\textbf{Output:} \  $\hats$, scale estimate solving~\eqref{eq:TLSscale}\;
\% Define boundaries and sort \\
$\vv = \text{sort}([s_1 - \alpha_1\barc, s_1 + \alpha_1\barc,\ldots,s_\nrTIM - \alpha_\nrTIM\barc, s_\nrTIM + \alpha_\nrTIM\barc])$ \label{line:boundaries} \\
\% Compute middle points \\
$m_i = \frac{\vv_{2i-1} + \vv_{2i}}{2}$ for $i=1,\ldots,2\nrTIM-1$ \label{line:midPoints} \\
 \% Voting \\
 \For{$i = 1,\ldots,2\nrTIM-1$}{
 	$\calS_i = \emptyset$\\
 	\For{$k = 1,\ldots,\nrTIM$}{
		\If{$m_i \in [s_k - \alpha_k\barc, s_k + \alpha_k\barc]$}{
			$\calS_i = \calS_i \cup \{k\}$ \% add to consensus set \label{line:consensusSet} \\
		}
	}
}
\% Enumerate consensus sets and return best\\
 \textbf{return:} 
 $\!\argmin 
\left\{ 
f_s(\hats_i) : \hats_i = 
\left(\displaystyle\sum_{k\in\calC_i} \frac{1}{\alpha_k^2} \right)\inv \hspace{-3mm} \displaystyle\sum_{k\in\calC_i} \frac{s_k}{\alpha_k^2}, \forall i 
\right\}\!\!$  \label{line:enumeration}
 \caption{Adaptive Voting.\label{alg:adaptiveVoting}}
\end{algorithm}


The interested reader can find a discussion on the relation between \TLS and \emph{consensus maximization} (a popular approach for outlier detection~\cite{Speciale17cvpr-consensusMaximization,Liu18eccv-registration}) in the \supp.

\myParagraph{Maximal clique inlier selection (\mcis)} 
The graph theoretic interpretation of~\prettyref{thm:TIM} offers further opportunities to prune outliers.
Considering the \TRIMs as edges in the graph $\calG(\calV,\calE)$ (where the vertices $\calV$ are the 3D points and the 
edge set $\calE$ induces the \TIMs and \TRIMs per~\prettyref{thm:TIM}), 
we can use the scale estimate $\hats$ from Algorithm~\ref{alg:adaptiveVoting} to prune edges $(i,j)$ in the graph 
whose associated \TRIM $s_{ij}$ is such that $\frac{(s_{ij} - \hats)^2}{\alpha_{ij}^2} \!\!> \!\!\barcsq$.
This allows us 
to obtain a pruned graph $\calG'(\calV,\calE')$, with $\calE'\subseteq \calE$, where gross outliers are discarded.
The following result ensures that inliers form a clique in the graph $\calG'(\calV,\calE')$ enabling an even more substantial rejection of outliers. 

\begin{theorem}[Maximal Clique Inlier Selection]\label{thm:maxClique}
Edges corresponding to inlier \TIMs form a clique in $\calE'$, and there is at least one maximal clique in 
$\calE'$ that contains all the inliers. 
\end{theorem} 

A proof of Theorem~\ref{thm:maxClique} is presented in the \supp. Theorem~\ref{thm:maxClique} allows us to \edit{prune outliers} by finding the maximal cliques of $\calG'(\calV,\calE')$. Although finding the maximal cliques of a graph takes exponential time in general, there exist efficient approximation algorithms \edit{based on heuristics~\cite{Bron73acm-allCliques,Pattabiraman15im-maxClique,wu2015ejor-reviewMCPAlgs}}. In addition, under high outlier rates, the graph $\calG'(\calV,\calE')$ is sparse and the maximal clique problem can be solved quickly in practice~\cite{Eppstein10isac-maxCliques}. 
In this paper, we choose \edit{the maximal clique with largest cardinality, \emph{i.e.,} the \emph{maximum clique}},
as the inlier set to pass to rotation estimation. 
\prettyref{sec:separateSolver} shows that this method drastically reduces the number of outliers.

In summary, the function ${\tt estimate\_s}$ in Algorithm~\ref{alg:ITR} first calls  Algorithm~\ref{alg:adaptiveVoting}, then computes the \edit{maximum clique} in the resulting graph \edit{to reject all measurements outside the clique.} 

{\bf What if the scale is known?} 
In some registration problems, the scale is known, e.g., the scale of the two point clouds is the same. In such a case, we can skip Algorithm~\ref{alg:adaptiveVoting} and set $\hats$ to be the known scale. 
Moreover, we can still use the \mcis method to largely reduce the number of outliers.


\subsection{Robust Rotation Estimation}
\label{sec:rotationEstimation}

The generative model~\eqref{eq:TIM} describes measurements $\TIMb_{ij}$ affected 
by bounded noise $\| \veps_{ij} \| \leq \beta_{ij}$ including potential outliers (when $\vo_{ij} \neq \bm{0}$).
Again, we estimate $\MR$ from the estimated scale $\hats$, the measurements $\TIMb_{ij}$ and the bounds $\beta_{ij}$ using a 
\TLS estimator:
\bea
\label{eq:TLSrotation}
\hatMR = \argmin_{\MR \in \Othree} \sumAllIM \min
\left( \frac{  \|  \TIMb_k - \hats \MR \TIMa_k \|^2 }{  \beta^2_k }  
\;,\; 
\barcsq \right)
\eea
where for simplicity we numbered the measurements from $1$ to $K = |\calE|$ and adopted the notation $\TIMa_k, \TIMb_{k}$ instead of $\TIMa_{ij}, \TIMb_{ij}$. \edit{We have also relaxed $\MR \in \SOthree$ to $\MR \in \Othree$, where $\Othree \doteq \setdef{\MR\in\Real{3 \times 3}}{\MR\tran \MR = \eye_3}$ is the \emph{Orthogonal Group}, which includes proper rotations and reflections. } For simplicity of notation, in the following we drop $\hats$ and assume that \edit{$\TIMa_1,\ldots,\TIMa_K$} have been corrected by the scale ($\TIMa_k \leftarrow \hats \TIMa_k$).


A fundamental contribution of this paper is to develop a tight convex relaxation for~\eqref{eq:TLSrotation}. The relaxation is tight even in presence of a large number (\maxOutliersRot) of outliers and provides per-instance suboptimality guarantees. 
Before presenting the relaxation, we  
introduce a binary formulation that is instrumental to develop the proposed relaxation.

\myParagraph{Binary formulation and \nameCloning} 
The first insight behind our convex relaxation is the fact that we can write the \TLS cost~\eqref{eq:TLSrotation} in additive form using auxiliary binary variables (a property recently leveraged in a different context by~\cite{Lajoie19ral-DCGM}): 
\bea
\label{eq:TLSrotation2}
\min_{
\substack{\MR \in \Othree,\\
\theta_k \in\{-1,+1\}, \forall k }} 
\sumAllIM 
\frac{ (1 + \theta_k) }{ 2 } \frac{  \|  \TIMb_k -  \MR \TIMa_k \|^2 }{  \beta^2_k }  
+ 
\frac{ (1 - \theta_k) }{ 2 } 
\barcsq 
\eea
The equivalence can be easily understood from the fact that $\min(x,y)=\min_{\theta \in\{-1,+1\}} 
\frac{ (1 + \theta ) }{ 2 } x + \frac{ (1 - \theta) }{ 2 } y$.

We conveniently rewrite~\eqref{eq:TLSrotation2} by replacing the binary variables with suitable (orthogonal) matrices. 
\begin{proposition}[\nameCloning]\label{prop:cloning}
Problem~\eqref{eq:TLSrotation2}
is equivalent to the following optimization problem
\bea
\hspace{-3mm}
\min_{
\substack{\MR, \MR_k, \forall k }} %
\!\!&
\!\!\sumAllIM 
\frac{  \| \TIMb_k - \MR \TIMa_k 
+ \MR\tran \MR_k  \TIMb_k - \MR_k \TIMa_k \|^2 }{ 4 \beta^2_k }  
+ 
\frac{ (1 - \ve_1\tran \MR\tran \MR_k \ve_1) }{ 2 } 
\barcsq \hspace{-5mm} \nonumber 
\\
\subject 
\hspace{-7mm}&
\MR\tran\MR=\eye_3, \;\; \MR_k\tran\MR_k=\eye_3,
\nonumber 
 \\
& \MR\tran\MR_k \in\{-\eye,+\eye\},
\; k=1,\ldots,\nrTIM \hspace{-5mm}\label{eq:TLSrotation4}
\eea
where we introduced a matrix $\MR_k \in \Real{3\times3}$ for each $k=1,\ldots,\nrTIM$, and defined the vector $\ve_1 \doteq [1 \; 0 \; 0]\tran$. 
\end{proposition}

A formal proof of~\prettyref{prop:cloning} is given in the \supp. 
 We name the re-parametrization in~\prettyref{prop:cloning} \emph{binary cloning}, since we now have $\nrTIM$ clones of $\MR$ 
 (namely $\MR_k=\theta_k \MR \in \{\MR,-\MR\}$, $k=1,\ldots,\nrTIM$) that 
 are in charge of rejecting outliers: when $\MR_k = \MR$ the $k$-th term in the objective becomes $\frac{  \| \TIMb_k - \MR \TIMa_k \|^2 }{ \beta^2_k }$ (i.e., $k$ is treated as an inlier, similarly to choosing $\theta_k=+1$), while when $\MR_k = -\MR$ the $k$-th term is equal to $\barcsq$ (i.e., $k$ is treated as an outlier, similarly to choosing $\theta_k=-1$).
This reparametrization enables our relaxation.

\myParagraph{Convex relaxation} 
The proposed relaxation  is presented in~\prettyref{prop:TLSrotationRelax}. 
The main goal of this paragraph is to provide the intuition behind our relaxation, while the interested 
reader can find a formal derivation in the \supp. 

Let us define a $3 \times 3(\nrTIM+2)$ matrix $\MX = [\eye_3 \; \MR \; \MR_1 \; \ldots \; \MR_\nrTIM]$, stacking all unknown variables in~\eqref{eq:TLSrotation4}. We observe that the matrix $\MZ \doteq \MX\tran \MX$ contains all linear and quadratic terms in $\MR$ and $\MR_k$:
\bea
\label{eq:Z}
\MZ \doteq\MX\tran \MX  = 
\begin{array}{ccccccc}
\hspace{-4mm}\overbrace{}^{I} \hspace{1mm} \overbrace{}^{R} \hspace{1mm} \overbrace{}^{R_1} \hspace{5mm} \ldots \hspace{5mm}  \overbrace{}^{R_\nrTIM}  \\
\left[
\begin{array}{ccccccc}
\eye_3 & \MR & \MR_1 & \ldots & \MR_\nrTIM \\
\star &   \eye_3   &  \MR\tran\MR_1     &   \ldots     &      \MR\tran\MR_\nrTIM       \\
\star &    \star        &  \eye_3            &   \ldots     &        \MR_1\tran\MR_\nrTIM     \\
\star &    \vdots        &  \vdots         &   \ddots     &    \vdots        \\
\star &    \star        &  \star   &     \ldots    &    \eye_3        
\end{array}
\right]
\end{array}
\eea
Now it is easy to see that the cost function in~\eqref{eq:TLSrotation4} can be rewritten as a function  of $\MZ$, by 
noting that $\MR, \MR\tran \MR_k, \MR_k$ (appearing in the cost) are entries of $\MZ$, see~\eqref{eq:Z}. 
The constraints in~\eqref{eq:TLSrotation4} can be similarly written as a function of $\MZ$. For instance, 
the constraints $\MR\tran\MR=\eye_3$ and $\MR_k\tran\MR_k=\eye_3$ simply enforce that the block diagonal entries of $\MZ$ are identity matrices. Similarly, $\MR\tran\MR_k \in\{-\eye,+\eye\}$ can be rewritten as a (non-convex) constraint involving off-diagonal entries of $\MZ$.
Finally, the fact that $\MZ \doteq \MX\tran \MX$ implies $\MZ$ is positive semidefinite and has rank $3$ (number of rows in $\MX$, see~\eqref{eq:Z}).

According to the discussion so far, we can reparametrize problem~\eqref{eq:TLSrotation4} 
 using $\MZ$, and we can then develop a convex relaxation by relaxing all the resulting non-convex constraints.
 This is formalized in the following proposition.

\begin{proposition}[\TLS Rotation Estimation: Convex Relaxation]\label{prop:TLSrotationRelax}
The following convex program is a relaxation of~\eqref{eq:TLSrotation4}:  
\bea
\min_{
\substack{\MZ \succeq 0}} %
& \hspace{-5mm} \trace{\barMQ \MZ} \hspace{-3mm}\label{eq:TLSrotationRelax}
\\
\subject 
& \hspace{-5mm} [\MZ]_{RR} =\eye_3, 
\;\;\; [\MZ]_{R_k R_k} =\eye_3,
\;\;\; [\MZ]_{II} =\eye_3,
\nonumber \hspace{-3mm}
 \\
& \hspace{-5mm} [\MZ]_{RR_k} = (\ve_1\tran [\MZ]_{RR_k} \ve_1) \eye_3, \;\; \forall k \nonumber \hspace{-3mm}\\
& \hspace{-5mm} \|  [\MZ]_{IR} \pm [\MZ]_{IR_k} \| \leq  1 \pm (\ve_1\tran [\MZ]_{RR_k} \ve_1), \;\; \forall k \nonumber \hspace{-3mm}\\
& \hspace{-5mm} \|  [\MZ]_{IR_k} \pm [\MZ]_{IR_{k'}} \| \leq  1 \pm (\ve_1\tran [\MZ]_{R_k R_{k'}} \ve_1), \;\; \forall k,k' \nonumber \hspace{-3mm}
\eea
where $\bar\MQ$ is a known $3(\nrTIM+2) \times 3(\nrTIM+2)$ symmetric matrix (expression given in the supplementary), 
and $[\MZ]_{RR_k}$ denotes a $3\times3$ block of $\MZ$ whose row indices correspond to the location of $\MR$ in $\MX$ (\cf with indices at the top of the matrix in eq.~\eqref{eq:Z})
and column indices correspond to the location of $\MR_k$ in $\MX$ (similarly, for $[\MZ]_{RR}$, $[\MZ]_{R_k R_k}$, $[\MZ]_{II}$, etc.). 
\end{proposition}

The convex program~\eqref{eq:TLSrotationRelax} can be solved in polynomial time using off-the-shelf convex solvers, such 
as \cvx~\cite{CVXwebsite}. It is a relaxation, in the sense that the set of feasible solutions of~\eqref{eq:TLSrotationRelax} 
includes the set of feasible solutions of~\eqref{eq:TLSrotation4}. 
Moreover, it enjoys the typical per-instance guarantees of convex relaxations.

\begin{proposition}[Guarantees for \TLS Rotation Estimation]\label{prop:TLSrotation}
Let $\MZ^\star$ be the optimal solution of the relaxation~\eqref{eq:TLSrotationRelax}.
If $\MZ^\star$ has rank 3, then it can be factored as $\MZ^\star = (\MX^\star)\tran(\MX^\star)$, where $\MX^\star \!\doteq\! [\eye_3, {\MR}^\star, {\MR}^\star_1,\mydots, {\MR}^\star_\nrTIM]$ is the first block row of $\MZ^\star$. Moreover,  ${\MR}^\star, {\MR}^\star_1,\mydots, {\MR}^\star_\nrTIM$ is an optimal solution for~\eqref{eq:TLSrotation4}.
\end{proposition}

Empirically, we found that our relaxation is tight (i.e., \edit{numerically} produces a rank-3 solution) even when \maxOutliersRot of the \TIMs are outliers. 
 Even when the relaxation is not tight, one can still project $\MZ^\star$ to a feasible solution of~\eqref{eq:TLSrotation4} 
and obtain an upper-bound on how suboptimal the resulting solution is. 

In summary, the function ${\tt estimate\_R}$ in Algorithm~\ref{alg:ITR} solves the convex program~\eqref{eq:TLSrotationRelax} 
(e.g., using \cvx) to obtain a matrix $\MZ^\star$ and extracts the rotation estimate $\hatMR$ from $\MZ^\star$. 
In particular,  $\hatMR = [\MZ^\star]_{IR}$ if $\MZ^\star$ has rank 3 (\prettyref{prop:TLSrotation}), or $\hatMR$ is computed as the projection of 
$[\MZ^\star]_{IR}$ to \Othree otherwise.


\subsection{Robust Translation Estimation}
\label{sec:translationEstimation}
%
Since we already presented a polynomial-time solution for scalar \TLS in \prettyref{sec:scaleEstimation}, we propose to solve 
for the translation component-wise, i.e., 
%
we compute the entries $t_1, t_2, t_3$ of $\vt$ independently (see the \supp\ for details): 
 \bea
 \label{eq:TLSRegistrationT1}
 \min_{ \substack{ t_j } } 
 \sumAllPointsi \min \left( \frac{1}{\beta_i^2} \left|  [\vb_i - \hats \hatMR \va_i]_j - t_j \right|^2 \!\!,  \barcsq \right), \;\; j=1,2,3
 \eea
 where $[\cdot]_j$ denotes the $j$-th entry of a vector. 


In summary, the function ${\tt estimate\_t}$ in Algorithm~\ref{alg:ITR} calls Algorithm~\ref{alg:adaptiveVoting} three times 
(one for each entry of $\vt$) and returns the translation estimate $\hatvt = [t_1\;t_2\;t_3]$. 

\section{Experiments and Applications}
\label{sec:experiments}

The goal of this section is to (i) test the performance of our scale, rotation, translation solvers and the \mcis 
pruning (\prettyref{sec:separateSolver}), 
(ii) evaluate \name against related techniques in  benchmarking datasets (\prettyref{sec:benchmark}),
(iii) evaluate \name with \emph{extreme} outliers rates (\prettyref{sec:benchmarkExtreme}), and
(iv) show an application of \name for object localization in an RGB-D robotics dataset (\prettyref{sec:roboticsApplication}).
In all tests we set $\barcsq=1$. 

\myParagraph{Implementation details} We implemented \name in matlab and used \cvx to solve the convex relaxation~\eqref{eq:TLSrotationRelax}.
Moreover, we used the algorithm in~\cite{Eppstein10isac-maxCliques} to find all the maximal cliques in the pruned \TIM graph 
(see~\prettyref{thm:maxClique}). 

\subsection{Testing \name's Subproblems}
\label{sec:separateSolver}


\myParagraph{Testing setup}
We use the \bunny~\edit{point cloud} from the Stanford 3D Scanning Repository~\cite{Curless96siggraph} and resize 
\edit{it} to be within the $[0,1]^3$ cube.
 The \bunny is first downsampled to $\nrPoints=50$ points, and then a random transformation $(s, \MR, \vt)$ (with $1\leq s \leq 5$ and $\Vert \vt \Vert \leq 1$) is applied according to eq.~\eqref{eq:robustGenModel}. To generate the bounded noise $\veps_i$, we sample $\veps_i \sim \calN(\bm{0},\sigma^2 \MI)$, until the resulting vector satisfies $\|\veps_i\| \leq \beta_i=\beta$. 
 We set $\sigma = 0.01$ and $\beta = 0.0554$ such that $\mathbb{P}\left(\Vert \veps_i \Vert^2 / \sigma^2 > \beta^2 \right) \leq 10^{-6}$ 
 (this bound stems from the fact that for Gaussian $\veps_i$, $\Vert \veps_i \Vert^2$ follows a Chi-square distribution with 3 degrees of freedom).
  To generate outliers, we replace a fraction of $\vb_i$ with vectors uniformly sampled inside the sphere of radius 5.
  We test increasing outlier ratios $\{0,0.2,0.4,0.6,0.7,0.8,0.9\}$.
  All statistics are computed over 40 Monte Carlo runs.  

\begin{figure}[t]
	\begin{center}
	\begin{minipage}{\textwidth}
	\begin{tabular}{cc}%
	        \myhspace
			\begin{minipage}{\mpw}%
			\centering%
			\includegraphics[width=0.9\columnwidth]{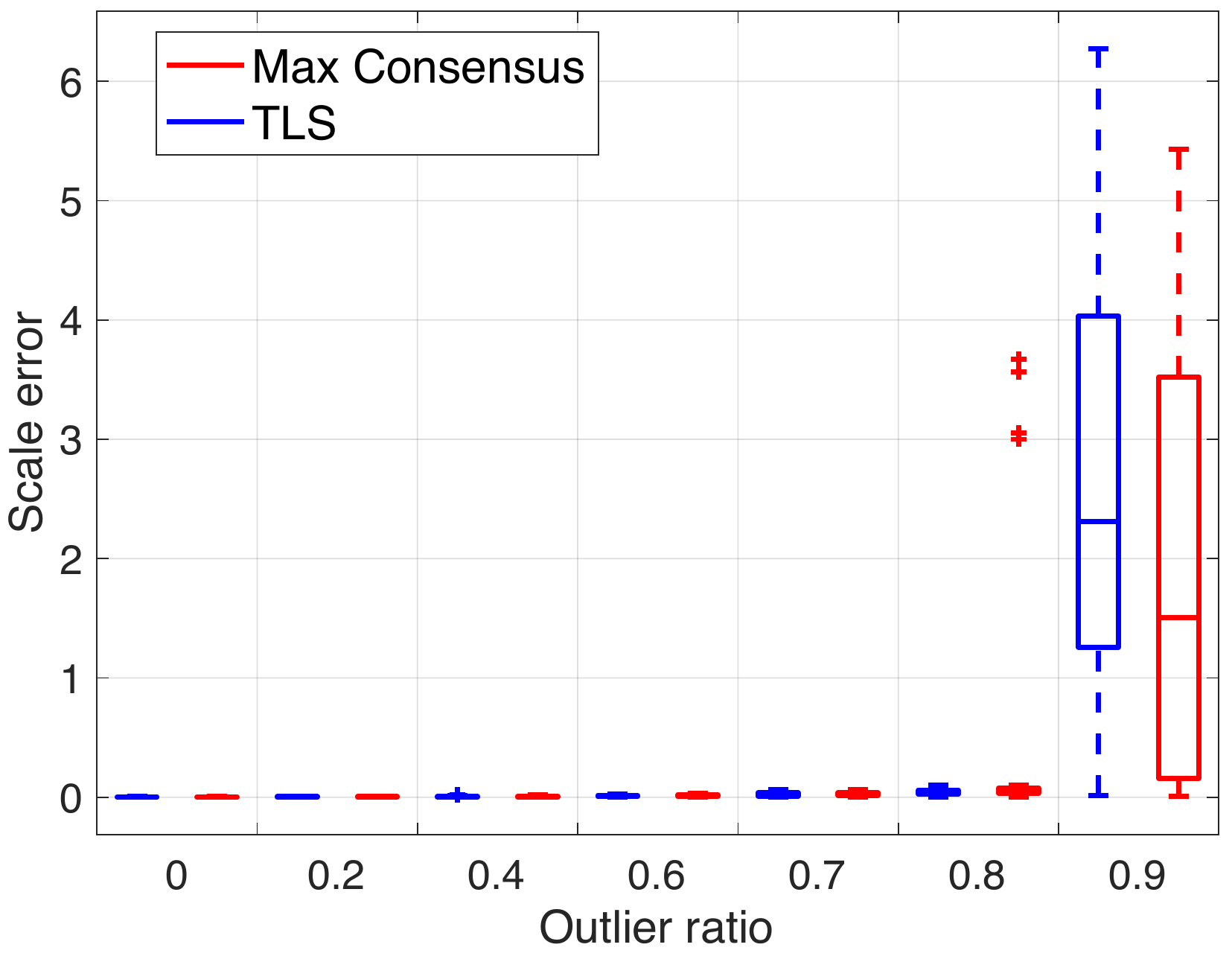} \\
			\vspace{-1mm}
			(a) Scale
			\end{minipage}
              & \myhspace 
			\begin{minipage}{\mpw}%
			\centering%
			\includegraphics[width=1\columnwidth]{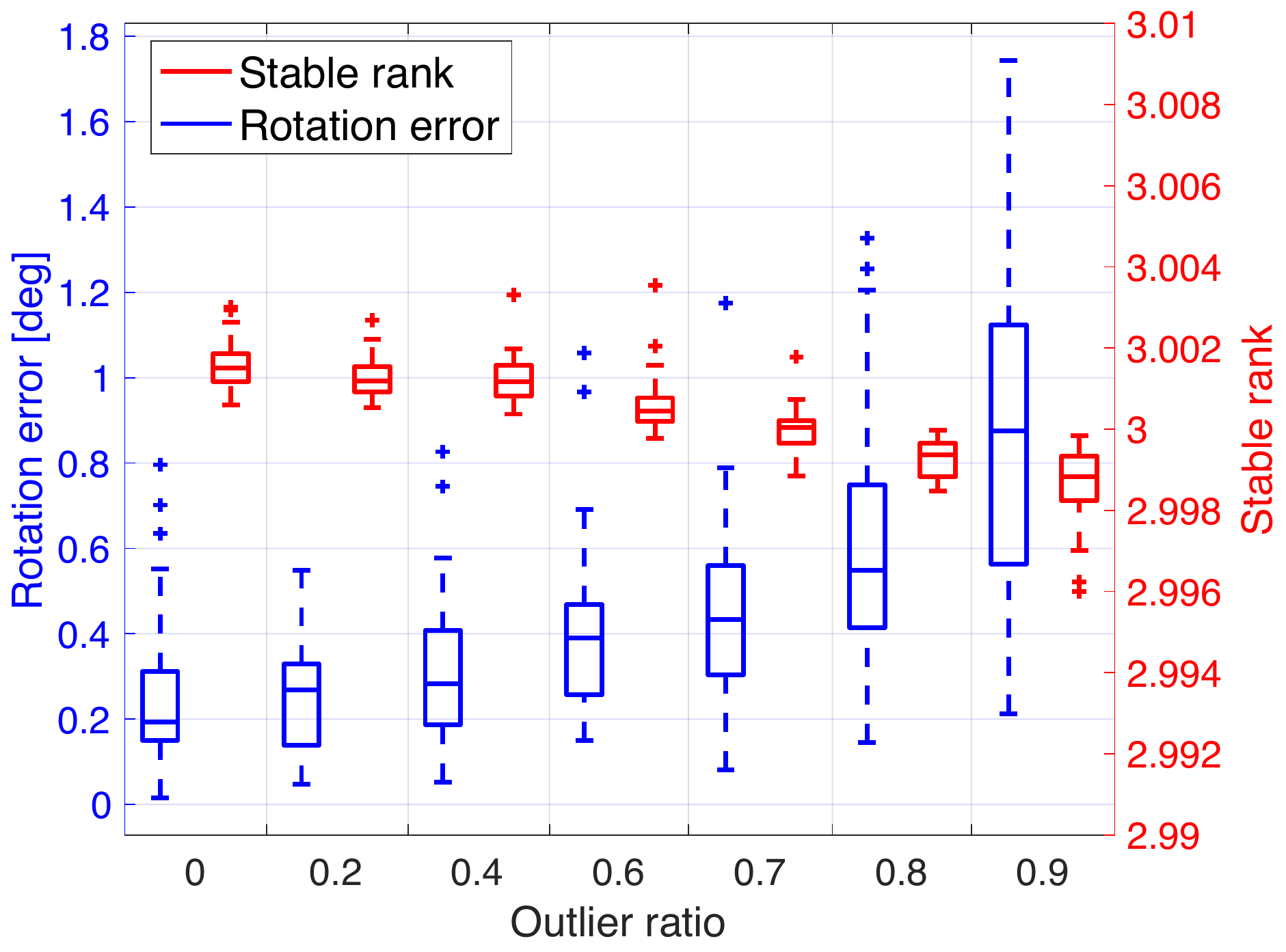} \\
			\vspace{-1mm}
			(b) Rotation
			\end{minipage} 
             \\
             \myhspace
			\begin{minipage}{\mpw}%
			\centering%
			\includegraphics[width=1\columnwidth]{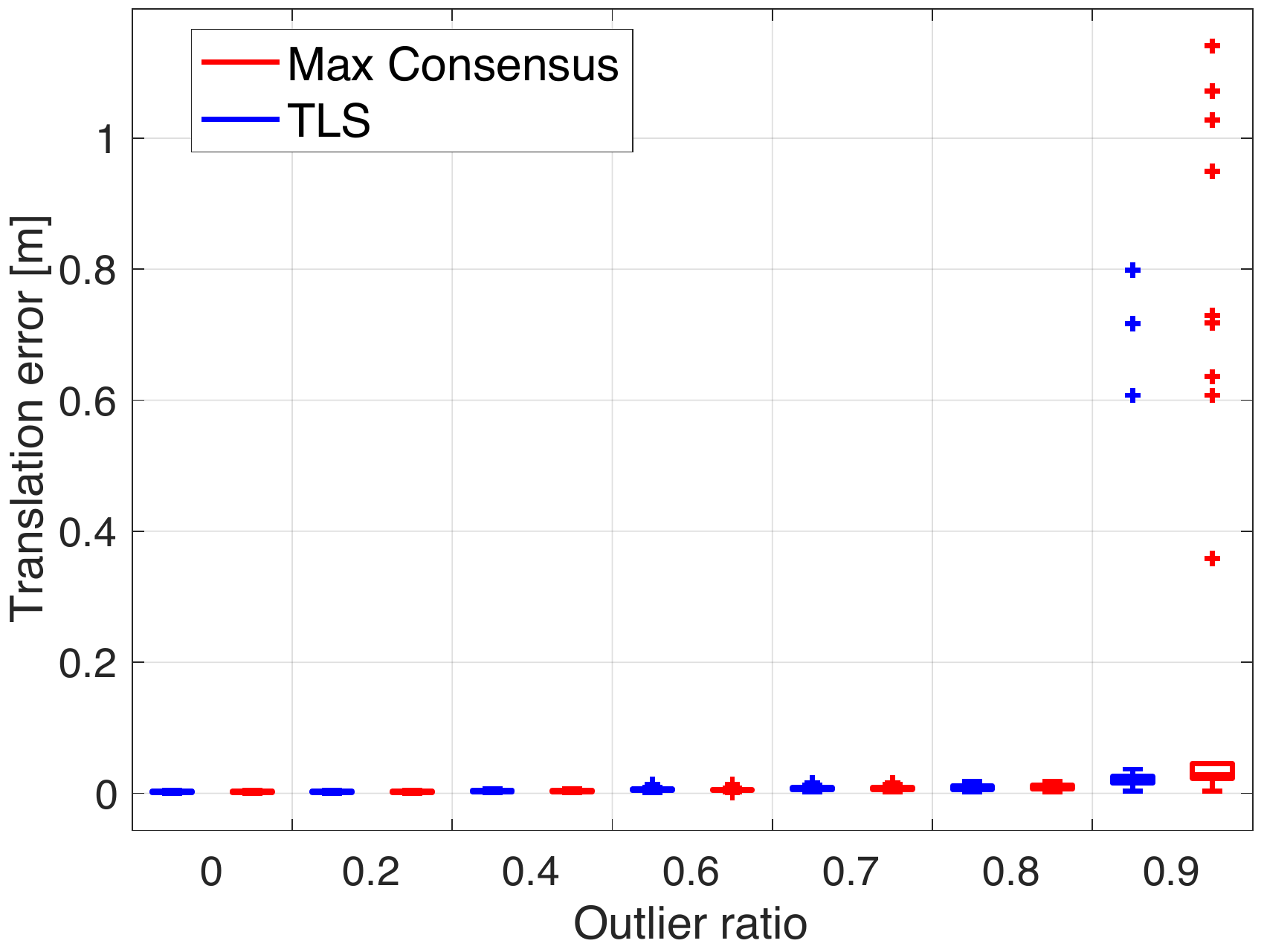} \\
			\vspace{-1mm}
			(c) Translation
			\end{minipage}
			& \myhspace
			\begin{minipage}{\mpw}%
			\centering%
			\includegraphics[width=0.95\columnwidth]{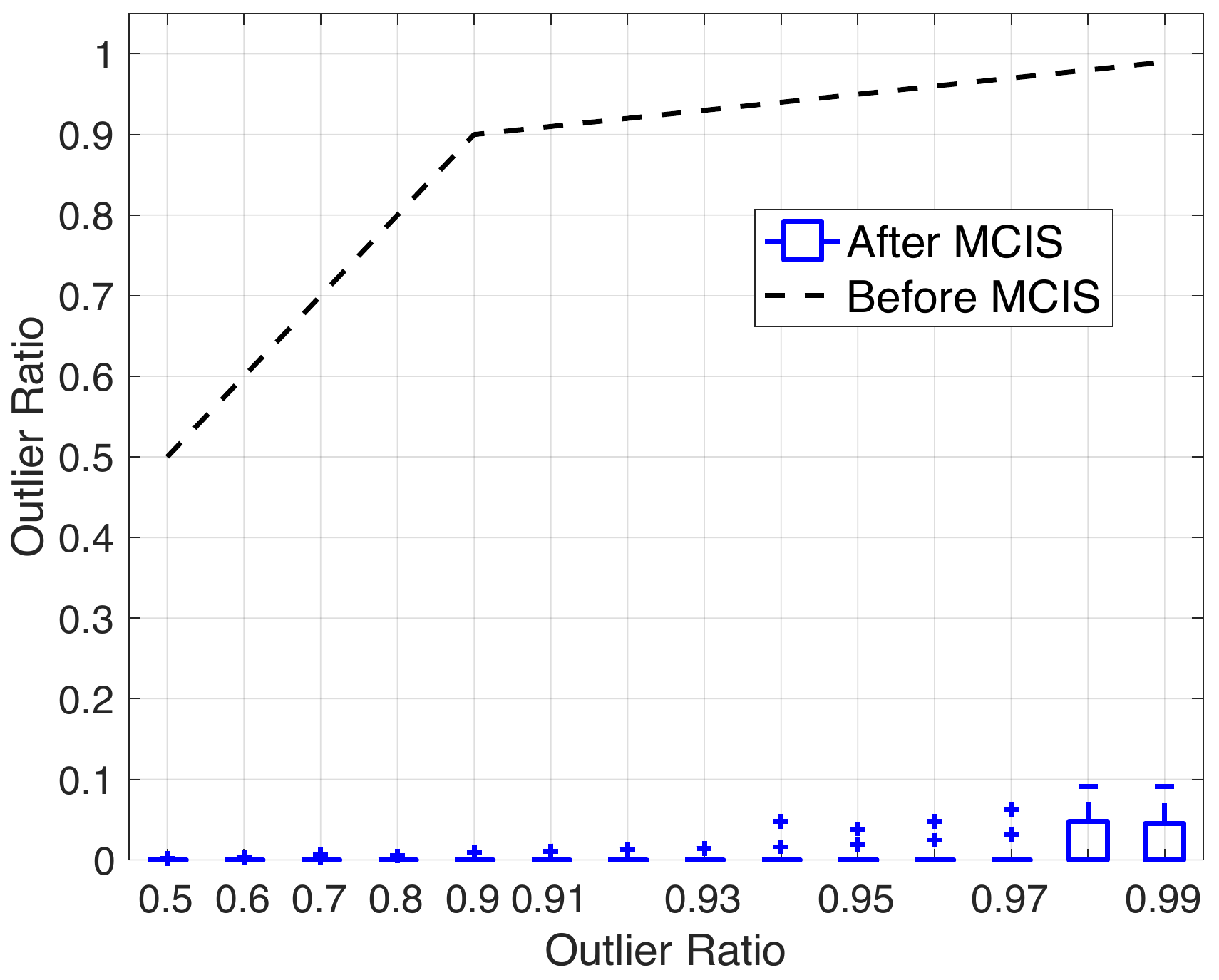} \\
			\vspace{-1mm}
			(d) \mcis
			\end{minipage}
		\end{tabular}
	\end{minipage}
	\begin{minipage}{\textwidth}
	\end{minipage}
	\vspace{-4mm} 
	\caption{Results for scale, rotation, translation estimation, and impact of  maximal clique pruning for increasing outlier ratios. 
	 \label{fig:benchmark_separate_solvers}}
	 	\vspace{-8mm} 
	\end{center}
\end{figure}

\myParagraph{Scale solver} Given two point clouds $\calA$ and $\calB$, we first create $\nrPoints (\nrPoints-1)/2$ \TIMs corresponding to a complete graph and then use Algorithm~\ref{alg:adaptiveVoting} to solve for the scale. We compute both \emph{maximum consensus}~\cite{Speciale17cvpr-consensusMaximization} and \TLS estimates of the scale.
Fig.~\ref{fig:benchmark_separate_solvers}(a) shows box plots of the scale error with increasing outlier ratios. \edit{The scale error is computed as $|s^\star - s_{gt}|$, where $s^\star$ is the scale estimate and $s_{gt}$ is the ground-truth}. \edit{We observe the \TLS solver is robust against 80\% outliers, while maximum consensus failed three times in that regime.}


\myParagraph{Rotation Solver} 
We apply a random rotation $\MR$ to the \bunny, and fix $s=1$ and $\vt = \zero$.
\edit{Two metrics are boxplotted in Fig.~\ref{fig:benchmark_separate_solvers}(b) to show the performance of the rotation solver: (i) the \emph{stable rank} of $\MZ^\star$, the optimal solution of SDP relaxation~\eqref{eq:TLSrotationRelax}, where the \emph{stable rank} is defined by the squared ratio between the Frobenius norm and the spectral norm; 
(ii) the rotation estimation error, defined as 
$\left| \arccos \left( \left(\trace{\MR_{gt}\tran \MR^\star} - 1 \right) / 2 \right) \right|$,  
\ie~the geodesic distance between the rotation estimate $\MR^\star$ and the ground-truth $\MR_{gt}$.}
We observe the stable rank is numerically close to 3 (\prettyref{prop:TLSrotationRelax}) even with \maxOutliersRot of outliers, and the rotation error remains below 2 degrees.

\myParagraph{Translation Solver} 
We apply a random translation $\vt$ to the \bunny, and fix $s=1$ and $\MR = \eye_3$.
\edit{Fig.~\ref{fig:benchmark_separate_solvers}(c) shows component-wise translation estimation using both \emph{maximum consensus} and \TLS are robust against 80\% outliers.}
\edit{ The translation error is defined as $\| \vt^\star - \vt_{gt} \|$, the  2-norm of the difference between the estimate $\vt^\star$ and the ground-truth $\vt_{gt}$. }

\myParagraph{Maximal Clique Inlier Selection} 
We downsample \bunny to $\nrPoints=1000$ and fix the scale to $s=1$ when applying the random transformation. 
We first prune the outlier \TIMs/\TRIMs (edges) that are not consistent with the scale $s=1$, while keeping all the points (nodes), to obtain 
the graph $\calG'$. Then we compute the \edit{\emph{maximum clique}}
in $\calG'$ using the algorithm in~\cite{Eppstein10isac-maxCliques}, 
and remove all edges and nodes outside the clique, obtaining a pruned graph $\calG"$.
Fig.~\ref{fig:benchmark_separate_solvers}(d) shows the outlier ratio in $\calG'$ (label: ``Before \mcis'') and $\calG"$ (label: ``After \mcis'').
The \mcis procedure effectively reduces the amount of outliers to below 10\%, facilitating 
rotation and translation estimation, which, in isolation, can already tolerate more than \maxOutliersRot outliers. 


\newcommand{\mpwthree}{6cm}

\begin{figure*}[h]
	\begin{center}
	\begin{minipage}{\textwidth}
	\hspace{-0.2cm}
	\begin{tabular}{ccc}%
			\begin{minipage}{\mpwthree}%
			\centering%
			\includegraphics[width=0.8\columnwidth]{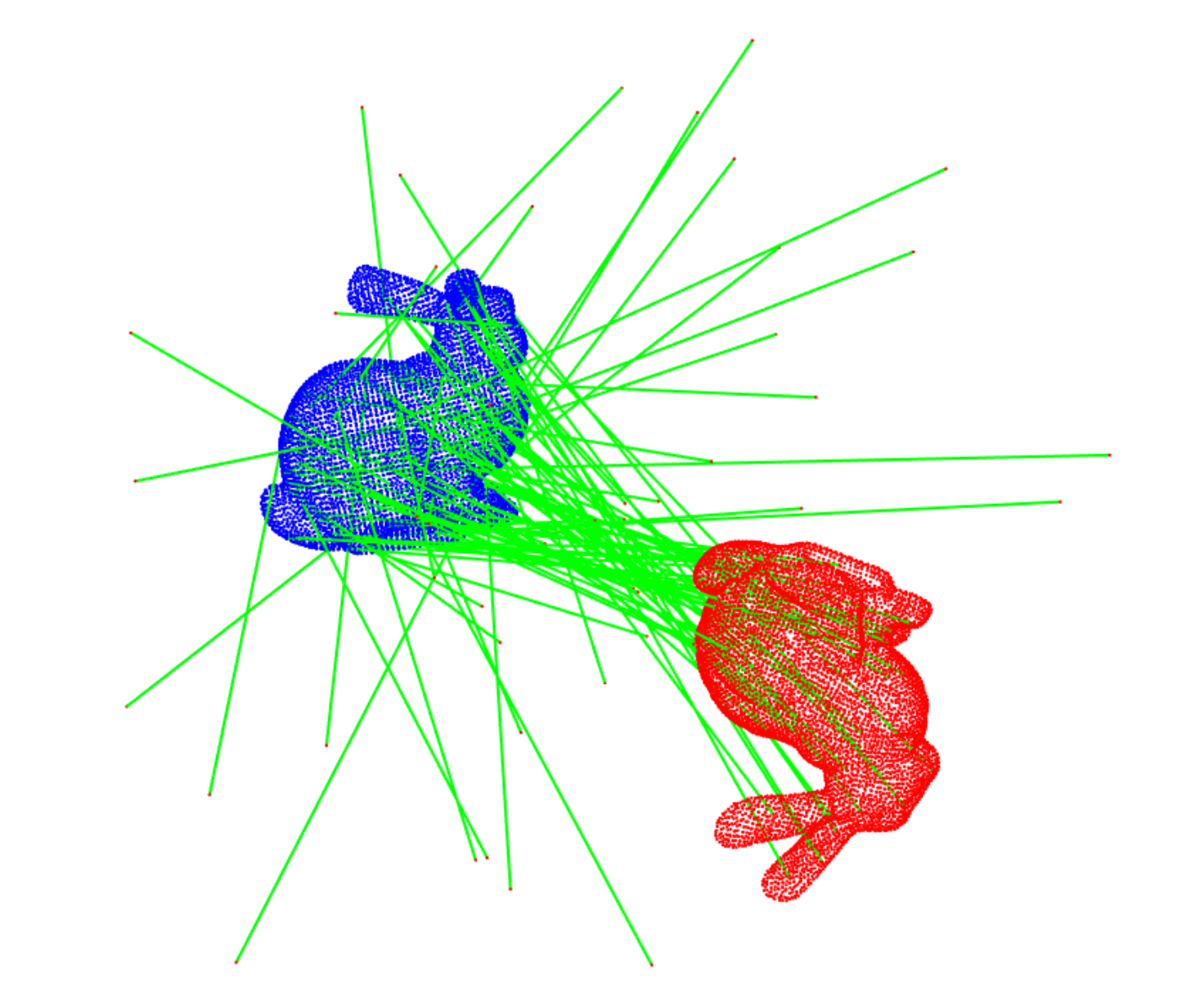} \\
			\end{minipage}
		& \myhspace
			\begin{minipage}{\mpwthree}%
			\centering%
			\includegraphics[width=0.9\columnwidth]{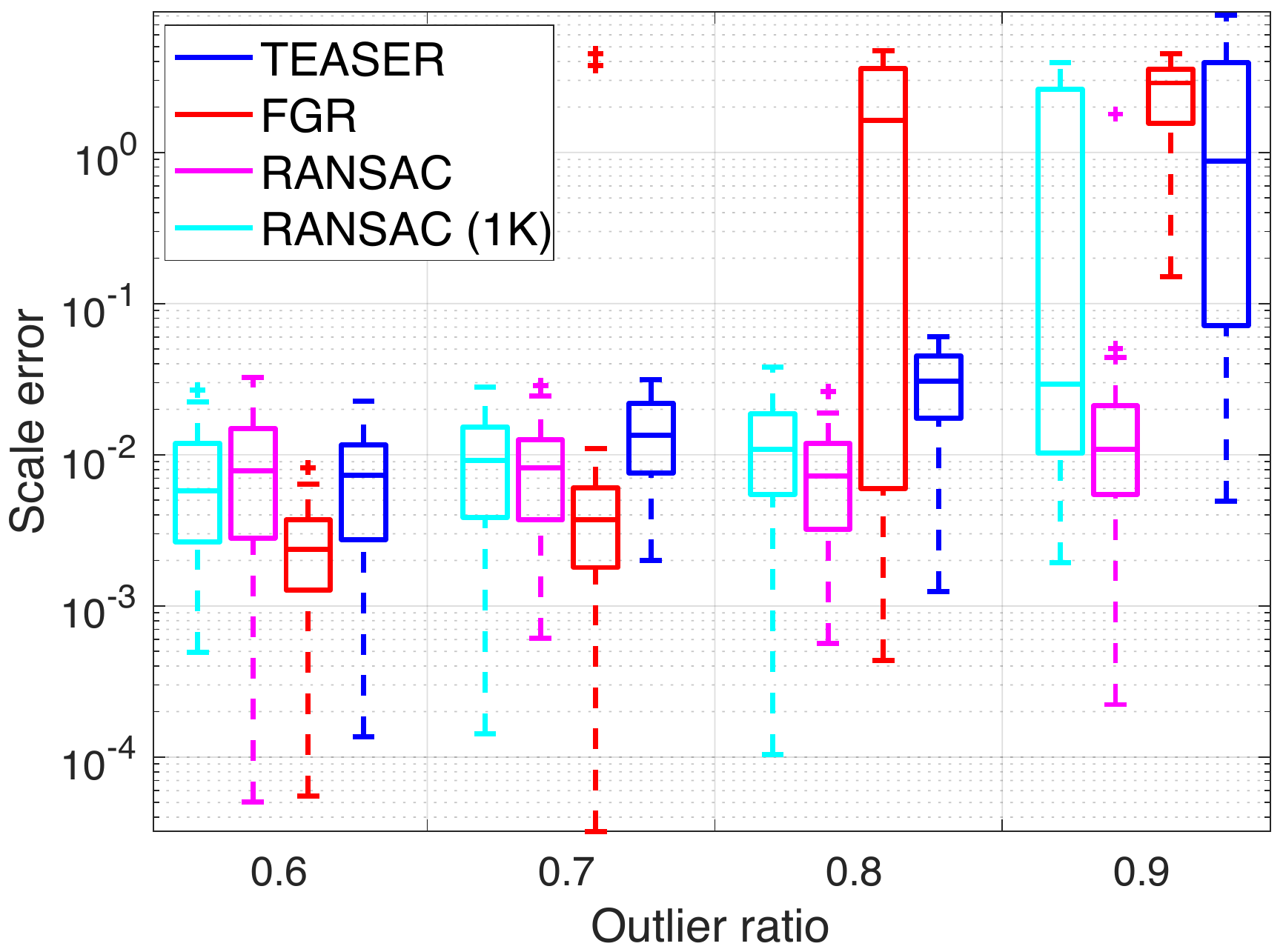} \\
			\end{minipage}
			\vspace{-1mm}
		& \myhspace
			\begin{minipage}{\mpwthree}%
			\centering%
			\includegraphics[width=0.9\columnwidth]{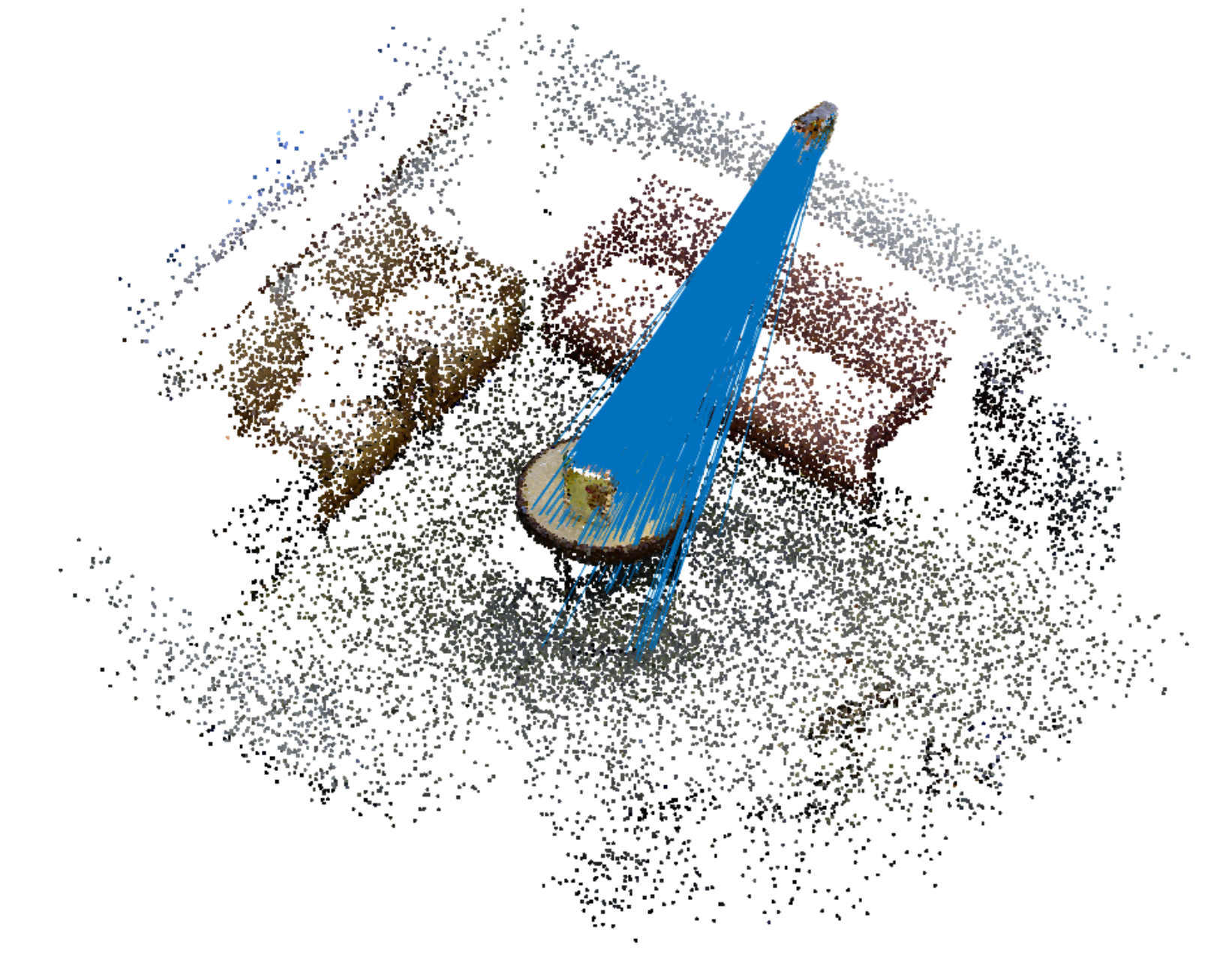} \\
			\end{minipage}  \\
			
		\begin{minipage}{\mpwthree}%
			\centering%
			\includegraphics[width=0.9\columnwidth]{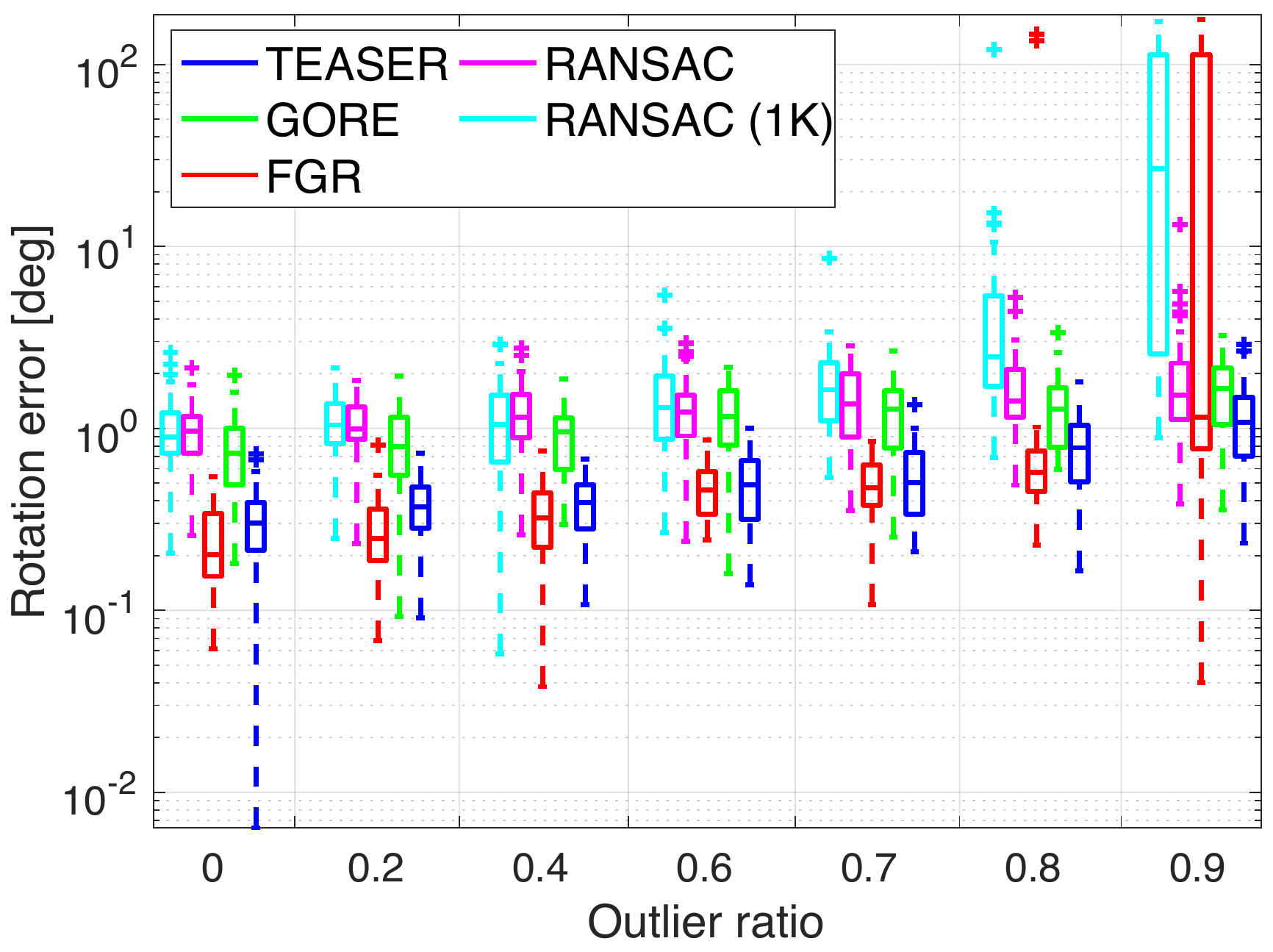} \\
			\end{minipage}
		& \myhspace
			\begin{minipage}{\mpwthree}%
			\centering%
			\includegraphics[width=0.9\columnwidth]{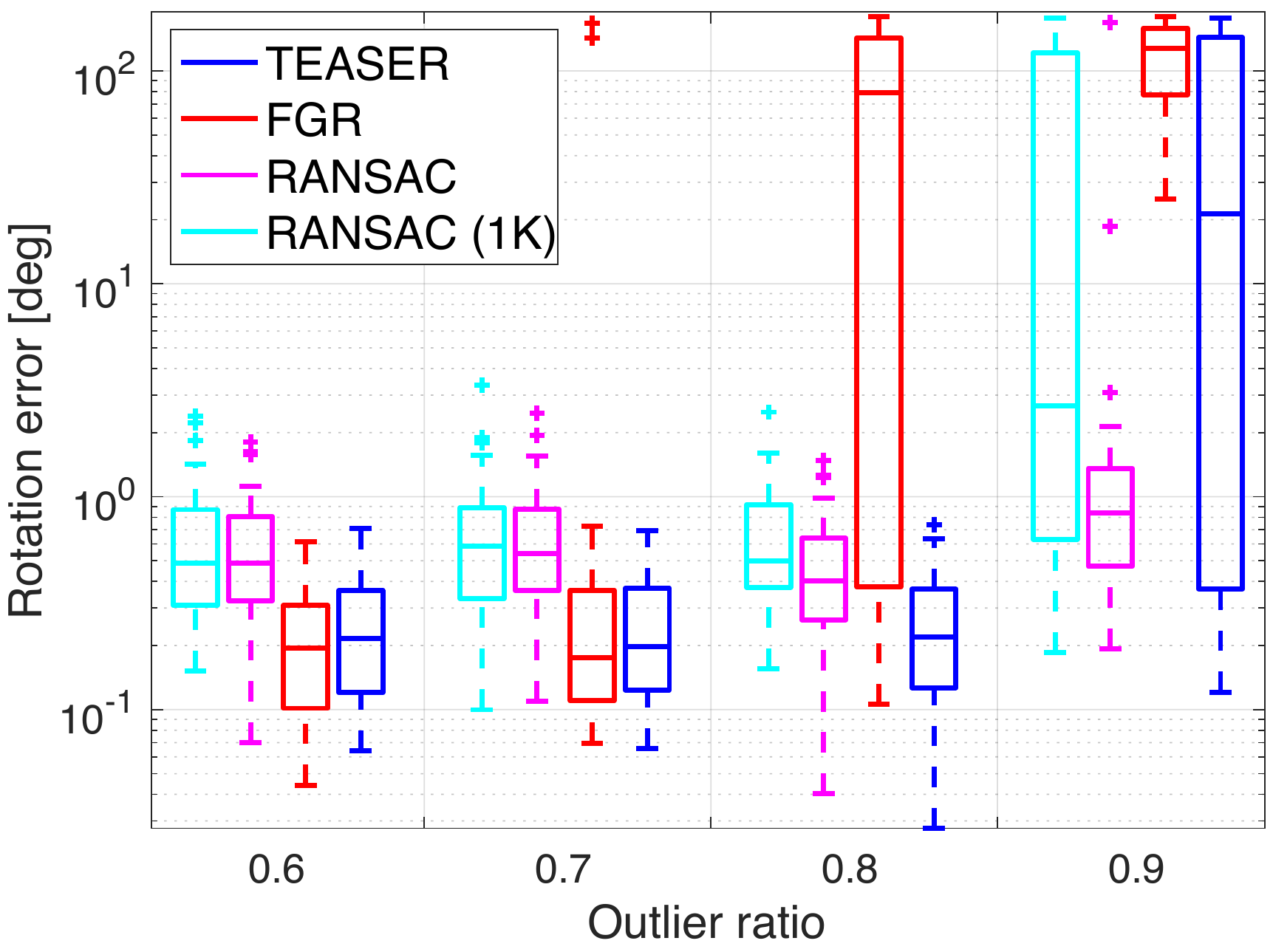} \\
			\end{minipage}
		& \myhspace
			\begin{minipage}{\mpwthree}%
			\centering%
			\includegraphics[width=0.9\columnwidth]{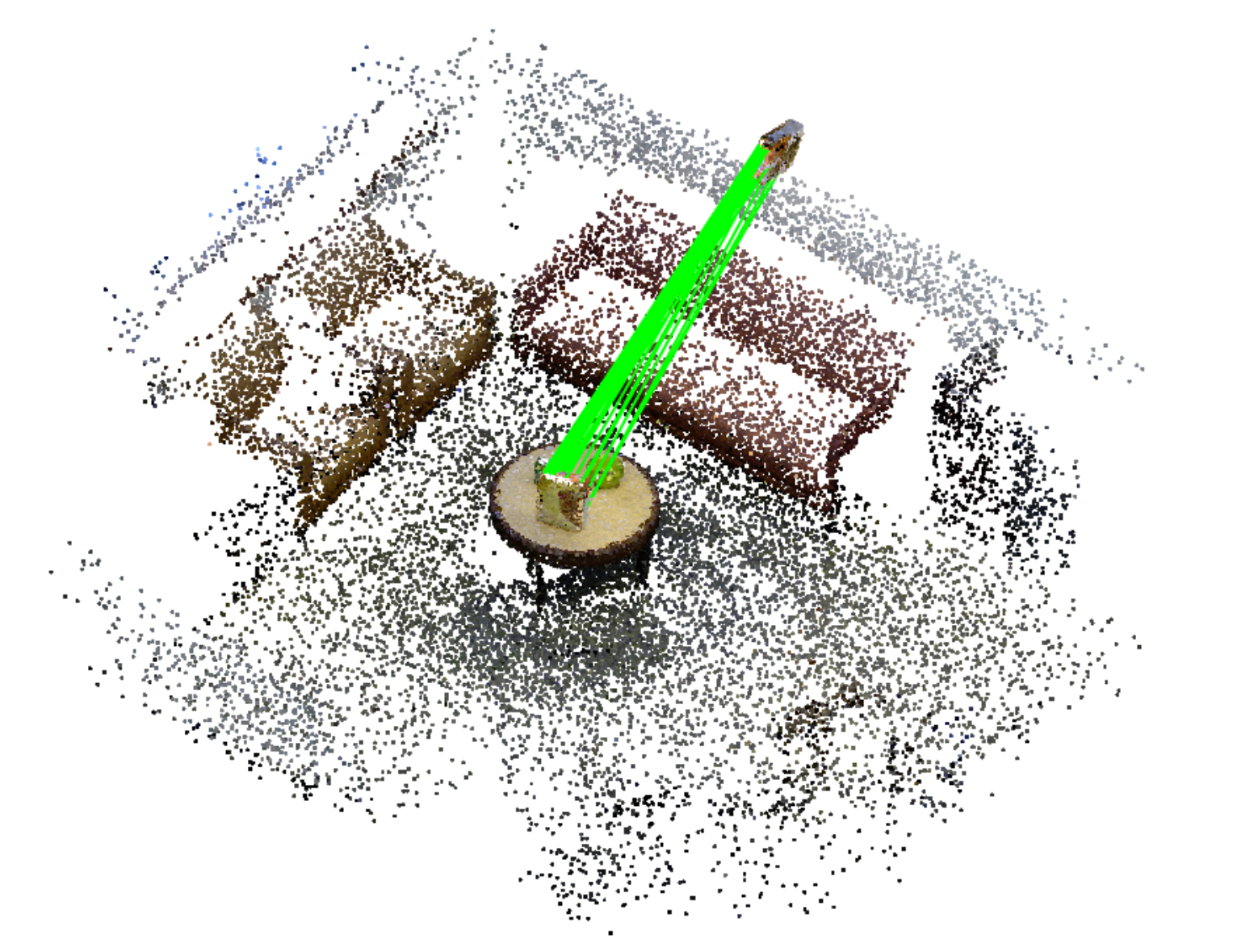} \\
			\end{minipage}\\
			
		\begin{minipage}{\mpwthree}%
			\centering%
			\includegraphics[width=0.9\columnwidth]{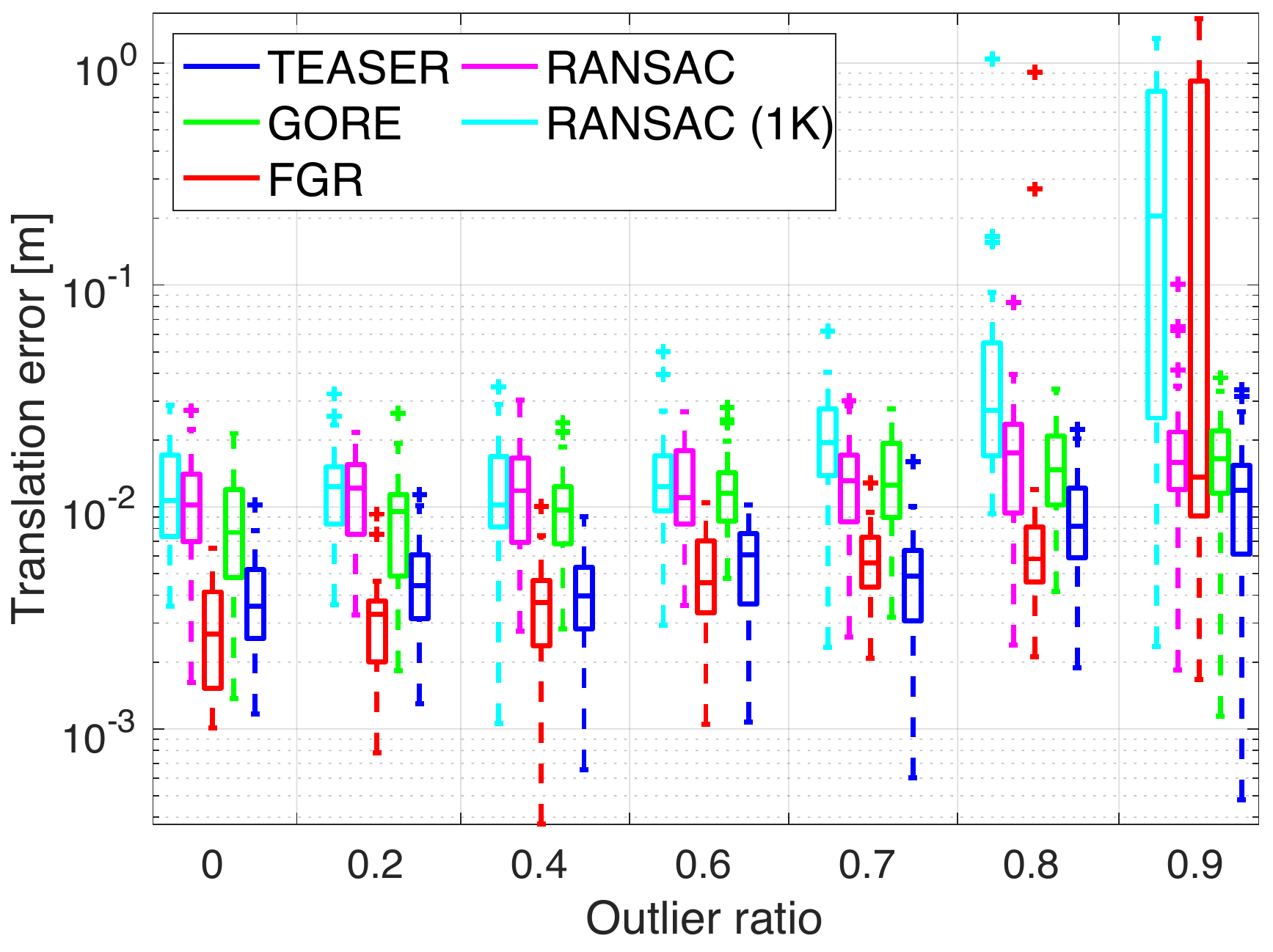} \\
			(a) Registration with known scale
			\end{minipage}
		& \myhspace
			\begin{minipage}{\mpwthree}%
			\centering%
			\includegraphics[width=0.9\columnwidth]{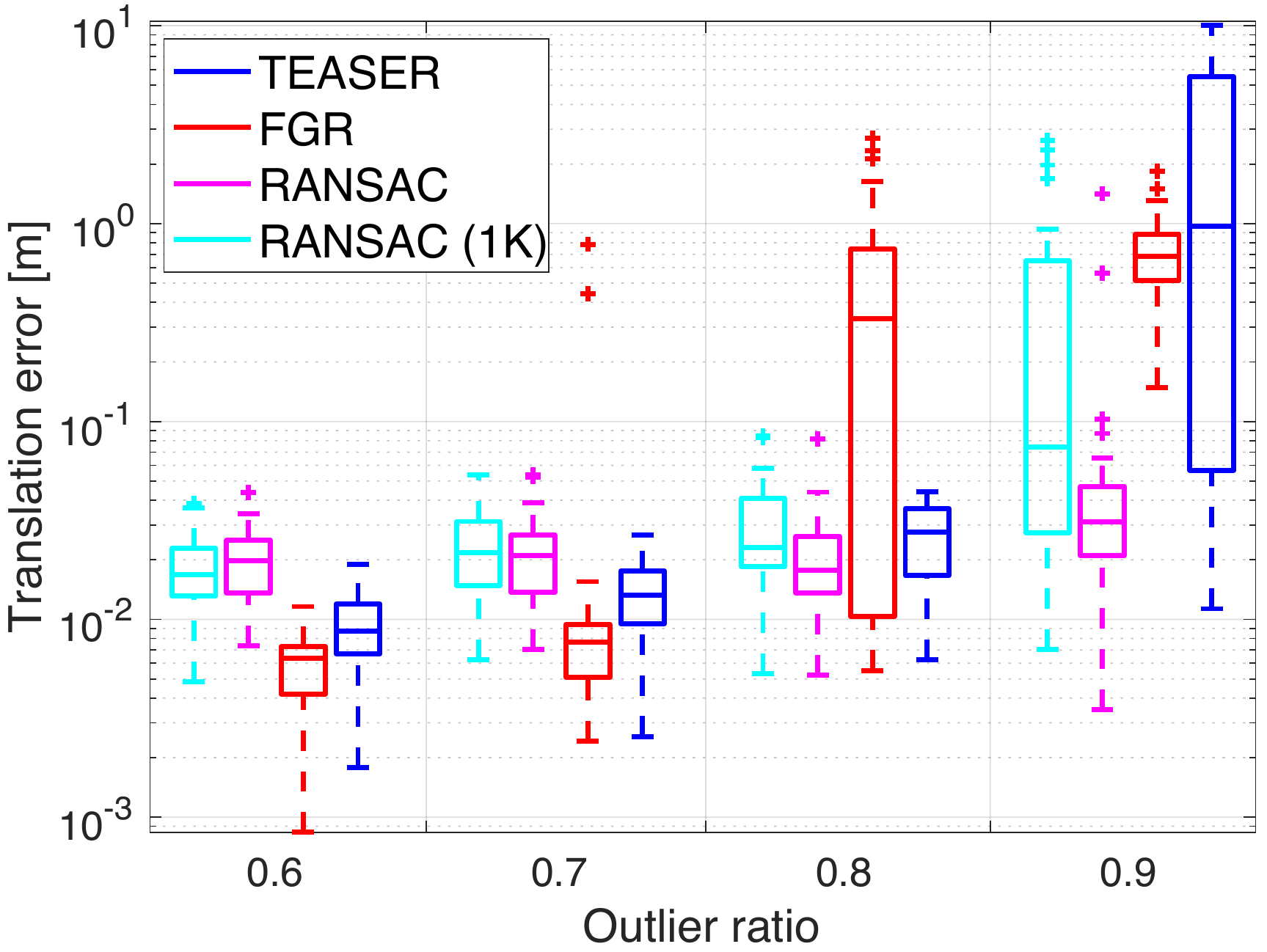} \\
			(b) Registration with unknown scale
			\end{minipage}
		& \myhspace
			\begin{minipage}{\mpwthree}%
			\centering%
			\includegraphics[width=0.85\columnwidth]{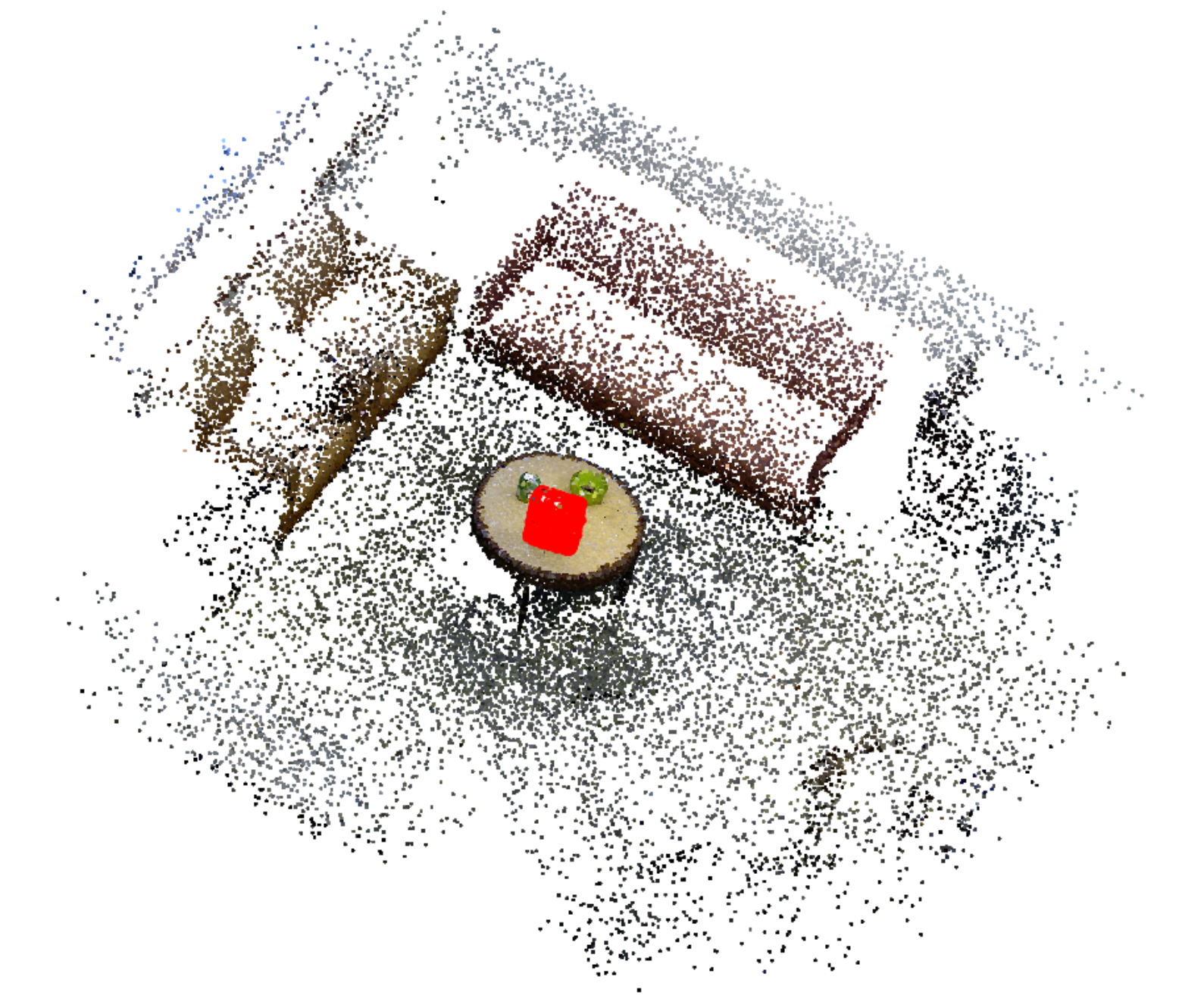} \\
			\vspace{-3mm}
			(c) Registration on robotics datasets
			\end{minipage}
		\end{tabular}
	\end{minipage}
	\vspace{-3mm} 
	\caption{Benchmark result. (a) Boxplots of rotation and translation errors for the five compared methods on the \bunny dataset with known scale (the top figure shows an example with 50\% outliers). (b) Boxplots of scale, rotation and translation errors for four registration methods on the \bunny dataset with unknown scale. (c) Successful object pose estimation by \name on a real RGB-D dataset. Blue lines are the original FPFH~\cite{Rusu09icra-fast3Dkeypoints} correspondences with outliers, green lines are the inlier correspondences computed by \name, and the final registered object is highlighted in red.}
	 \label{fig:benchmark}
	\vspace{-7.5mm} 
	\end{center}
\end{figure*}

\subsection{Benchmarking on Standard Datasets}
\label{sec:benchmark}

\myParagraph{Testing setup}
We benchmark \name against two state-of-the-art robust registration techniques: \emph{Fast Global Registration} (\FGR)~\cite{Zhou16eccv-fastGlobalRegistration} and \emph{Guaranteed Outlier REmoval} (\GORE)~\cite{Bustos18pami-GORE}. 
In addition, we test two \ransac variants: a fast version where we terminate \ransac after a maximum of  1,000 iterations (\ransaconek) and 
a slow version where we terminate \ransac after 60s (\ransac). Four datasets, \emph{\bunny}, \emph{\armadillo}, \emph{\dragon} and \emph{\buddha}, from the Stanford 3D Scanning Repository are selected and downsampled to $\nrPoints=100$ points. 
The tests below follow the same protocol of \prettyref{sec:separateSolver}. \edit{In the \supp, we provide an example of the performance of \name on registration problems with high noise ($\sigma=0.1$).}

{\bf Known Scale.} We first evaluate the compared techniques with known scale $s=1$. 
Fig.~\ref{fig:benchmark}(a) shows the rotation and translation error at increasing outlier ratios for the \bunny dataset. \name, \GORE and \ransac are robust against up to 90\% outliers, although \name tends to produce more accurate estimates than \GORE, and \ransac typically requires over $10^5$ iterations for convergence at 90\% outlier rate. \FGR can only resist 70\% outliers and \ransaconek starts breaking at 60\% outlier rate. 
These conclusions are confirmed by the results on the other three datasets ({\armadillo}, {\dragon}, {\buddha), which are \edit{given} in the \supp due to space constraints.

{\bf Unknown Scale.} 
\GORE is unable to solve for the scale, hence we only benchmark \name against \FGR (although the original algorithm in~\cite{Zhou16eccv-fastGlobalRegistration} did not solve for the scale, we extend it by using Horn's method to compute the scale at each iteration), \ransaconek and \ransac. Fig.~\ref{fig:benchmark}(b) plots the scale, rotation and translation error for increasing outlier ratios on the \bunny dataset. 
All the compared techniques perform well when the outlier ratio is below 60\%. \FGR has the lowest breakdown point and fails at 80\%. \ransaconek and \name only fail at 90\% outlier ratio when the scale is unknown. Although \ransac with 60s timeout outperforms other methods at 90\% outlier rate, it typically requires more than $10^5$ iterations to converge, which is not practical for real-time applications.


\subsection{Testing under Extreme Outlier Rates}
\label{sec:benchmarkExtreme}
We further benchmark the performance of \name under extreme outlier rates from 95\% to 99\% with known scale and $N=1000$ correspondences on the \bunny. We replace \ransaconek with \ransactenk, since \ransaconek already performs poorly at 90\% outlier ratio.
Fig.~\ref{fig:benchmark_extreme_outlier} shows the boxplots of the rotation and translation errors.
Both \name and \GORE are robust against up to 99\% outliers, while \ransac with 60s timeout can resist 98\% outliers with about $10^6$ iterations. \ransactenk and \FGR perform poorly under extreme outlier ratios. While \GORE and \name are both robust against 99\% outliers, \name produces slightly lower estimation errors.

\begin{figure}[t]
	\begin{center}
	\begin{minipage}{\textwidth}
	\begin{tabular}{cc}%
	\myhspace
			\begin{minipage}{\mpw}%
			\centering%
			\includegraphics[width=\columnwidth]{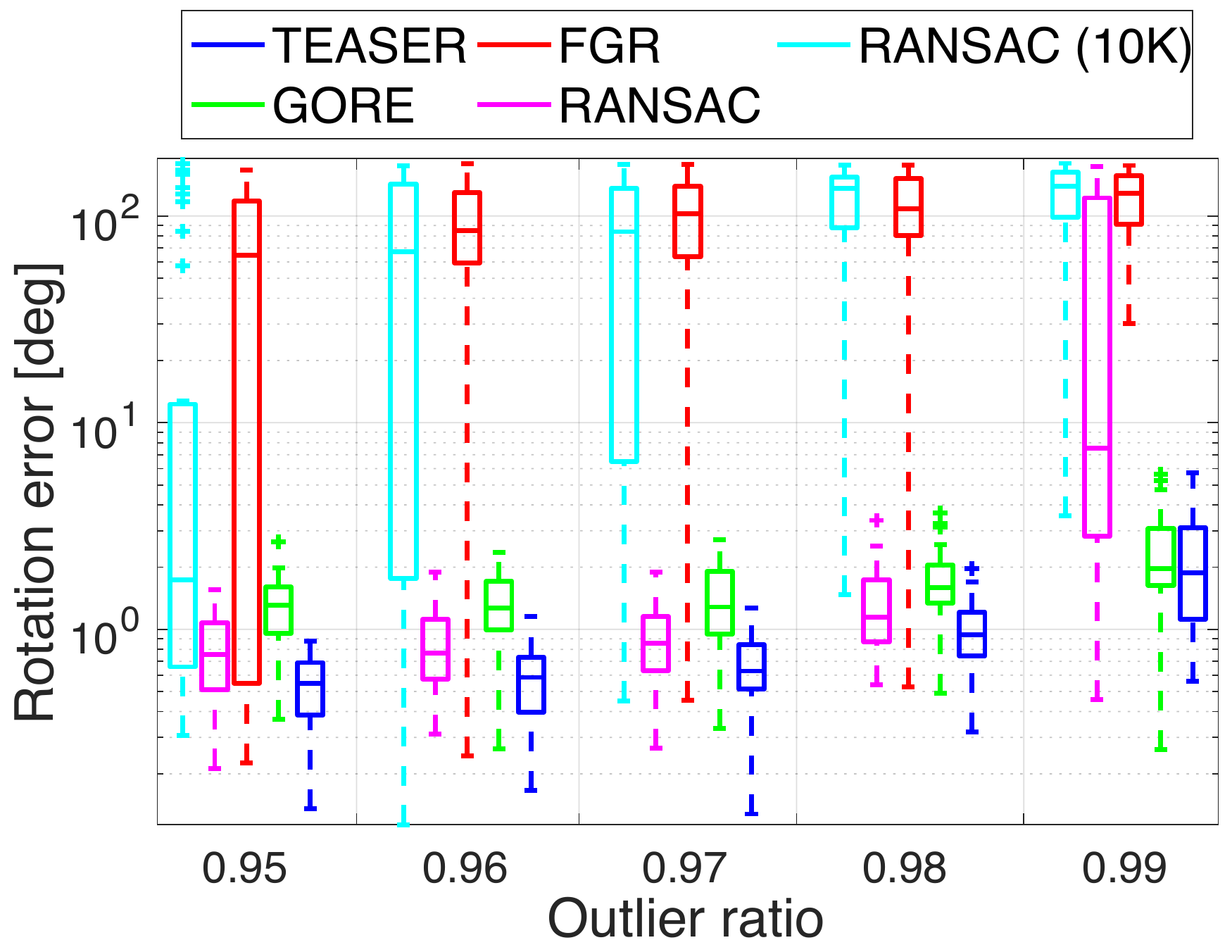} \\ \vspace{-1mm}
			(a) Rotation Error
			\end{minipage}
		& \myhspace 
			\begin{minipage}{\mpw}%
			\centering%
			\includegraphics[width=\columnwidth]{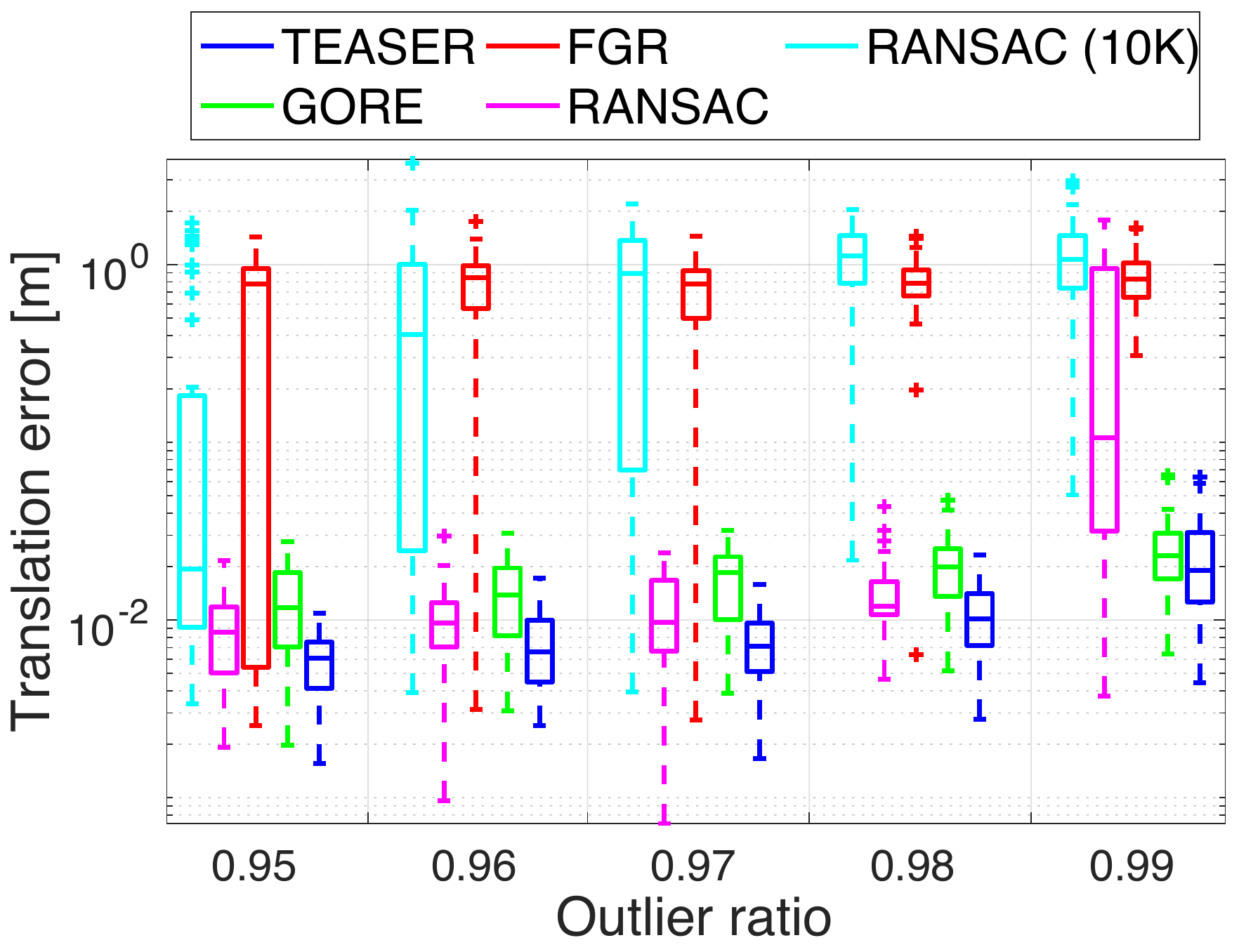} \\ \vspace{-1mm}
			(b) Translation Error
			\end{minipage}
		\end{tabular}
	\end{minipage}
	\begin{minipage}{\textwidth}
	\end{minipage}
	\vspace{-4mm} 
	\caption{Estimation errors under extreme outlier rates (known scale).
	 \label{fig:benchmark_extreme_outlier}}
	\vspace{-9.5mm} 
	\end{center}
\end{figure}

\subsection{Application: Object Pose Estimation and Localization}
\label{sec:roboticsApplication}
We use the large-scale point cloud datasets from~\cite{Lai11icra-largeRGBD} to test \name in \emph{object pose estimation and localization} applications. 
We first use the ground-truth object labels to extract the \emph{cereal box/cap} out of the scene and treat it as the 
object, then apply a random transformation to the scene, to get an object-scene pair. To register the object-scene pair, we first use FPFH feature descriptors~\cite{Rusu09icra-fast3Dkeypoints} to establish putative correspondences.
Then, \name is used to find the relative pose. Fig.~\ref{fig:benchmark}(c) shows the noisy FPFH correspondences, the inlier correspondences obtained by \name, and successful localization and pose estimation of the \emph{cereal box}. \edit{The \supp provides results on more object-scene pairs.}

%
%
%
%

\section{Conclusion} 
\label{sec:conclusion}

We propose a \emph{Truncated Least Squares} approach to compute the relative transformation (scale, rotation, translation) 
that aligns two point clouds in the presence of extreme outlier rates. 
We present a general graph-theoretic framework to decouple rotation, translation, and scale estimation.
%
We provide a polynomial-time solution for each subproblem: 
 scale and (component-wise) translation estimation can be solved exactly via an \emph{adaptive voting} scheme, 
while rotation estimation can be relaxed to a semidefinite program. 
%
The resulting \edit{polynomial-time approach}, \edit{named \name} (\emph{\nameLong}),  
outperforms \ransac and robust local optimization techniques, and  
 favorably compares with \edit{\bnb} methods.
 \name can tolerate 
up to $\maxOutliers$ outliers and returns highly-accurate solutions. 

\edit{While running in polynomial time, the general-purpose SDP solver used in our current implementation scales poorly in the problem size. Current research effort is devoted to developing specialized SDP solvers to allow using \name to solve large-scale registration
 problems in real-time.}

\section*{Acknowledgments}

\edit{This work was partially funded by ARL DCIST CRA W911NF-17-2-0181, ONR RAIDER
N00014-18-1-2828, and the Google Daydream Research Program.}





\renewcommand{\theequation}{A\arabic{equation}}
\renewcommand{\thetheorem}{A\arabic{theorem}}
\renewcommand{\thefigure}{A\arabic{figure}}
\renewcommand{\thetable}{A\arabic{table}}
\renewcommand{\theequation}{A\arabic{equation}}

\large
\begin{center}
{\bf Supplementary Material}
\end{center}

\normalsize

\section{Proof of~\prettyref{thm:TIM}}
\label{sec:proof:thm:TIM}

 Using the vector notation $\va \in \Real{3\nrPoints}$ and $\vb \in \Real{3\nrPoints}$ already introduced in the statement of the theorem, we can write the generative model~\eqref{eq:robustGenModel} compactly as:
 \bea
 \label{eq:robustGenModelVect}
 \vb = s (\eye_\nrPoints \kron \MR) \va + (\ones_\nrPoints \kron \vt) + \vo + \veps
 \eea
 where $\vo \doteq [\vo_1\tran \; \ldots \; \vo_\nrPoints\tran]\tran$,
  $\veps \doteq [\veps_1\tran \; \ldots \; \veps_\nrPoints\tran]\tran$, and
  $\ones_\nrPoints$ is a column vector of ones of size $\nrPoints$. Denote $K=|\calE|$ as the cadinality of $\calE$, such that $\MA \in \Real{K \times \nrPoints}$.
  Let us now multiply both members by $(\MA \kron \eye_3)$:
  \bea
  \label{eq:proof-tim1}
 \TIMb = 
  (\MA \kron \eye_3) [s (\eye_\nrPoints \kron \MR) \va +  (\ones_\nrPoints \kron \vt) + (\vo + \veps)]
 \eea
 Using the property of the Kronecker product we simplify:
 \beal
 (i) &  (\MA \kron \eye_3)(\eye_\nrPoints \kron \MR) \va = (\MA \kron \MR) \va \\
 &=  (\eye_K \kron \MR) (\MA \kron \eye_3) \va \\
 &=(\eye_K \kron \MR) \TIMa\\
 (ii)& (\MA \kron \eye_3)(\ones_\nrPoints \kron \vt) = (\MA \ones_\nrPoints \kron \vt ) = \zero
 \eeal
where we used the fact that $\ones_\nrPoints$ is in the Null space of the 
 incidence matrix $\MA$~\cite{Chung96book}. Using (i) and (ii), eq.~\eqref{eq:proof-tim1} becomes:
 \bea
  \label{eq:proof-tim2}
 \TIMb = s (\eye_K \kron \MR)  \TIMa + (\MA \kron \eye_3)(\vo + \veps)
 \eea
 which is invariant to the translation $\vt$, concluding the proof.  

\section{Summary of Invariant Measurements}
Table~\ref{tab:summaryIMs} below provides a summary of the invariant measurements from the main document.

\begin{table*}[t]
\centering
\begin{tabular}{cccc}
Measurements & Points & \TIMs & \TRIMs \\
\hline 
Symbol & $\va_i$, $\vb_i$ & $\barva_{ij}$, $\barvb_{ij}$ & $s_{ij}$ \\
\hline 
Definition & - & $\begin{cases} \barva_{ij} = \barva_j - \barva_i \\ \barvb_{ij} = \barvb_j - \barvb_i \end{cases}$ & $s_{ij} = \frac{\| \barvb_{ij} \| }{ \| \barva_{ij} \| }$ \\
\hline 
Generative model & $\vb_i = s\MR\va_i + \vt + \vo_i + \vepsilon_i $ & $\barvb_{ij} = s\MR \barva_{ij} + \vo_{ij} + \vepsilon_{ij}$ & $s_{ij} = s + o_{ij}^s + \epsilon_{ij}^s$ \\
\hline 
Noise bounds & $\|\vepsilon_i\| \leq \beta_i$ & $\|\vepsilon_{ij}\| \leq \beta_{ij} \doteq \beta_i+\beta_j$ & $|\epsilon_{ij}^s| \leq \alpha_{ij} \doteq \beta_{ij}/\|\barva_{ij}\|$ \\
\hline 
Dependent transformations & $(s,\MR,\vt)$ & $(s,\MR)$ & s \\
\hline
Number & $N$ & $K \leq \frac{N(N-1)}{2}$ & $K$ \\
\hline
\end{tabular}
\caption{Summary of invariant measurements.}
\label{tab:summaryIMs}
\vspace{-5mm}
\end{table*}

\section{Novelty of \TIMs and \TRIMs}
\label{sec:noveltyIMs}

\begin{remark}[Novelty]
We remark that the idea of using \emph{translation invariant measurements} and \emph{rotation invariant measurements} has been proposed in recent work~\cite{Li19arxiv-fastRegistration,Bustos18pami-GORE,Liu18eccv-registration,agarwal2017icra-RFM-SLAM} while (i) the graph theoretic interpretation of~\prettyref{thm:TIM}
is novel and generalizes previously proposed methods, and (ii) the notion of \emph{translation and rotation invariant measurements} (\TRIMs) is completely new. We also remark that while related work uses invariant measurements 
to filter-out outliers~\cite{Bustos18pami-GORE} or to speed up \bnb~\cite{Li19arxiv-fastRegistration,Liu18eccv-registration}, we show that they also allow computing a \emph{polynomial-time} robust registration solution. 
\end{remark}

\section{Proof of~\prettyref{thm:scalarTLS}}
\label{sec:proof:thm:scalarTLS}

Let us first prove that there are at most $2\nrTIM-1$ different non-empty consensus sets. 
We attach a confidence interval $[s_k - \alpha_k\barc, s_k + \alpha_k\barc]$ 
to each measurement $s_k$, $\forall k \in \{1,\ldots,\nrTIM\}$.
For a given scalar $s\in \Real{}$, a measurement $k$ is in the consensus set of $s$ if $s\in [s_k-\alpha_k\barc, s_k+\alpha_k\barc]$ (satisfies $\frac{\|s-s_k\|^2}{\alpha_k^2} \leq \barcsq$), see Fig.~\ref{fig:consensusMax}(a).
Therefore, the only points on the real line where the 
consensus set may change are the boundaries (shown in red in Fig.~\ref{fig:consensusMax}(a)) of the intervals 
$[s_k-\alpha_k\barc, s_k+\alpha_k\barc]$, $k \in \{1,\ldots,\nrTIM\}$. Since there are at most   $2\nrTIM-1$  such intervals, there are at most $2\nrTIM-1$ non-empty consensus sets (Fig.~\ref{fig:consensusMax}(b)), concluding the first part of the proof. 
The second part follows from the fact that the consensus set of $\hats$ is necessarily one of the $2\nrTIM-1$ possible consensus sets, and problem~\eqref{eq:TLSscale} simply computes the least squares estimate of the measurements in the consensus set of the solution and choose the estimate that induces the lowest cost as the optimal estimate. Therefore, we can just enumerate every possible consensus set and compute a least squares estimate for each of them, to find the solution that induces the smallest cost.

\section{Comments on TLS vs. Consensus Maximization}
\label{sec:exampleTLSMaxConsensus}

\TLS (and Algorithm~\ref{alg:adaptiveVoting}) are related to \emph{consensus maximization}, a popular approach for outlier detection in vision~\cite{Speciale17cvpr-consensusMaximization,Liu18eccv-registration}. 
Consensus maximization looks for an estimate that maximizes the number of inliers, i.e. $\max_s |\calC(s)|$, where $|\cdot|$ denotes the cardinality of a set.
While consensus maximization is intractable in general, by following the same lines of Theorem~\ref{thm:scalarTLS}, it is easy to show that consensus maximization can be solved in polynomial time in the scalar case as $\argmax_{i,\ldots,2K-1} |\calC_i|$. While we expect the \TLS solution to maximize the set of inliers, consensus maximization and \TLS will not return the same solution in general, since \TLS may prefer to discard measurements that induce a large bias in the estimate, as shown by the simple example below.

\begin{example}[\TLS vs. Consensus Maximization]
Consider a simple scalar estimation problem, where we are given three measurements $s_1\!=\!s_2\!=\!0$ 
and $s_3 = 3$. Assume $\alpha_k = 2$, $k=1,2,3$ and $\barc=1$. 
Then, it is possible to see that $s_{m} = 1.5$ attains a maximum consensus set including all measurements $\{1,2,3\}$, while the \TLS estimate is $\hats = 0$ which attains a cost $f_s(\hats) = 1$, and has consensus set $\calC(\hats) = \{1,2\}$.
\end{example}

\section{Proof of~\prettyref{thm:maxClique}}
\label{sec:proof:thm:maxClique}

Consider a graph $\calG'(\calV,\calE')$ whose edges where selected as inliers during scale estimation. 
An edge $(i,j)$ (and the corresponding \TIM) is an inlier if both $i$ and $j$ are correct correspondences (see discussion before~\prettyref{thm:TIM}). Therefore, $\calG'$ contains edges connecting all points for which we have inlier correspondences. 
Therefore, these points are vertices of a clique in the graph $\calG'$ 
and the edges (or equivalently the \TIMs) connecting those points form a clique in  $\calG'$.
We conclude the proof by observing 
that the clique formed by the inliers has to belong to at least one maximal clique of $\calG'$.

\section{Proof of~\prettyref{prop:cloning}}
\label{sec:proof:prop:cloning}

Here we prove the equivalence between Problem~\eqref{eq:TLSrotation2} and problem~\eqref{eq:TLSrotation4}.
Towards this goal, we show that the latter is simply a reparametrization of the former.

Let us rewrite~\eqref{eq:TLSrotation2} by making the orthogonality constraint $\MR \in \Othree$ explicit 
(recall $\Othree \doteq \setdef{\MR\in\Real{3 \times 3}}{\MR\tran \MR = \eye_3}$): 
\bea
\label{eq:TLSrotationA2}
\min_{
\substack{\MR,
\theta_k, \forall k }} 
& \sumAllIM 
 \frac{ (1 + \theta_k) }{ 2 } \frac{  \|  \TIMb_k -  \MR \TIMa_k \|^2 }{  \beta^2_k }  
+ 
\frac{ (1 - \theta_k) }{ 2 } 
\barcsq  \\
\subject 
&
\MR\tran\MR=\eye_3, \;\; \theta_k \in\{-1,+1\}, k=1,\ldots,\nrTIM \nonumber 
\eea
%

Noting that the term $\frac{(1 + \theta_k)}{2} \in \{0,1\}$, we can safely move it inside the squared norm and rewrite
eq.~\eqref{eq:TLSrotationA2} as:
\bea
\min_{
\substack{\MR, \theta_k, \forall k }} %
&
\sumAllIM 
\frac{  \| \TIMb - \MR \TIMa 
+ \theta_k \TIMb - \theta_k \MR \TIMa \|^2 }{ 4 \beta^2_k }  
+ 
\frac{ (1 - \theta_k) }{ 2 } 
\barcsq  \label{eq:TLSrotationA3}
\\
\subject 
&
\MR\tran\MR=\eye_3, \;\; \theta_k \in\{-1,+1\}, k=1,\ldots,\nrTIM \nonumber 
\eea
Now, we reparametrize the problem by introducing matrices $\MR_k \doteq \theta_k \MR$.
 In particular, we note that these matrices satisfy:
 \bit
 \item $\MR_k\tran \MR_k = (\theta_k\MR)\tran(\theta_k\MR) = \theta_k^2 \eye_3 = \eye_3$ (i.e., $\MR_k \in \Othree$)
 \item $\MR\tran \MR_k = \theta_k \cdot \eye_3 \in \{-\eye_3,+\eye
_3\}$, which also implies 
\item $\theta_k =\ve_1\tran \MR\tran \MR_k\ve_1$, where $\ve_1 \doteq [1 \; 0 \; 0]\tran$.
\eit
These three properties allow writing~\eqref{eq:TLSrotationA3} as:
\bea
\min_{
\substack{\MR, \MR_k, \forall k }} %
&
\sumAllIM 
\frac{  \| \TIMb - \MR \TIMa 
+ \MR\tran \MR_k  \TIMb - \MR_k \TIMa \|^2 }{ 4 \beta^2_k }  
+ 
\frac{ (1 - \ve_1\tran \MR\tran \MR_k \ve_1) }{ 2 } 
\barcsq  \nonumber 
\\
\subject 
\hspace{-5mm}&
\MR\tran\MR=\eye_3, \;\; \MR_k\tran\MR_k=\eye_3,
\nonumber 
 \\
& \MR\tran\MR_k \in\{-\eye,+\eye\},
\; k=1,\ldots,\nrTIM \label{eq:TLSrotationA4}
\eea
%
which matches~\eqref{eq:TLSrotation4}, concluding the proof.

%

\section{Proof of~\prettyref{prop:TLSrotationRelax}}
\label{sec:proof:prop:TLSrotationRelax}

Here we prove that problem~\eqref{eq:TLSrotationRelax} is a 
 convex relaxation of~\eqref{eq:TLSrotation4}, in the sense that the two problems have the same objective and the feasible set of~\eqref{eq:TLSrotationRelax} contains the feasible set of~\eqref{eq:TLSrotation4}.  
 We start by providing a matrix reparametrization of the problem, given in the following lemma.

\begin{lemma}[Matrix $\MX$ Formulation]
\label{lem:TLSrotation_X}
Problem~\eqref{eq:TLSrotation4}
is equivalent to the following optimization problem:

\vspace{-3mm}
\small
\bea
\hspace{-3mm}
\min_{
\substack{\MX}} %
& \hspace{-5mm} \trace{\barMQ \MX\tran\MX} \hspace{-3mm}\label{eq:TLSrotation_X}
\\
\hspace{-3mm}\subject \hspace{-3mm}
& \hspace{-5mm} [\MX\tran\MX]_{RR} =\eye_3, 
\;\;\; [\MX\tran\MX]_{R_k R_k} =\eye_3,
\;\;\; [\MX\tran\MX]_{II} =\eye_3,
\nonumber \hspace{-3mm}
 \\
& \hspace{-5mm} [\MX\tran\MX]_{RR_k} = (\ve_1\tran [\MX\tran\MX]_{RR_k} \ve_1) \eye_3, \;\; \forall k \nonumber \hspace{-3mm}\\
& \hspace{-5mm} \|  [\MX\tran\MX]_{IR} \pm [\MX\tran\MX]_{IR_k} \| \leq  1 \pm (\ve_1\tran [\MX\tran\MX]_{RR_k} \ve_1), \;\; \forall k \nonumber \hspace{-3mm}\\
& \hspace{-5mm} \|  [\MX\tran\MX]_{IR_k} \pm [\MX\tran\MX]_{IR_{k'}} \| \leq  1 \pm (\ve_1\tran [\MX\tran\MX]_{R_k R_{k'}} \ve_1), \;\; \forall k,k' \nonumber \hspace{-5mm}
\eea

\normalsize
\end{lemma}  

\myParagraph{Proof of~\prettyref{lem:TLSrotation_X}}
To prove the lemma, we reparametrize problem~\eqref{eq:TLSrotation4} using the matrix $\MX = [\eye_3 \; \MR \; \MR_1 \; \ldots \; \MR_\nrTIM]$. 

We first rewrite each summand in the objective of~\eqref{eq:TLSrotation4} as a \emph{quadratic} function of $\MX$:
\beal
\frac{  \| \TIMb_k - \MR \TIMa_k 
+ \MR\tran \MR_k  \TIMb_k - \MR_k \TIMa_k \|^2 }{ 4 \beta^2_k }  
+ 
\frac{ (1 - \ve_1\tran \MR\tran \MR_k \ve_1) }{ 2 } 
\barcsq = \nonumber \\
\expl{developing the squared norm} 
\frac{  \| \TIMb_k - \MR \TIMa_k \|^2
+  \| \MR\tran \MR_k  \TIMb_k - \MR_k \TIMa_k \|^2
 }{ 4 \beta^2_k }  
+  \nonumber \\
\frac{ 2\trace{ (\TIMb_k - \MR \TIMa_k)\tran  
(\MR\tran \MR_k  \TIMb_k - \MR_k \TIMa_k)}
 }{ 4 \beta^2_k } +
\frac{ (1 - \ve_1\tran \MR\tran \MR_k \ve_1) }{ 2 } 
\barcsq = \nonumber \\
\expl{leveraging that $\MR_k \in \{-\MR,+\MR\}$ and $\| \pm \vxx\| = \|\vxx\|$} 
\frac{  2\| \TIMb_k - \MR \TIMa_k \|^2
 }{ 4 \beta^2_k }  
+  \nonumber \\
\frac{ 2\trace{ (\TIMb_k - \MR \TIMa_k)\tran  
(\MR\tran \MR_k  \TIMb_k - \MR_k \TIMa_k) }
 }{ 4 \beta^2_k } +
\frac{ (1 - \ve_1\tran \MR\tran \MR_k \ve_1) }{ 2 } 
\barcsq = \nonumber \\
\expl{developing the first and the second summand} 
\frac{  \TIMb_k\tran \TIMb_k + \TIMa_k\tran \TIMa_k 
 }{ 2 \beta^2_k }  - \frac{ 2 \TIMb_k\tran \MR \TIMa_k 
 }{ 2 \beta^2_k }  
+  \nonumber \\
\frac{ 2\trace{ (\TIMb_k - \MR \TIMa_k)\tran  
(\MR\tran \MR_k  \TIMb_k - \MR_k \TIMa_k}
 }{ 4 \beta^2_k } +
\frac{ (1 - \ve_1\tran \MR\tran \MR_k \ve_1) }{ 2 } 
\barcsq = \nonumber \\
\expl{developing the product in the trace}
\frac{  \TIMb_k\tran \TIMb_k + \TIMa_k\tran \TIMa_k 
 }{ 2 \beta^2_k }  - \frac{ 2 \TIMb_k\tran \MR \TIMa_k 
 }{ 2 \beta^2_k }  +
\frac{ (1 - \ve_1\tran \MR\tran \MR_k \ve_1) }{ 2 } 
\barcsq
+  \nonumber \\
\frac{ \trace{ 
\TIMb_k\tran \MR\tran \MR_k  \TIMb_k 
- \TIMb_k\tran \MR_k  \TIMa_k
- \TIMa_k\tran \MR\tran\MR\tran\MR_k  \TIMb_k
+ \TIMa_k\tran \MR\tran \MR_k \TIMa_k
}
 }{ 2 \beta^2_k }  = \nonumber \\ 
\expl{noting that $\MR\tran\MR\tran\MR_k = \MR_k\tran$ and simplifying}
\frac{  \TIMb_k\tran \TIMb_k + \TIMa_k\tran \TIMa_k 
 }{ 2 \beta^2_k }  - \frac{ 2 \TIMb_k\tran \MR \TIMa_k 
 }{ 2 \beta^2_k }  +
\frac{ (1 - \ve_1\tran \MR\tran \MR_k \ve_1) }{ 2 } 
\barcsq
+  \nonumber \\
\frac{ \trace{ 
\TIMb_k\tran \MR\tran \MR_k  \TIMb_k 
- 2\TIMb_k\tran \MR_k  \TIMa_k
+ \TIMa_k\tran \MR\tran \MR_k \TIMa_k
}
 }{ 2 \beta^2_k }  = \nonumber \\ 
\eeal
\beal
 \expl{rearranging terms and noting that for $x \in \Real{},\trace{x} = x$}
 \frac{  \TIMb_k\tran \TIMb_k + \TIMa_k\tran \TIMa_k + \beta_k^2\barcsq 
 }{ 2 \beta^2_k }  +
\frac{ (- \ve_1\tran \MR\tran \MR_k \ve_1) }{ 2 } 
\barcsq
+  \nonumber \\
\frac{ \trace{ 
\TIMb_k\tran \MR\tran \MR_k  \TIMb_k 
- 2\TIMb_k\tran \MR_k  \TIMa_k
- 2 \TIMb_k\tran \MR \TIMa_k 
+ \TIMa_k\tran \MR\tran \MR_k \TIMa_k
}
 }{ 2 \beta^2_k }  = \nonumber \\ 
 \expl{noting that $x \in \Real{}, \trace{x \eye_3} = 3 x$}
 \trace{ 
 \frac{  \TIMb_k\tran \TIMb_k + \TIMa_k\tran \TIMa_k + \beta_k^2\barcsq 
 }{ 6 \beta^2_k }   \eye_3} 
 +
 \trace{ 
- \frac{ \barcsq }{ 6 } 
\MR\tran \MR_k 
}
+  \nonumber \\
\frac{ \trace{ 
\TIMb_k\tran \MR\tran \MR_k  \TIMb_k 
- 2\TIMb_k\tran \MR_k  \TIMa_k
- 2 \TIMb_k\tran \MR \TIMa_k 
+ \TIMa_k\tran \MR\tran \MR_k \TIMa_k
}
 }{ 2 \beta^2_k }  = \nonumber \\
 \expl{using linearity and the cyclic property of the trace}
 \trace{ 
 \frac{  \TIMb_k\tran \TIMb_k + \TIMa_k\tran \TIMa_k + \beta_k^2\barcsq 
 }{ 6 \beta^2_k }   \eye_3} 
 +
 \trace{ 
- \frac{ \barcsq }{ 6 } 
\MR\tran \MR_k 
}
+  \nonumber \\
\trace{ 
\frac{1}{2\beta_k^2}
\TIMb_k \TIMb_k\tran \MR\tran \MR_k 
} 
-
\trace{ 
\frac{1}{\beta_k^2}
 \TIMa_k \TIMb_k\tran \MR_k  
} 
\nonumber \\
-
\trace{ 
\frac{1}{\beta_k^2}
 \TIMa_k  \TIMb_k\tran \MR 
} 
+
\trace{ 
\frac{1}{2\beta_k^2}
\TIMa_k \TIMa_k\tran \MR\tran \MR_k 
}  \nonumber \\
 \expl{grouping terms including $\MR\tran \MR_k$ and using $\trace{x \eye_3} = 3 x$}
 \trace{ 
 \frac{  \TIMb_k\tran \TIMb_k + \TIMa_k\tran \TIMa_k + \beta_k^2\barcsq 
 }{ 6 \beta^2_k }   \eye_3} 
  \nonumber \\
+ \trace{ 
\frac{  \TIMb_k\tran \TIMb_k + \TIMa_k\tran \TIMa_k -
 \beta_k^2\barcsq 
 }{ 6 \beta^2_k }
\MR\tran \MR_k 
} 
\nonumber\\
-
\trace{ 
\frac{1}{\beta_k^2}
 \TIMa_k \TIMb_k\tran \MR_k  
} 
-
\trace{ 
\frac{1}{\beta_k^2}
 \TIMa_k  \TIMb_k\tran \MR 
}  \nonumber \\
\expl{noting that the expression is linear in $\MX\tran\MX$, \cf eq.~\eqref{eq:Z}}
\doteq \barMQ_k \MX\tran\MX  
\eeal
where $\barMQ_k$ is a symmetric matrix with $3\times3$ blocks $[\barMQ_k]_{UV}$ (below we use the block indices from eq.~\eqref{eq:Z}) defined as:
\beq
\; [\barMQ_k]_{UV} = 
\left\{
\begin{array}{ll}
\frac{\TIMb_k\tran \TIMb_k + \TIMa_k\tran \TIMa_k + \beta_k^2\barcsq 
 }{ 6 \beta^2_k } \eye_3  & \substack{\text{if } U=I \text{ and } V = I}
 \vspace{2mm}
 \\
 \frac{  \TIMb_k\tran \TIMb_k + \TIMa_k\tran \TIMa_k - \beta_k^2\barcsq 
 }{ 12 \beta^2_k } \eye_3 & 
 \substack{\text{if }U=R \text{ and } V = R_k 
 \\ \text{or } U=R_k \text{ and } V = R}
\vspace{2mm}
 \\
 -( \frac{1}{2\beta_k^2}
 \TIMa_k \TIMb_k\tran  )\tran 
   & 
    \substack{
    \text{if }U=I \text{ and } V = R 
 \\ \text{or } U=I \text{ and } V = R_k}
\vspace{2mm}
 \\
 -( \frac{1}{2\beta_k^2}
 \TIMa_k \TIMb_k\tran  )
   & 
    \substack{
    \text{if }U=R \text{ and } V = I
 \\ \text{or } U=R_k \text{ and } V = I}
 \end{array}  
\right.
\eeq
Since each summand in the objective of~\eqref{eq:TLSrotation4} can be written as $\barMQ_k \MX\tran\MX$, the linearity of the trace allows writing the overall objective  as $\barMQ \MX\tran\MX$, where:
\beq
\barMQ = \sumAllIM  \barMQ_k
\eeq
concluding the first part of the proof.

To finish the proof of the equivalence between~\eqref{eq:TLSrotation_X} and~\eqref{eq:TLSrotation4}, we show that the constraints in~\eqref{eq:TLSrotation4} can be written as in~\eqref{eq:TLSrotation_X}, while the latter also includes extra \emph{redundant} constraints. 
From the structure of $\MX\tran\MX$ in eq.~\eqref{eq:Z}, it is trivial to see that the orthogonality constraints on $\MR$ and $\MR_k$ are equivalent to the first three constraints (first line) in~\eqref{eq:TLSrotation_X}, where we also added a constraint $[\MX\tran\MX]_{II} =\eye_3$ on the top-left identity block of $\MX\tran\MX$, \cf eq.~\eqref{eq:Z}.
Now we note that the constraint $\MR\tran\MR_k \in \{ -\eye_3, +\eye_3\}$ is equivalent to: 
imposing that 
(i)
$\MR\tran\MR_k \in \Othree$ and (ii) $\MR\tran\MR_k = x \eye_3$, for some scalar $x$ (in words, $\MR\tran\MR_k$ is a multiple of the identity matrix).  
This can be easily seen by noting that the two conditions imply:
\bea
(\MR\tran\MR_k)\tran \MR\tran\MR_k \overset{(i)}{=}  \eye_3 
\quad
\overset{(ii)}{\Rightarrow}
\quad
x^2 \eye_3 = \eye_3
\eea
which is only true for $x \in \{-1,+1\}$. 
Since we already imposed that $\MR$ and $\MR_k$ are orthogonal, then also $\MR\tran\MR_k$ must be orthogonal,
and we are only left to add the constraint $\MR\tran\MR_k = x \eye_3$. To avoid adding extra variables, we rewrite the constraint $\MR\tran\MR_k = x \eye_3$ as:
\bea
\label{eq:idiag11}
[\MX\tran\MX]_{RR_k} = (\ve_1\tran [\MX\tran\MX]_{RR_k} \ve_1) \eye_3
\eea
where $\ve_1\tran [\MX\tran\MX]_{RR_k} \ve_1$ simply selects the top-left entry of $[\MX\tran\MX]_{RR_k}$.
Eq.~\eqref{eq:idiag11} corresponds to the second line in~\eqref{eq:TLSrotation_X}.

Finally, we note that the last two constraints in~\eqref{eq:TLSrotation_X} are \emph{redundant}, i.e., they are trivially satisfied given the other constraints. To prove this fact we observe:
\bea
\|  [\MX\tran\MX]_{IR} \pm [\MX\tran\MX]_{IR_k} \| \leq  1 \pm (\ve_1\tran [\MX\tran\MX]_{RR_k} \ve_1)
& \iff \nonumber\\
\expl{using~\eqref{eq:Z}}
\| \MR \pm \MR_k \| \leq 1 \pm (\ve_1\tran \MR \MR_k \ve_1)
& \iff \nonumber\\
\expl{recalling $\MR_k = \theta \MR$, for some $\theta \in
\{-1,+1\}$}
\| (1 \pm \theta) \MR \| \leq 1 \pm \theta
&
\iff \nonumber\\
\expl{since $\|\MR\|=1$ and $(1\pm\theta)\geq0$ for $\theta \in
\{-1,+1\}$}
1 \pm \theta \leq 1 \pm \theta
\eea
which is trivially satisfied (a proof of the redundancy of the last constraint proceeds in a similar manner). 
\qed

\begin{remark}[Redundant constraints]
The proof of~\prettyref{lem:TLSrotation_X} shows that the last two constraints in~\eqref{eq:TLSrotation_X} are \emph{redundant}, i.e., they can be omitted from~\eqref{eq:TLSrotation_X} without altering the result.
 While these redundant constraints do not play any role in~\eqref{eq:TLSrotation_X}, empirical evidence shows that they largely improve the quality of our convex relaxation (presented below), and this is the reason why we added them to~\eqref{eq:TLSrotation_X} in the first place.
\end{remark}

We are now ready to prove~\prettyref{prop:TLSrotationRelax}.

\myParagraph{Proof of~\prettyref{prop:TLSrotationRelax}}
Using~\prettyref{lem:TLSrotation_X}, it is now trivial to show that~\eqref{eq:TLSrotationRelax} is a 
 convex relaxation of~\eqref{eq:TLSrotation4}.
 Towards this goal, we first observe that from~\prettyref{lem:TLSrotation_X}, problem~\eqref{eq:TLSrotation4} is equivalent to~\eqref{eq:TLSrotation_X}.
 Second, we observe that~\eqref{eq:TLSrotation_X} can be rewritten using a matrix $\MZ = \MX\tran\MX$, which is a positive-semidefinite rank-3 matrix:

 \vspace{-3mm}
\small
\bea
\hspace{-3mm}
\min_{
\substack{\MZ}} %
& \hspace{-5mm} \trace{\barMQ \MZ} \hspace{-3mm}\label{eq:TLSrotation_ZZ}
\\
\hspace{-3mm}\subject \hspace{-3mm}
& \hspace{-5mm} [\MZ]_{RR} =\eye_3, 
\;\;\; [\MZ]_{R_k R_k} =\eye_3,
\;\;\; [\MZ]_{II} =\eye_3,
\nonumber \hspace{-3mm}
 \\
& \hspace{-5mm} [\MZ]_{RR_k} = (\ve_1\tran [\MZ]_{RR_k} \ve_1) \eye_3, \;\; \forall k \nonumber \hspace{-3mm}\\
& \hspace{-5mm} \|  [\MZ]_{IR} - [\MZ]_{IR_k} \| \leq  1 - (\ve_1\tran [\MZ]_{RR_k} \ve_1), \;\; \forall k \nonumber \hspace{-3mm}\\
& \hspace{-5mm} \|  [\MZ]_{IR_k} - [\MZ]_{IR_{k'}} \| \leq  1 - (\ve_1\tran [\MZ]_{R_k R_{k'}} \ve_1), \;\; \forall k,k' \nonumber \hspace{-3mm} \\
& \MZ \succeq 0, \quad \rank{\MZ} = 3
\eea

\normalsize
All the constraints in~\eqref{eq:TLSrotation_ZZ} are convex except the rank constraint. Therefore, we can obtain a convex relaxation by dropping the rank constraint from~\eqref{eq:TLSrotation_ZZ} and obtain~\eqref{eq:TLSrotationRelax}. 
\qed

\section{Proof of~\prettyref{prop:TLSrotation}}
\label{sec:proof:prop:TLSrotation}

Here we state and prove an extended version of~\prettyref{prop:TLSrotation}, which we omitted from the paper for space reasons.

\begin{proposition}[Guarantees for \TLS Rotation Estimation]\label{prop:TLSrotation-others}
Let $\MZ^\star$ be the optimal solution of the relaxation~\eqref{eq:TLSrotationRelax}.
The following two facts hold true:
\begin{enumerate}
\item 
If $\MZ^\star$ has rank 3, then it can be factored as $\MZ^\star = (\MX^\star)\tran(\MX^\star)$ and the 
first block row of $\MX^\star \doteq [\eye_3, {\MR}^\star, {\MR}^\star_1,\ldots, {\MR}^\star_\nrTIM]$ is an optimal solution for problem~\eqref{eq:TLSrotation4}.
\item If we call $\hat{\MR}, \hat{\MR}_1,\ldots, \hat{\MR}_\nrTIM$ the \emph{rounded estimate}, where $\hat{\MR} = \text{proj}_{\Othree}( [\MZ^\star]_{IR} )$ ($\text{proj}_S$ is the projection onto a set $S$), and 
$\hat{\MR}_k = \text{proj}_{\{-\hat{\MR},+\hat{\MR}\}} ([\MZ^\star]_{IR_k})$, then the rounded estimate is \emph{feasible} for the original problem~\eqref{eq:TLSrotation4}, and the suboptimality of the rounded estimate is bounded by: 
\beq
\hat{f}_R - f_R^\star \leq \hat{f}_R - f_R^\circ
\eeq
where  
$f_R^\star$ is the optimal objective of~\eqref{eq:TLSrotation4},
  $\hat{f}_R$ is the objective attained in~\eqref{eq:TLSrotation4} by the rounded estimate, and $f_R^\circ$ is the optimal objective of the convex relaxation~\eqref{eq:TLSrotationRelax}.
\end{enumerate}
\end{proposition}

\myParagraph{Proof} 
We first observe that $f_R^\circ \leq f_R^\star$: by definition, a relaxation can only attain a better objective with respect to the original problem, since it operates on a larger feasible set.

Now we prove the first claim in~\prettyref{prop:TLSrotation-others}.
The proof easily follows from~\prettyref{lem:TLSrotation_X} and the proof of~\prettyref{prop:TLSrotationRelax}.
In particular, in the proof of~\prettyref{prop:TLSrotationRelax}, we showed that 
(i) problem~\eqref{eq:TLSrotation4} is equivalent to problem~\eqref{eq:TLSrotation_ZZ}, and 
(ii) \eqref{eq:TLSrotation_ZZ} is the same as the convex relaxation~\eqref{eq:TLSrotationRelax}, except from the fact that in the relaxation we dropped the rank constraint. 
Therefore, if the relaxation~\eqref{eq:TLSrotationRelax} produces a rank-3 solution, such solution is also optimal for~\eqref{eq:TLSrotation_ZZ} (such solution attains a cost $f_R^\circ \leq f_R^\star$, hence when it is feasible, it must be optimal).
Moreover, since such solution is rank 3, it can be factored as $\MZ^\star = (\MX^\star)\tran(\MX^\star)$ 
and $\MX^\star$ solves~\eqref{eq:TLSrotation_X}.
Since~\eqref{eq:TLSrotation_X} is a reparametrization of the original problem~\eqref{eq:TLSrotation4}, we can always build an optimal solution to~\eqref{eq:TLSrotation4} using the blocks of $\MX^\star$, concluding the proof of the first claim.

The second claim follows from the fact that 
(i) the rounded estimate satisfies the constraints in~\eqref{eq:TLSrotation4}, hence must be feasible,
and 
(ii)  $f_R^\circ \leq f_R^\star$ (already observed earlier in the proof).
 Using (ii) we obtain:
 \bea
f_R^\circ \leq f_R^\star
\Rightarrow
 - f_R^\star \leq -f_R^\circ
 \nonumber\\
 \expl{adding $\hat{f}_R$ to both members}
 \Rightarrow
 \hat{f}_R- f_R^\star \leq \hat{f}_R-f_R^\circ
 \eea
 which concludes the proof.

 \begin{remark}[Rounding]
 The projections required to compute the rounded solution can be computed efficiently. In particular,
 for any matrix $\MM \in \Real{3\times3}$, $\text{proj}_{\Othree}(\MM)$ can be computed via SVD~\cite{Hartley13ijcv}, while $\text{proj}_{\{-\hat{\MR},+\hat{\MR}\}}(\MM)$ can be computed by inspection since the set $\{-\hat{\MR},+\hat{\MR}\}$ only contains two elements.
 \end{remark}

\section{Robust Translation Estimation}
\label{sec:translationEstimationSupp}

After obtaining an estimate of the scale $\hats$ and the rotation $\hatMR$, we can substitute them into the original formulation~\eqref{eq:TLSRegistration} and solve the resulting optimization problem to compute an estimate of the translation $\vt$. 
Since we already presented a poly-time solution for scalar \TLS in \prettyref{sec:scaleEstimation}, we propose to solve 
for the translation component-wise, i.e., 
%
we compute the entries $t_1, t_2, t_3$ of $\vt$ independently as the minimizers of: 
 \bea
 \label{eq:TLSRegistrationT1}
 \min_{ \substack{ t_j } } 
 \sumAllPointsi \min \left( \frac{1}{\beta_i^2} \left|  [\vb_i - \hats \hatMR \va_i]_j - t_j \right|^2 \!\!,  \barcsq \right)
 \eea
 where $[\cdot]_j$, $j=1,2,3$, denotes the $j$-th entry of a vector. 
While the solution to~\eqref{eq:TLSRegistrationT1} will not match in general the solution to~\eqref{eq:TLSRegistration}, 
problem~\eqref{eq:TLSRegistrationT1} 
also has the effect
to bound the maximum admissible error for each inlier measurement 
to $\beta_i$, but while~\eqref{eq:TLSRegistration} operates on the $\ell_2$ norm of the vector, operates component-wise~\eqref{eq:TLSRegistrationT1} (in other words, while the error in our original noise model was confined within the $\ell_2$-norm ball,
it has been now relaxed to the $\ell_\infty$-norm ball).


\renewcommand{\mpwthree}{6cm}
\renewcommand{\myhspace}{\hspace{-3mm}}

\begin{figure*}[t]
	\begin{center}
	\begin{minipage}{\textwidth}
	\hspace{-0.2cm}
	\begin{tabular}{ccc}%
			\begin{minipage}{\mpwthree}%
			\centering%
			\includegraphics[width=\columnwidth]{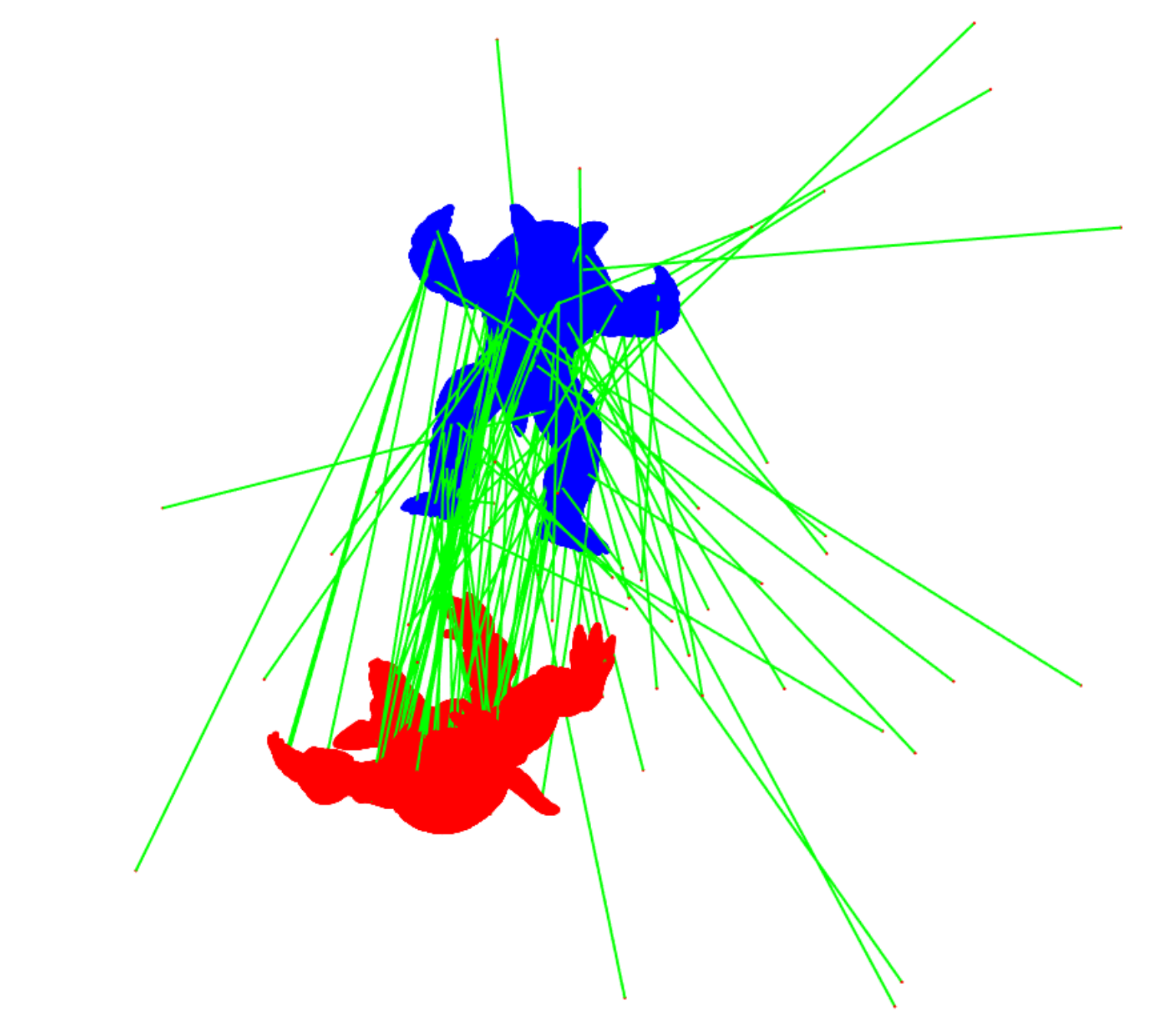} \\
			\end{minipage}
		& \myhspace
			\begin{minipage}{\mpwthree}%
			\centering%
			\includegraphics[width=\columnwidth]{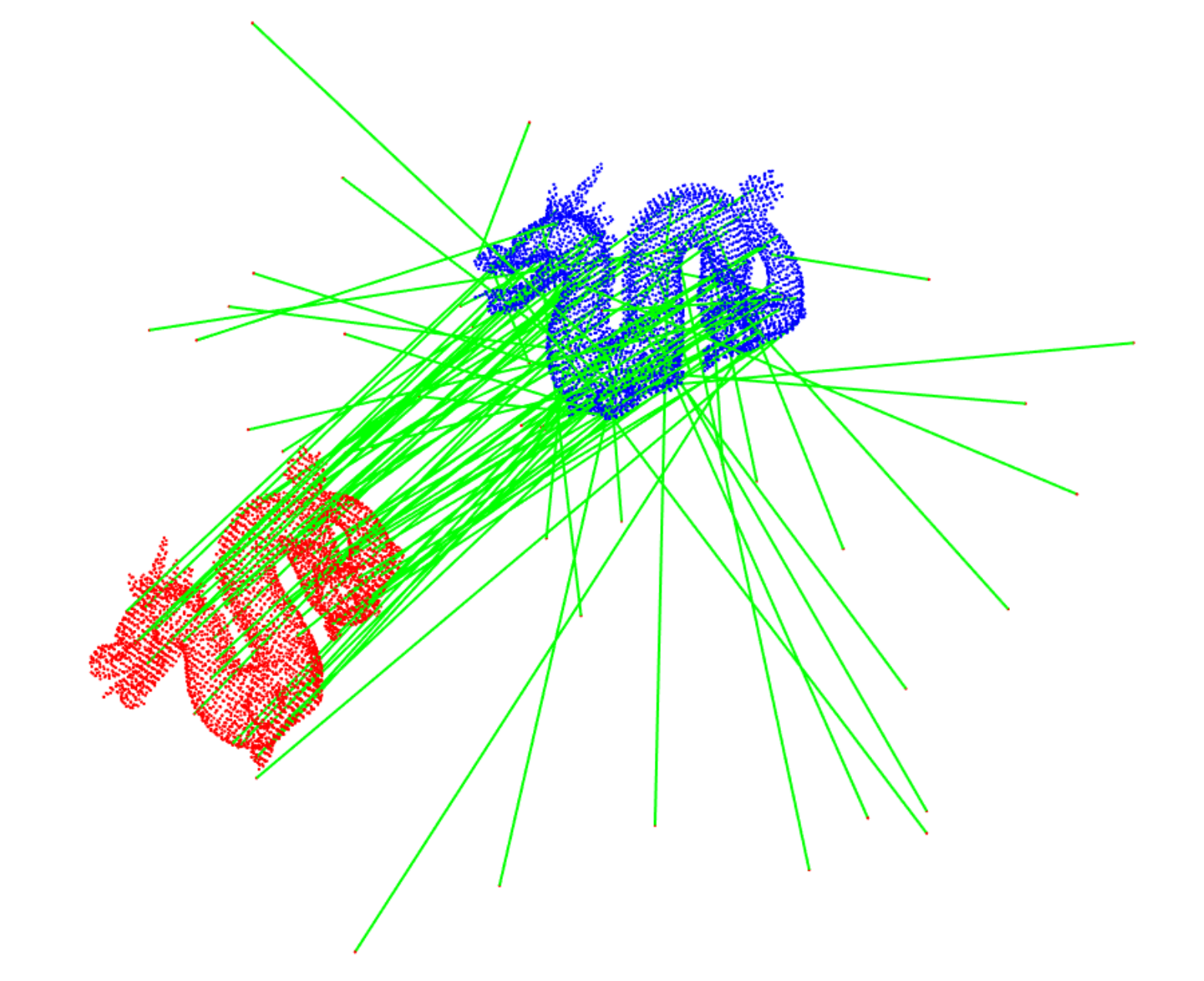} \\
			\end{minipage}
		& \myhspace
			\begin{minipage}{\mpwthree}%
			\centering%
			\includegraphics[width=\columnwidth]{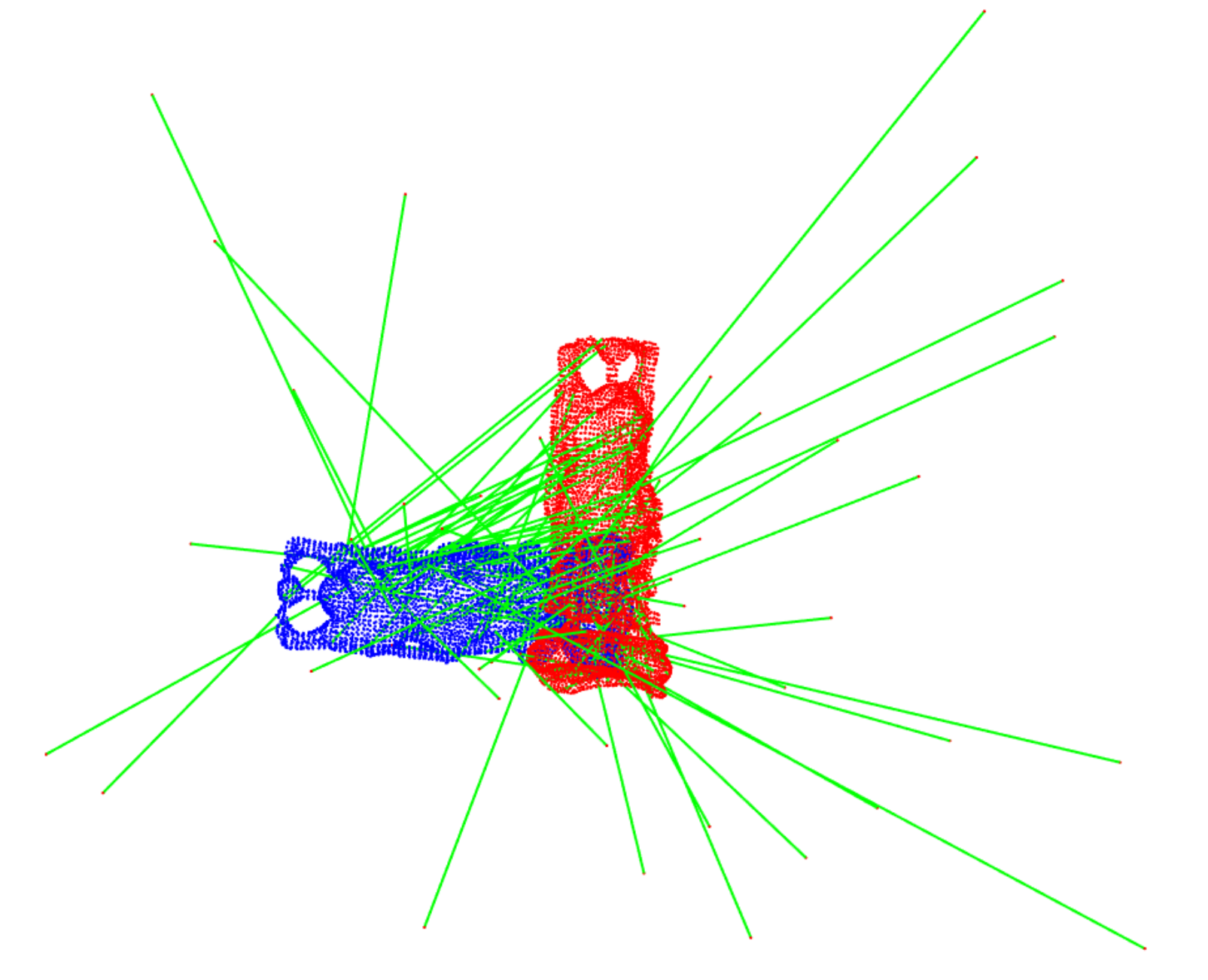} \\
			\end{minipage}  \\
			
		\begin{minipage}{\mpwthree}%
			\centering%
			\includegraphics[width=\columnwidth]{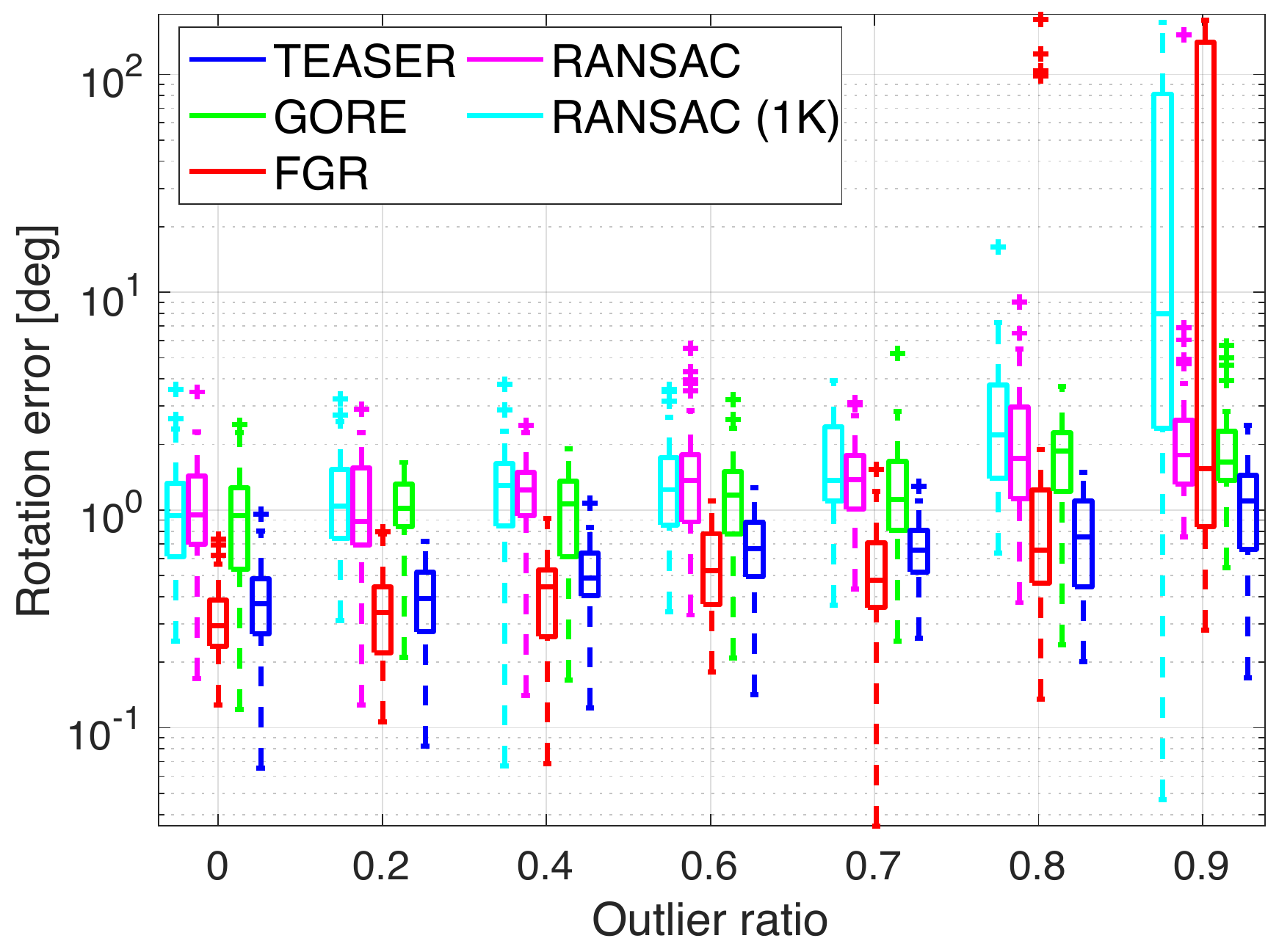} \\
			\end{minipage}
		& \myhspace
			\begin{minipage}{\mpwthree}%
			\centering%
			\includegraphics[width=\columnwidth]{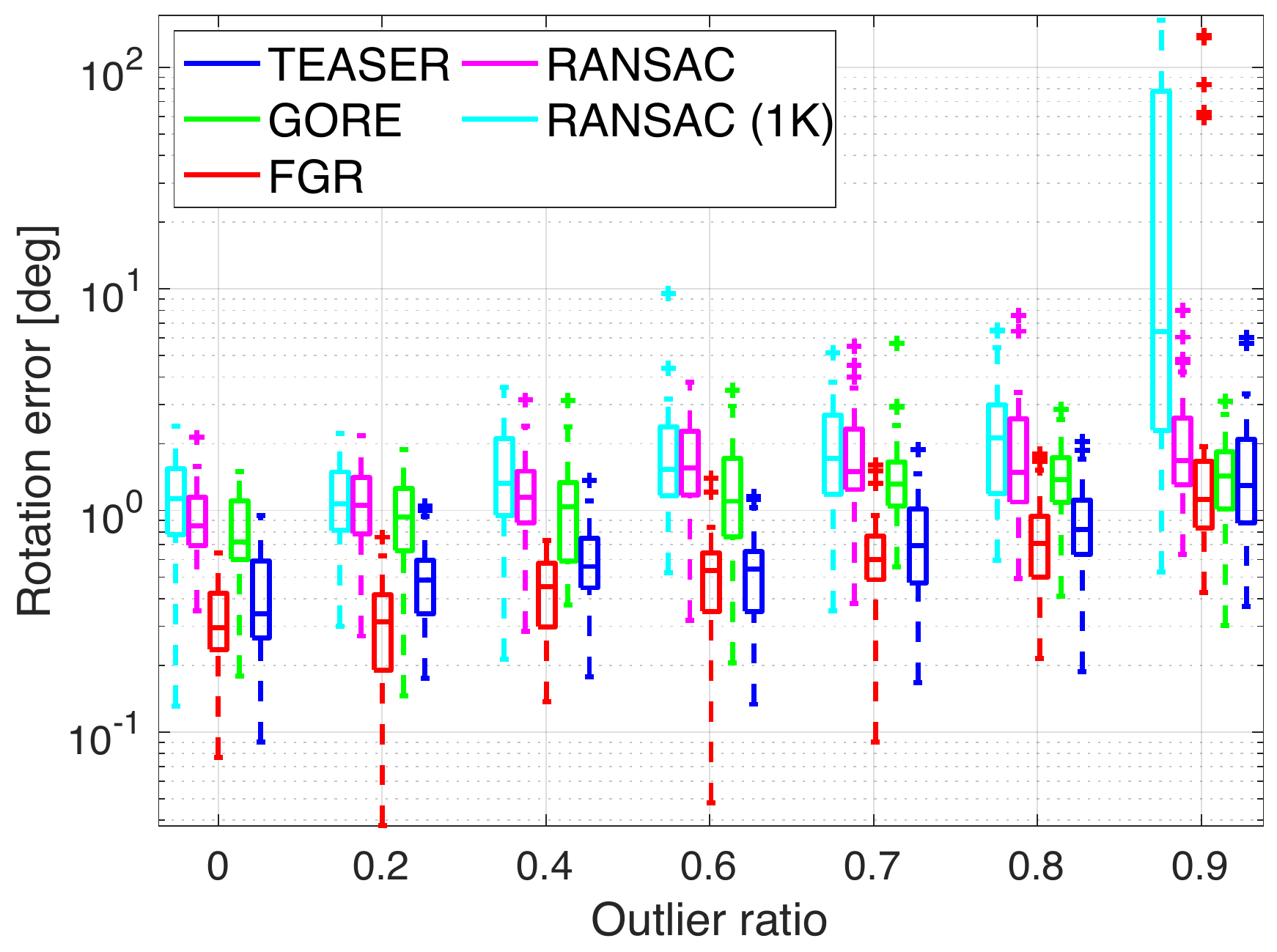} \\
			\end{minipage}
		& \myhspace
			\begin{minipage}{\mpwthree}%
			\centering%
			\includegraphics[width=\columnwidth]{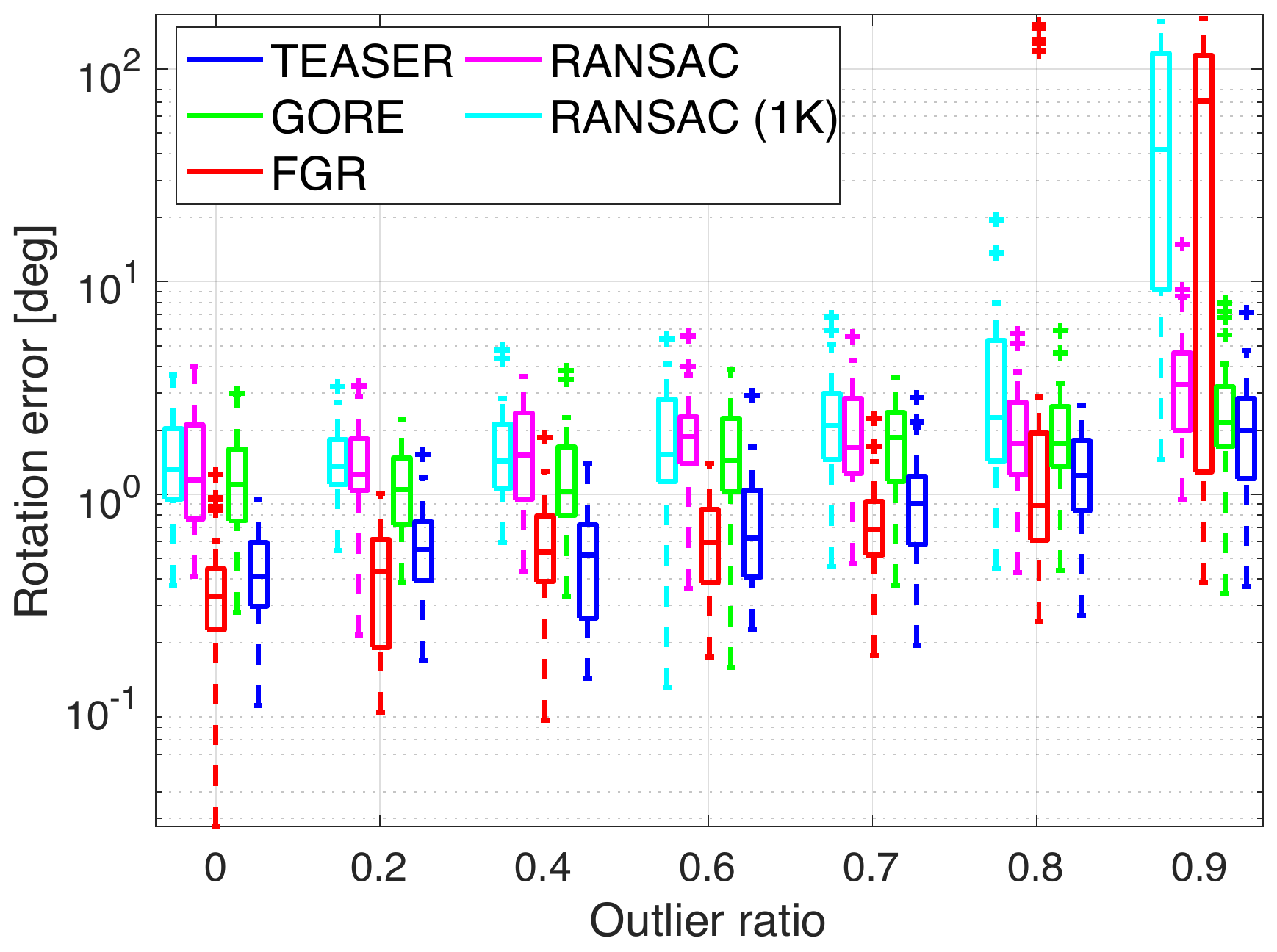} \\
			\end{minipage}\\
			
		\begin{minipage}{\mpwthree}%
			\centering%
			\includegraphics[width=\columnwidth]{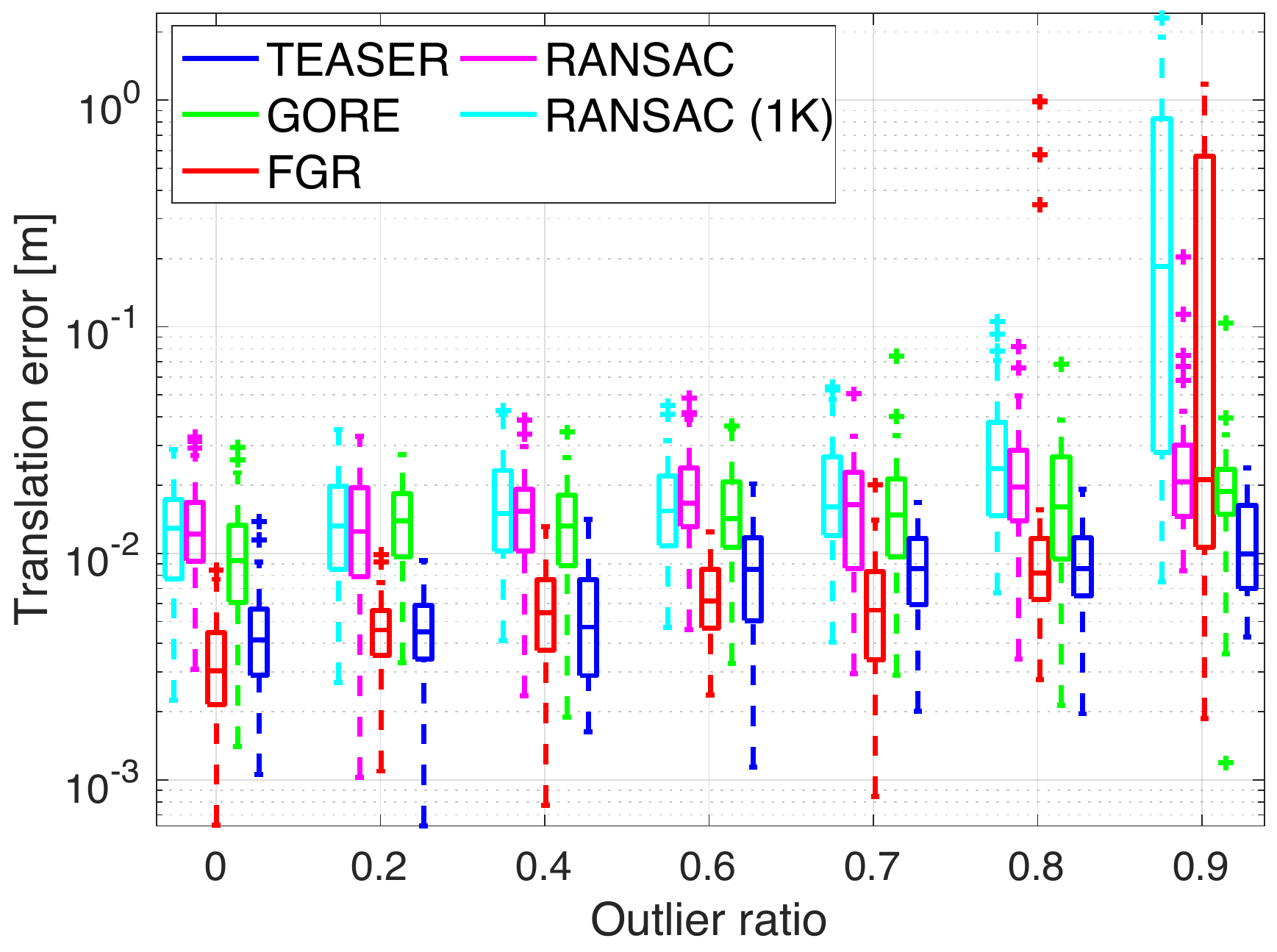} \\
			(a) \armadillo
			\end{minipage}
		& \myhspace
			\begin{minipage}{\mpwthree}%
			\centering%
			\includegraphics[width=\columnwidth]{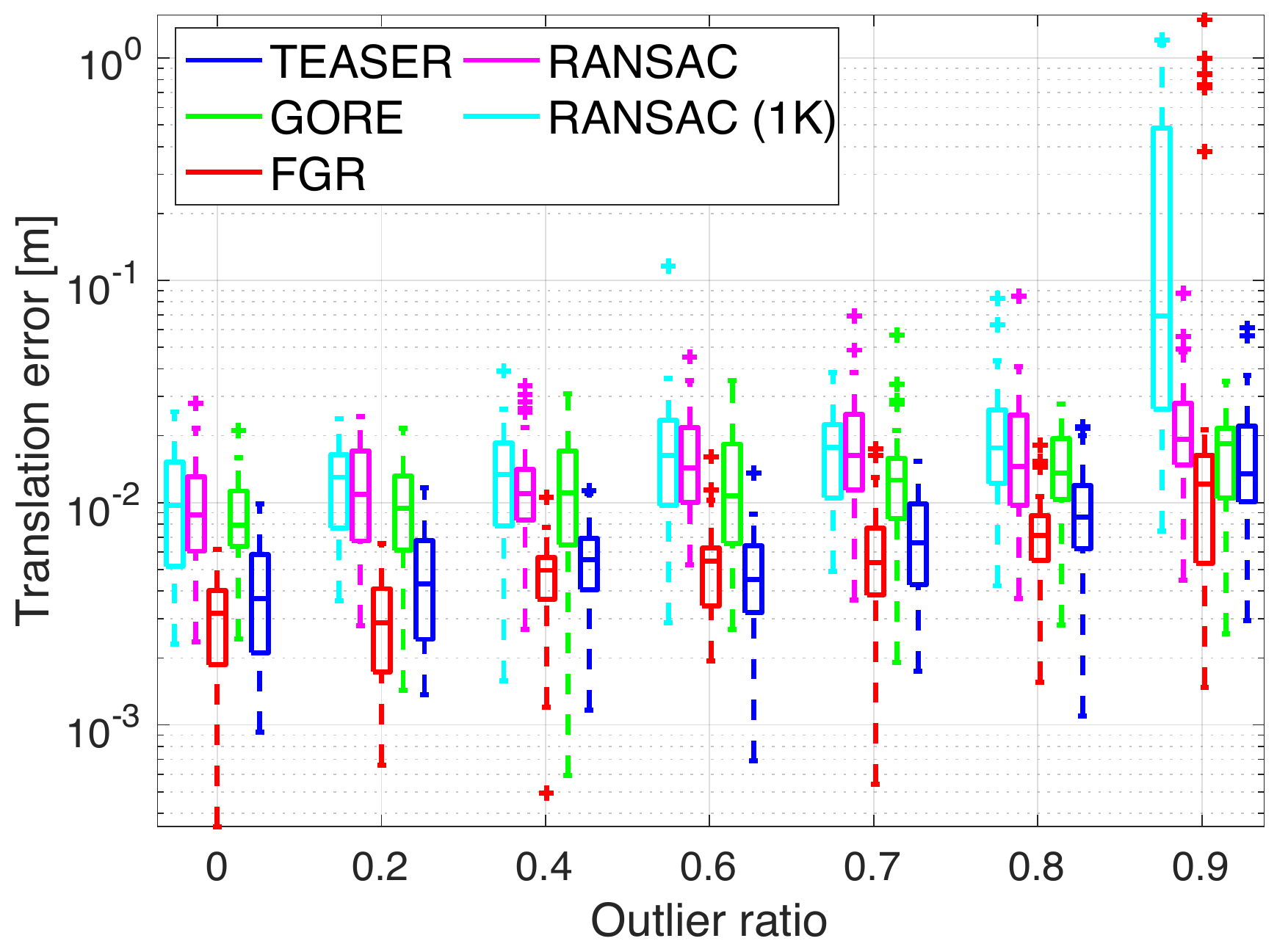} \\
			(b) \dragon
			\end{minipage}
		& \myhspace
			\begin{minipage}{\mpwthree}%
			\centering%
			\includegraphics[width=\columnwidth]{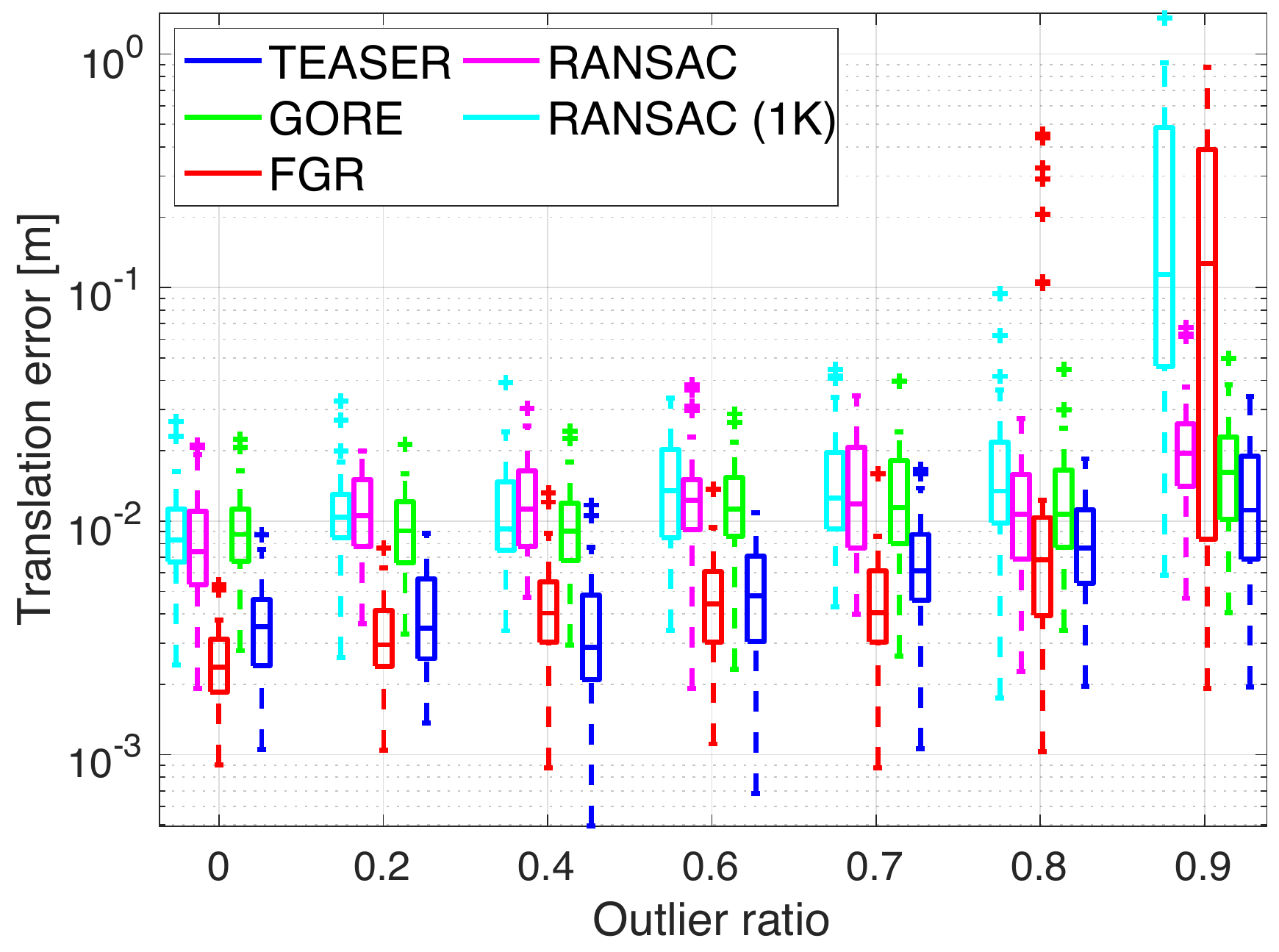} \\
			(c) \buddha
			\end{minipage}
		\end{tabular}
	\end{minipage}
	\vspace{-3mm} 
	\caption{Results for the \armadillo, \dragon and \buddha datasets at increasing levels of outliers. First row: example of putative correspondences with 50\% outliers. Blue points are the model point cloud, red points are the transformed scene model. Second row: rotation error produced by five methods including \name. Third row: translation error produced by the five methods. }
	 \label{fig:benchmarkSupp}
	\vspace{-8mm} 
	\end{center}
\end{figure*}

\section{Experiments and Applications}
\label{sec:experiments_supp}
In all the following experiments, $\bar{c}^2=1$ for scale, rotation and translation estimation.

\subsection{Benchmark on Synthetic Datasets}
\label{sec:benchmark_supp}
As stated in \prettyref{sec:benchmark} in the main document, we have benchmarked \name against four state-of-the-art methods in point cloud registration (\FGR~\cite{Zhou16eccv-fastGlobalRegistration}, \GORE~\cite{Bustos18pami-GORE}, \ransaconek, \ransac) at increasing level of outliers, using four datasets from the Stanford 3D Scanning Repository~\cite{Curless96siggraph}: \bunny, \armadillo, \dragon and \buddha. Results for the \bunny are showed in the main document and here we show the results for the other three datasets in Fig.~\ref{fig:benchmarkSupp}.

\subsection{\name on High Noise}
\edit{In the main document, the standard deviation $\sigma$ of the isotropic Gaussian noise is set to be $0.01$, with the point cloud scaled inside a unit cube $[0,1]^3$. In this section, we give a visual illustration of how corrupted the point cloud is after adding noise of such magnitude. In addition, we increase the noise standard deviation $\sigma$ to be $0.1$, visualize the corresponding corrupted point cloud and show that \name can still recover the ground-truth transformation with reasonable accuracy. Fig.~\ref{fig:bunny_with_noise} illustrates a \emph{clean} (noise-free) \bunny model scaled inside the unit cube and its variants by adding different levels of isotropic Gaussian noise and outliers. From Fig.~\ref{fig:bunny_with_noise}(b)(c), we can see $\sigma=0.01$ 
(noise used in the main document) is a reasonable noise standard deviation for real-world applications, while $\sigma=0.1$ destroys the geometric structure of the \bunny and it is beyond the noise typically encountered in robotics and computer vision applications. 

Fig.~\ref{fig:bunny_with_noise}(d) shows the \bunny with high noise ($\sigma = 0.1$) and 50\% outliers. Under this setup, we run \name to register the two point clouds and recover the relative transformation. \name can still return rotation and translation estimates that are  close to the ground-truth transformation. However, due to the severe noise corruption, even a successful registration 
fails to yield a visually convincing registration result to human perception. For example, Fig.~\ref{fig:bunny_teaser_high_noise} shows a representative \name registration result with $\sigma=0.1$ noise corruption and $50\%$ outlier rate, where accurate transformation estimates are obtained (rotation error is $3.42^{\circ}$ and translation error is $0.098$m), but the \bunny in the cluttered scene is hardly visible, even when super-imposed to the clean model.
}


\renewcommand{\myhspace}{\hspace{-6mm}}
\renewcommand{\mpw}{4.5cm}

\begin{figure}[h]
	\begin{center}
	\begin{minipage}{\textwidth}
	\begin{tabular}{cc}%
	\myhspace
			\begin{minipage}{\mpw}%
			\centering%
			\includegraphics[width=\columnwidth]{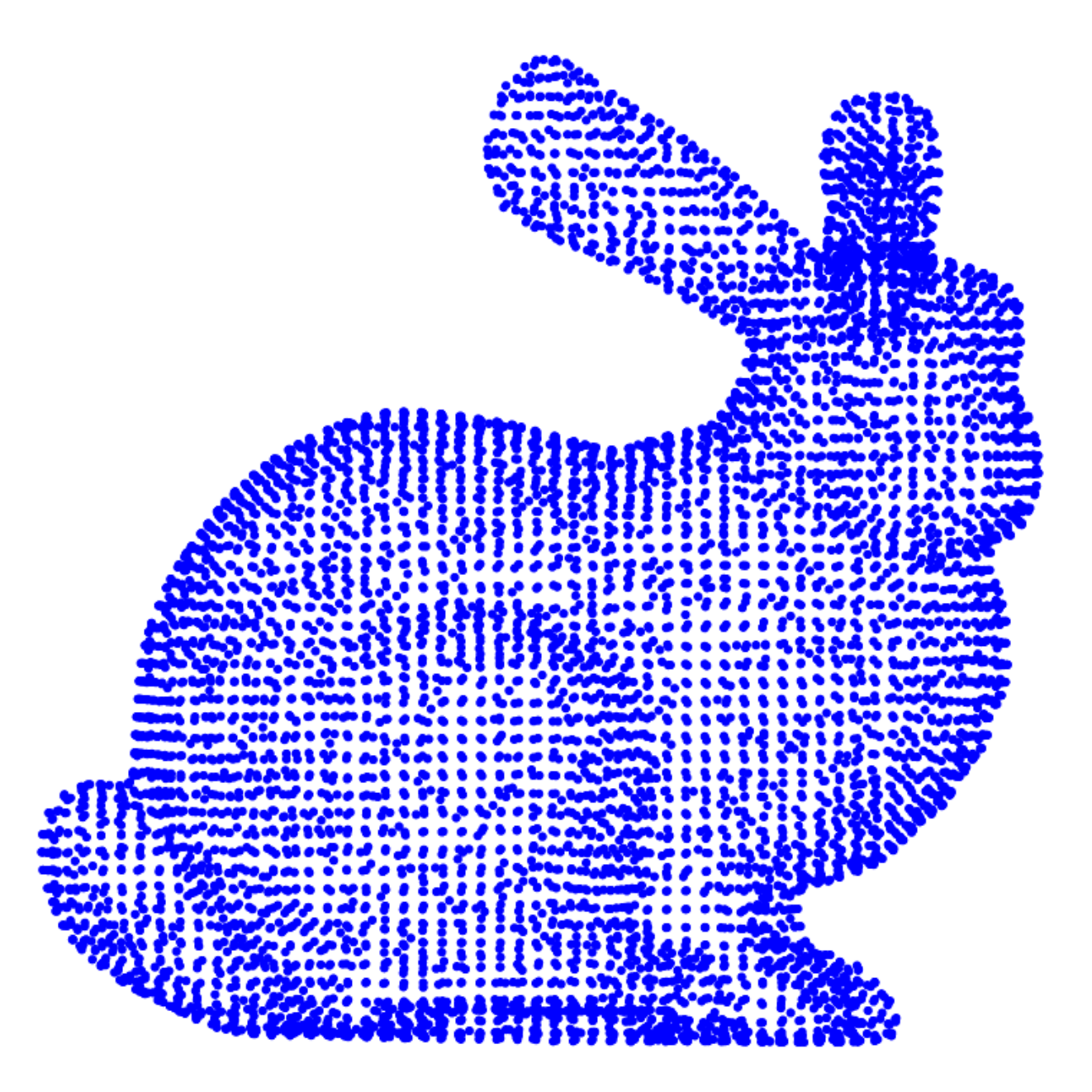} \\
			(a) \bunny model
			\end{minipage}
		& \myhspace 
			\begin{minipage}{\mpw}%
			\centering%
			\includegraphics[width=\columnwidth]{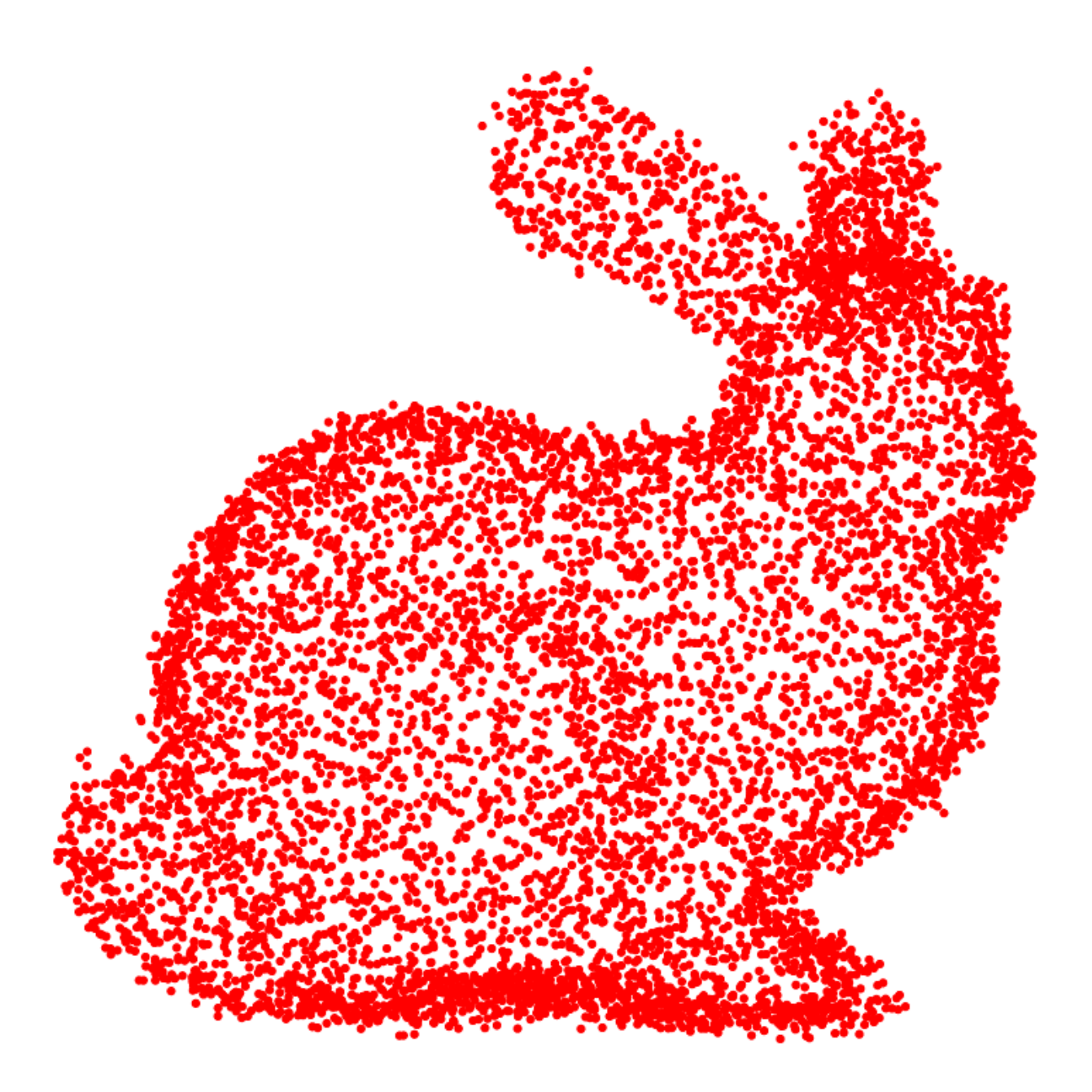} \\
			(b) \bunny scene, $\sigma = 0.01$
			\end{minipage} \\
		 \myhspace 
			\begin{minipage}{\mpw}%
			\centering%
			\includegraphics[width=\columnwidth]{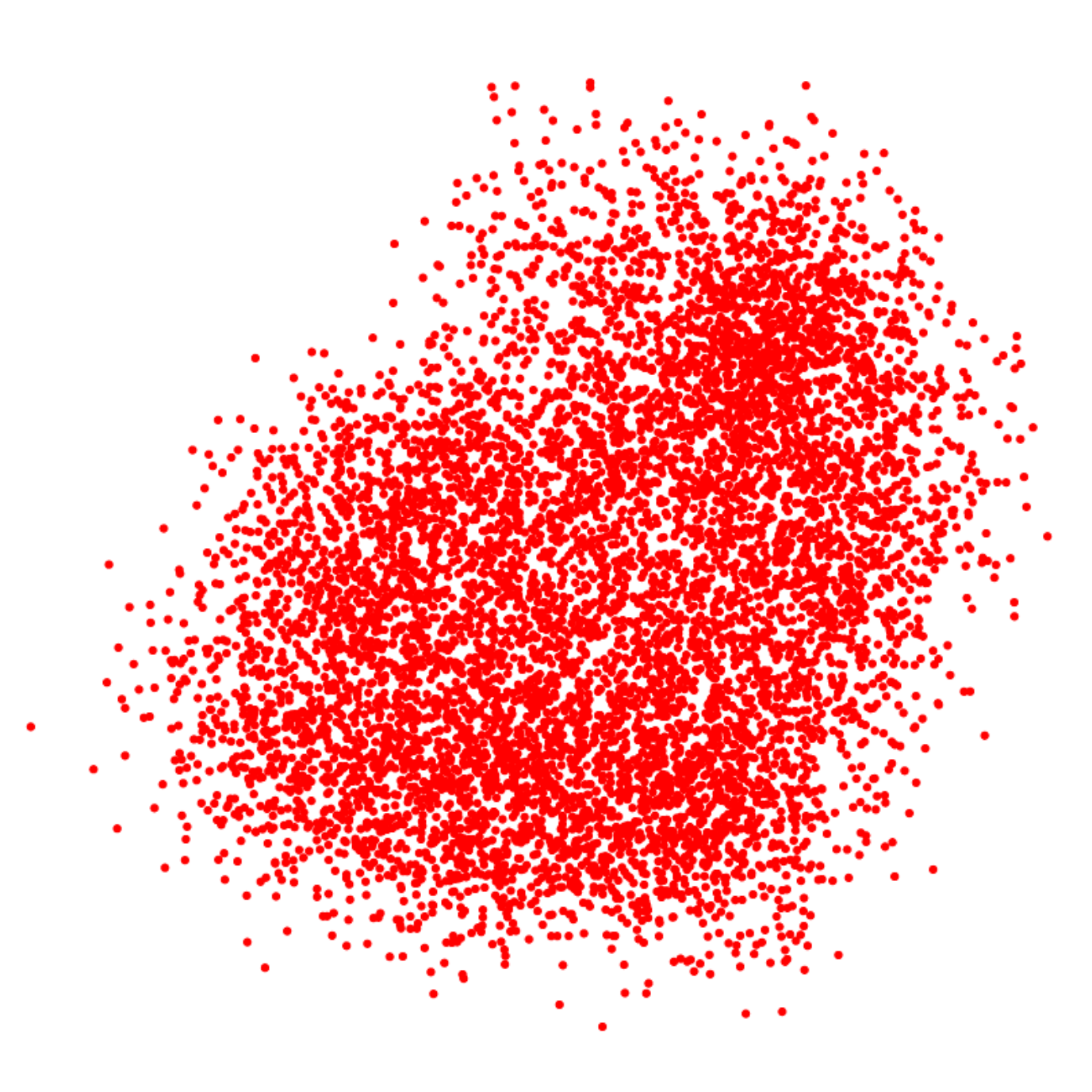} \\
			(c) \bunny scene, $\sigma = 0.1$
			\end{minipage}
		& \myhspace 
			\begin{minipage}{\mpw}%
			\centering%
			\includegraphics[width=\columnwidth]{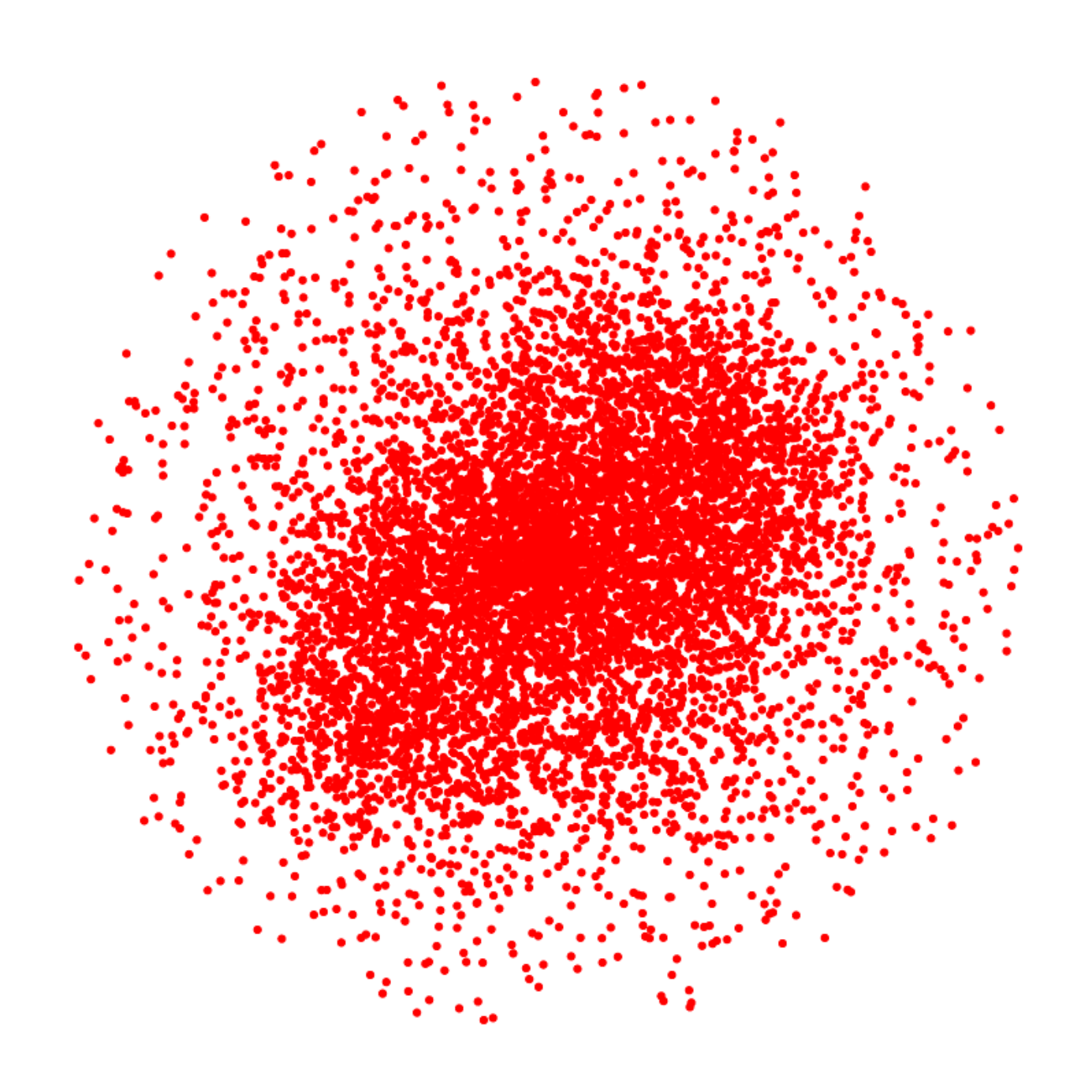} \\
			(d) \bunny scene, $\sigma = 0.1$, \\ 50\% outliers
			\end{minipage}
		\end{tabular}
	\end{minipage}
	\begin{minipage}{\textwidth}
	\end{minipage}
	\vspace{-3mm} 
	\caption{\bunny point cloud scaled inside unit cube $[0,1]^3$ and corrupted by different levels of noise and outliers, all viewed from the same perspective angle. (a) Clean \bunny model point cloud, scaled inside unit cube $[0,1]^3$. (b) \bunny scene, generated from (a) by adding isotropic Gaussian noise with standard deviation $\sigma=0.01$. (c) \bunny scene, generated from (a) by adding isotropic Gaussian noise with $\sigma=0.1$. (d) \bunny scene, generated from (a) by adding isotropic Gaussian noise with $\sigma = 0.1$ and $50\%$ random outliers.
	 \label{fig:bunny_with_noise}}
	\end{center}
\end{figure}

\renewcommand{\myhspace}{\hspace{-6mm}}
\renewcommand{\mpw}{4.5cm}

\begin{figure}[h]
	\begin{center}
	\begin{minipage}{\textwidth}
	\begin{tabular}{cc}%
	\myhspace
			\begin{minipage}{\mpw}%
			\centering%
			\includegraphics[width=\columnwidth]{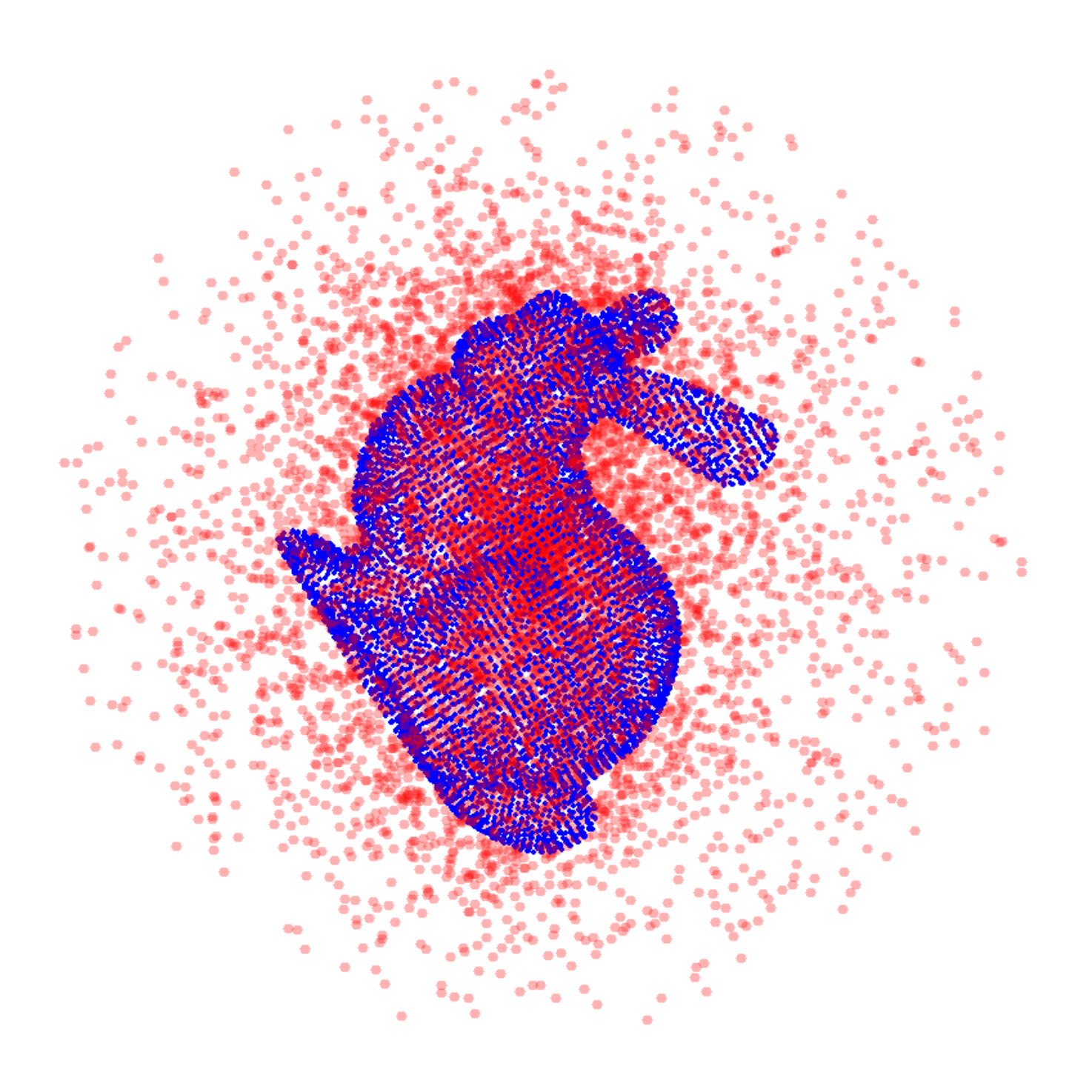} \\
			\end{minipage}
		& \myhspace 
			\begin{minipage}{\mpw}%
			\centering%
			\includegraphics[width=0.97\columnwidth]{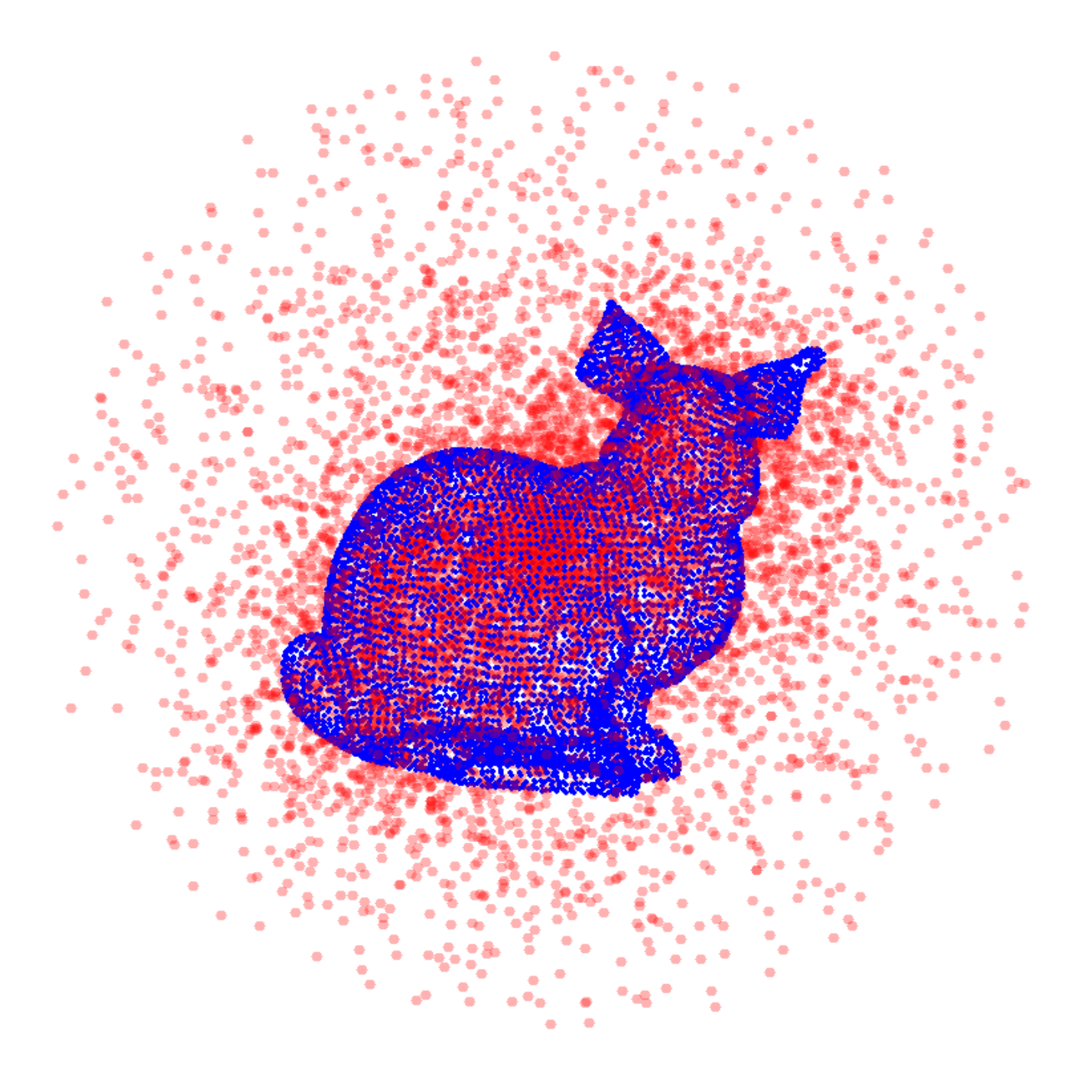} \\
			\end{minipage} 
		\end{tabular}
	\end{minipage}
	\begin{minipage}{\textwidth}
	\end{minipage}
	\vspace{-3mm} 
	\caption{ A single representative \name registration result with $\sigma=0.1$ noise corruption and $50\%$ outliers, viewed from two distinctive perspectives. Clean \bunny model showed in blue and cluttered scene showed in red. Although the rotation error compared to ground-truth is $3.42^{\circ}$ and the translation error compared to ground-truth is $0.098$m, it is challenging for a human 
	to confirm the correctness of the registration result. 
	 \label{fig:bunny_teaser_high_noise}}
	\vspace{-7mm} 
	\end{center}
\end{figure}

\subsection{Object Localization and Pose Estimation}
\label{sec:objectPoseEstimation}

We use the large-scale point cloud datasets from the University of Washington~\cite{Lai11icra-largeRGBD} to test \name in \emph{object pose estimation} applications. Seven scenes containing a \emph{cereal box} and one scene containing a \emph{cap} are selected from the dataset. We first use the ground-truth object labels to extract the \emph{cereal box/cap} out of the scene and treat it as the object, then apply a random rotation and translation to the scene, to get an object-scene pair. To register the object-scene pair of point clouds, we first use FPFH feature descriptors~\cite{Rusu09icra-fast3Dkeypoints} as implemented in the Point Cloud Library (PCL) to establish putative feature correspondences. 
Given correspondences from FPFH, \name is used to find the relative pose between the object and scene. We downsample the object and scene  using the same ratio (about 0.1) to make the object have 2,000 points, so that FPFH and \name can run in reasonable time. Fig.~\ref{fig:objectPoseEstimation} shows the registration results for all 8 selected scenes. The inlier correspondence ratios for \emph{cereal box} are all below 10\% and typically below 5\%. \name is able to compute a highly-accurate estimate of the relative pose using a handful of inliers. 
Due to the distinctive shape of \emph{cap}, FPFH produces a higher number of inliers and \name is able to register the object-scene pair without problems. 
Table~\ref{tab:objectPoseEstimation} shows the mean and standard deviation (STD) of the rotation and translation errors, the number of FPFH correspondences, and the FPFH inlier ratio estimated by \name on the eight scenes.


\begin{table}
\centering
\begin{tabular}{ccccc}
& \shortstack{Rotation \\ error [rad]} & \shortstack{Translation \\ error [m]} & \shortstack{\# of FPFH \\ correspondences} &\shortstack{FPFH inlier \\ ratio [\%]} \\
\hline
Mean & 0.0665 & 0.0695 & 525 & 6.53 \\
\hline
STD & 0.0435 & 0.0526 & 161 & 4.59\\
\hline
\end{tabular}
\caption{Registration results on eight scenes of the RGB-D dataset~\cite{Lai11icra-largeRGBD}.}
\label{tab:objectPoseEstimation}
\end{table}


\renewcommand{\mpw}{6cm}

\begin{figure*}[h]
	\begin{center}
	\begin{minipage}{\textwidth}
	\hspace{-0.2cm}
	\begin{tabular}{ccc}%
			\begin{minipage}{\mpw}%
			\centering%
			\includegraphics[width=1.0\columnwidth]{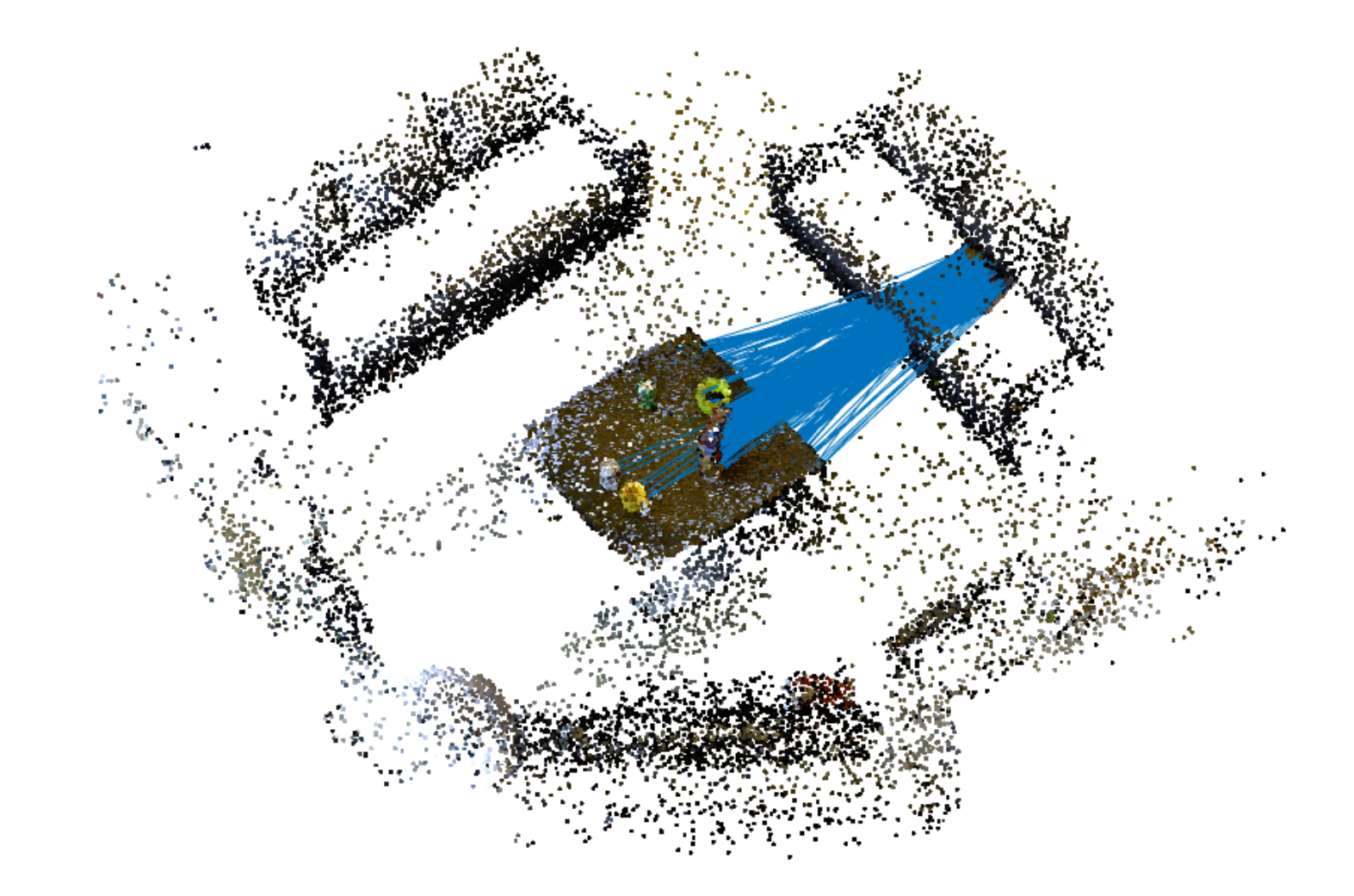} \\
			\end{minipage}
		& \myhspace
			\begin{minipage}{\mpw}%
			\centering%
			\includegraphics[width=1.0\columnwidth]{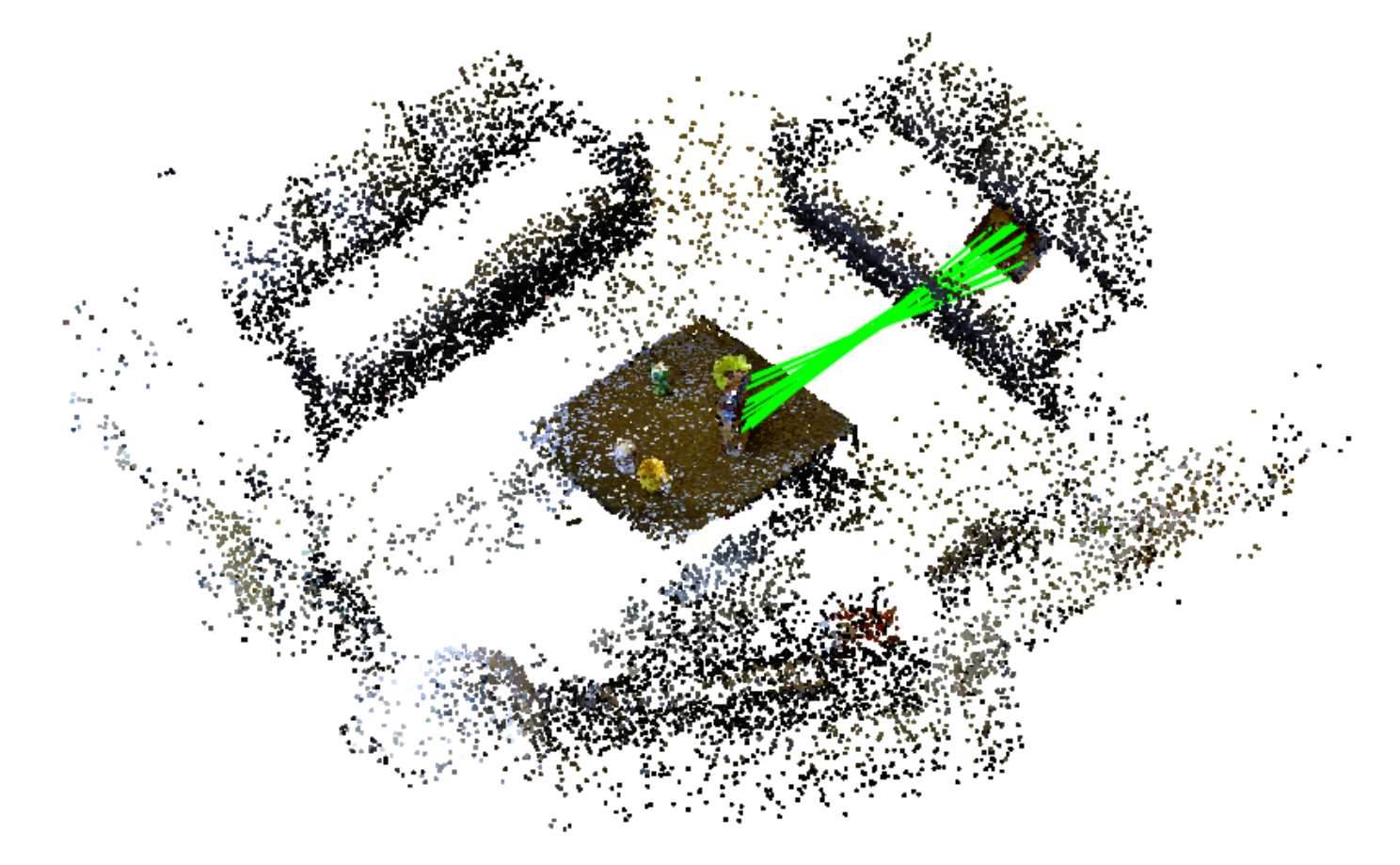} \\
			\end{minipage}
		& \myhspace
			\begin{minipage}{\mpw}%
			\centering%
			\includegraphics[width=1.0\columnwidth]{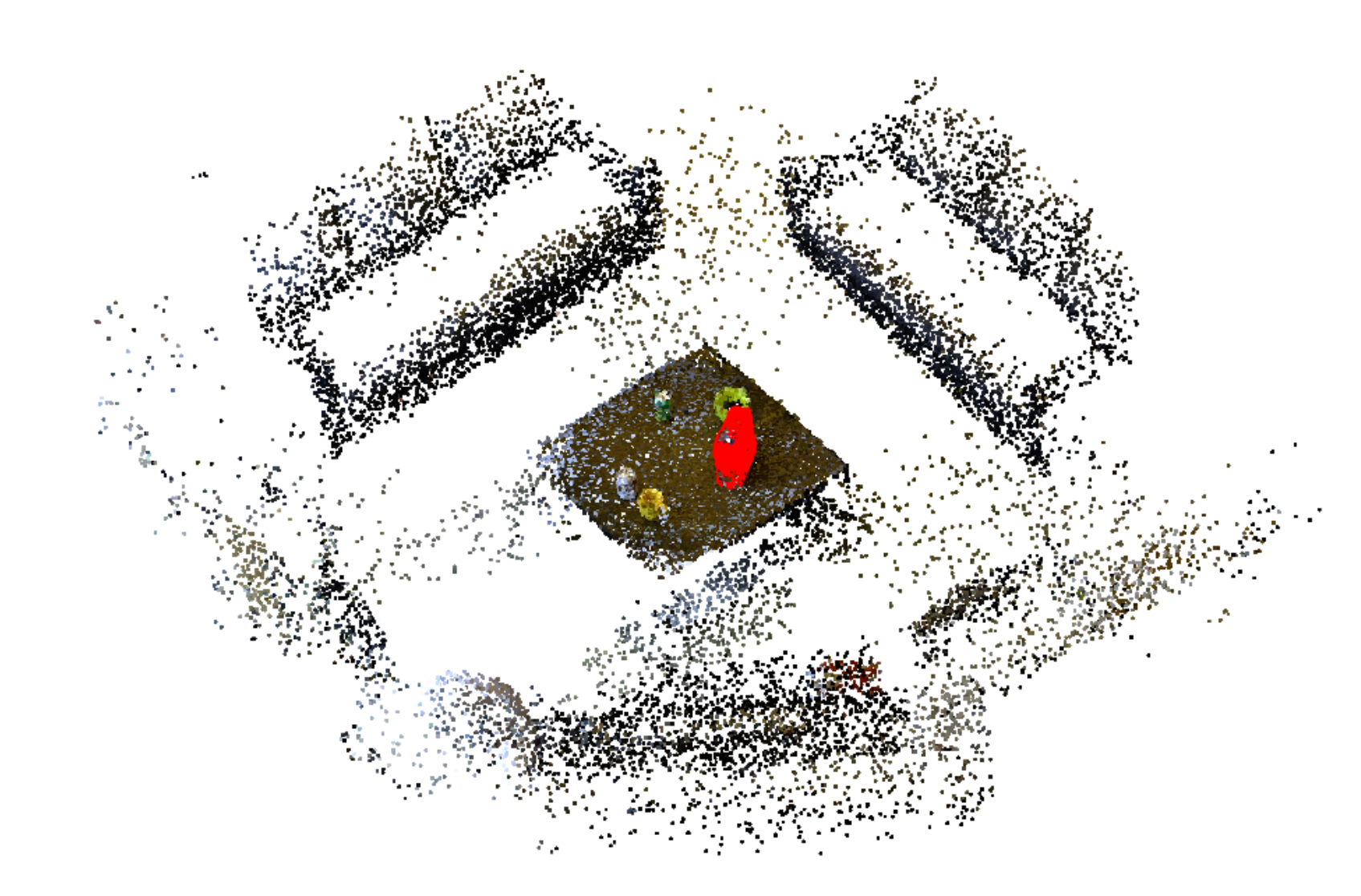} \\
			\end{minipage} \\
		\multicolumn{3}{c}{\emph{scene-2}, \# of FPFH correspondences: 550, Inlier ratio: 4.55\%, Rotation error: 0.120, Translation error: 0.052.}\\
		
		\begin{minipage}{\mpw}%
			\centering%
			\includegraphics[width=1.0\columnwidth]{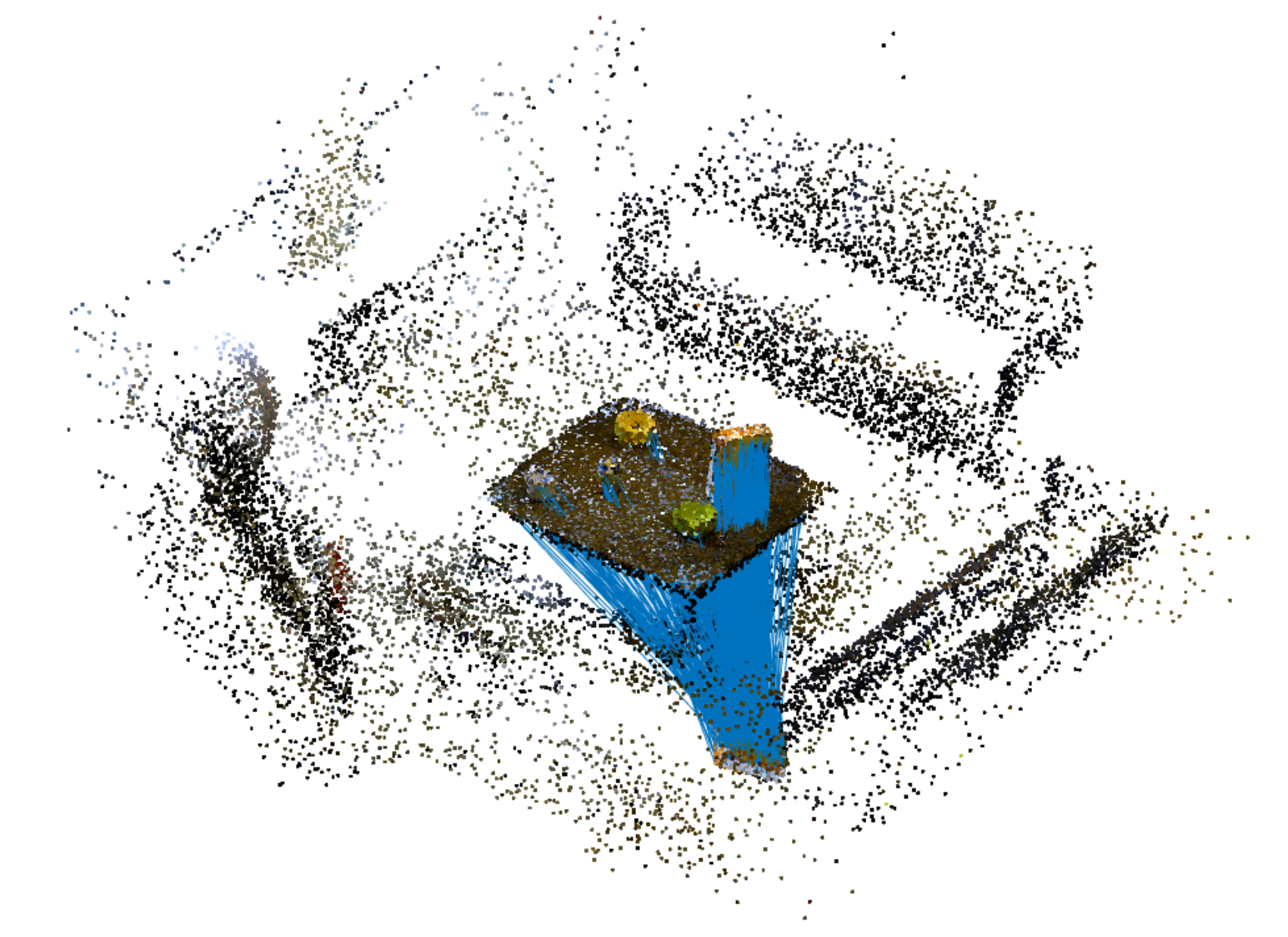} \\
			\end{minipage}
		& \myhspace
			\begin{minipage}{\mpw}%
			\centering%
			\includegraphics[width=1.0\columnwidth]{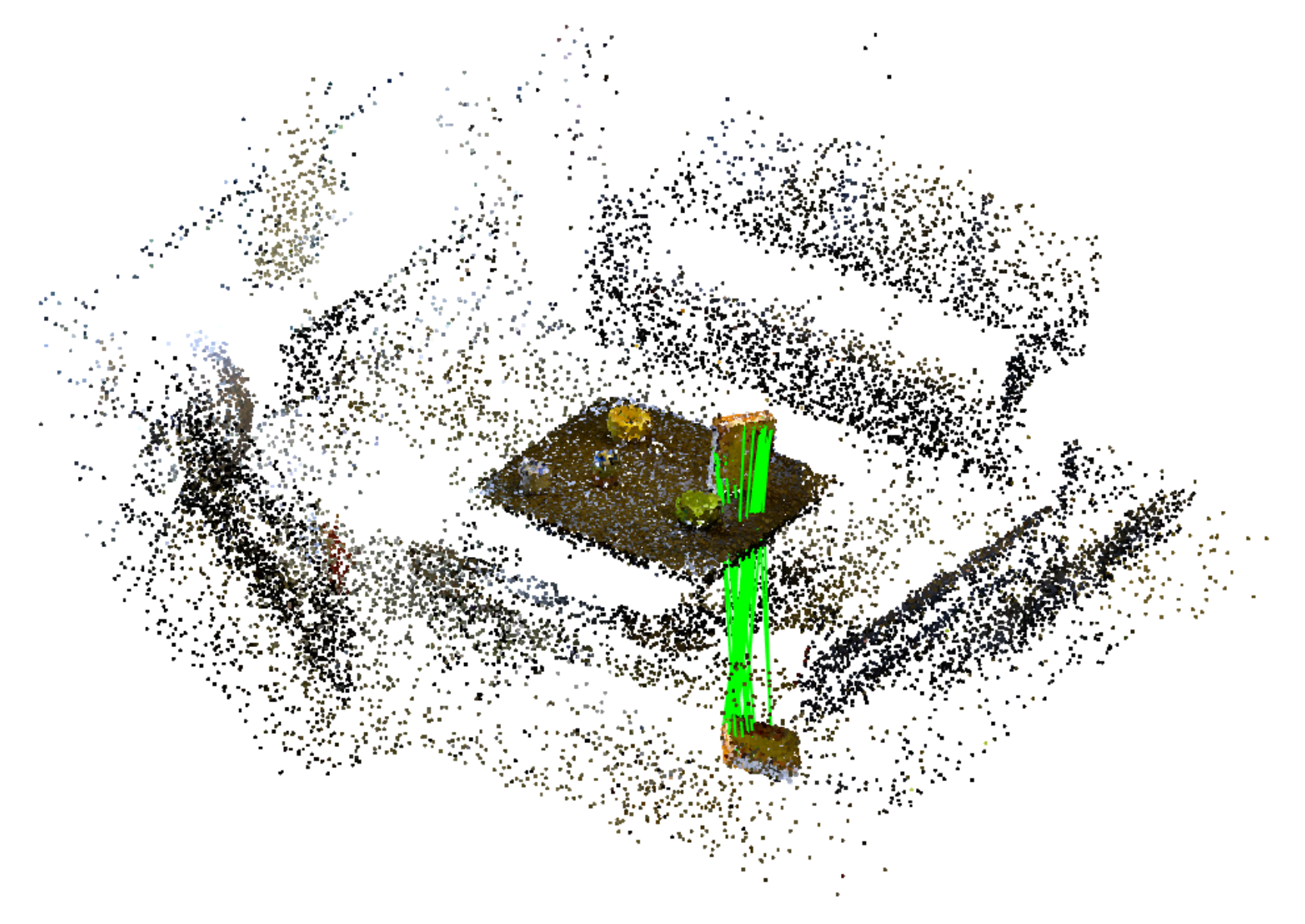} \\
			\end{minipage}
		& \myhspace
			\begin{minipage}{\mpw}%
			\centering%
			\includegraphics[width=1.0\columnwidth]{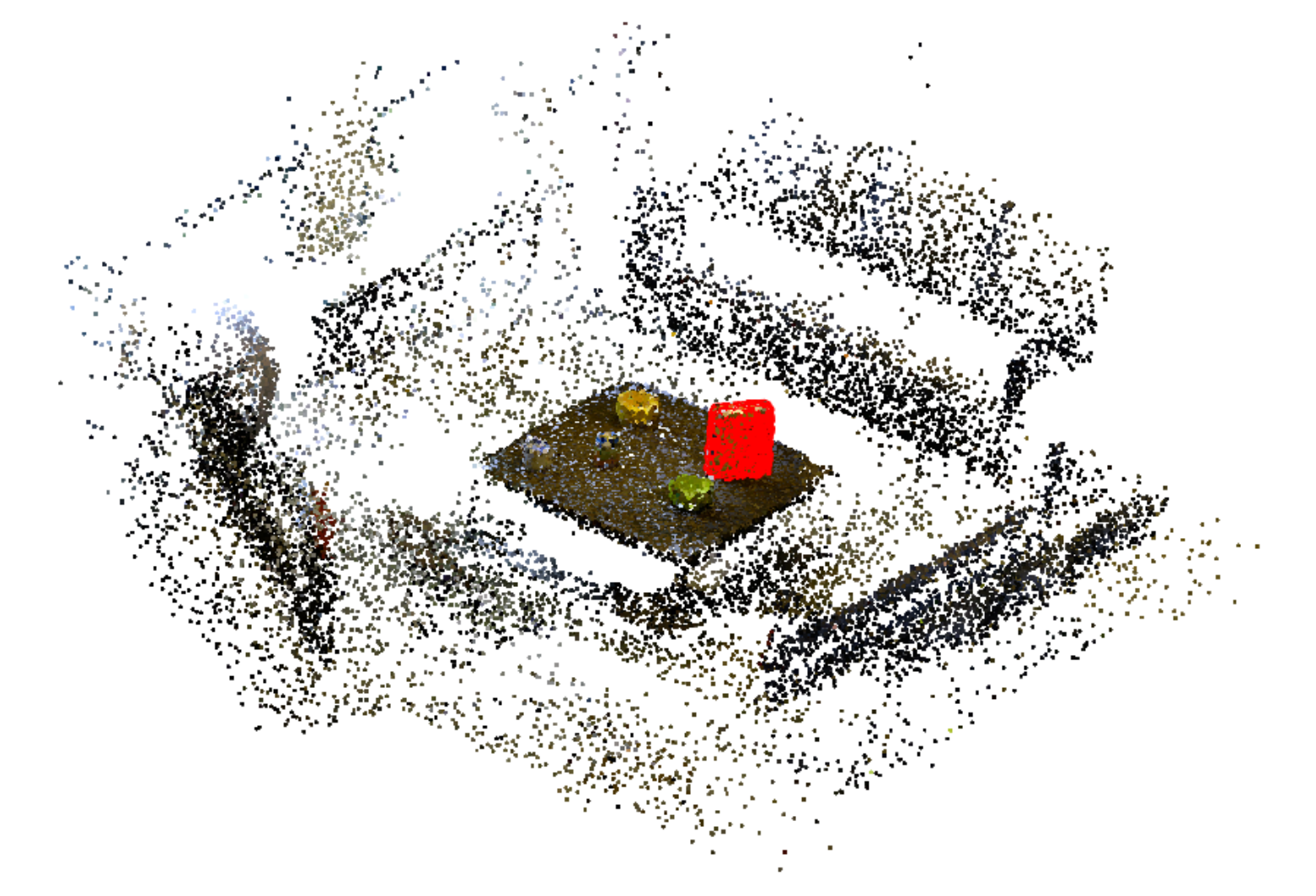} \\
			\end{minipage} \\
\multicolumn{3}{c}{\emph{scene-4}, \# of FPFH correspondences: 636, Inlier ratio: 4.56\%, Rotation error: 0.042, Translation error: 0.051.}\\

		\begin{minipage}{\mpw}%
			\centering%
			\includegraphics[width=1.0\columnwidth]{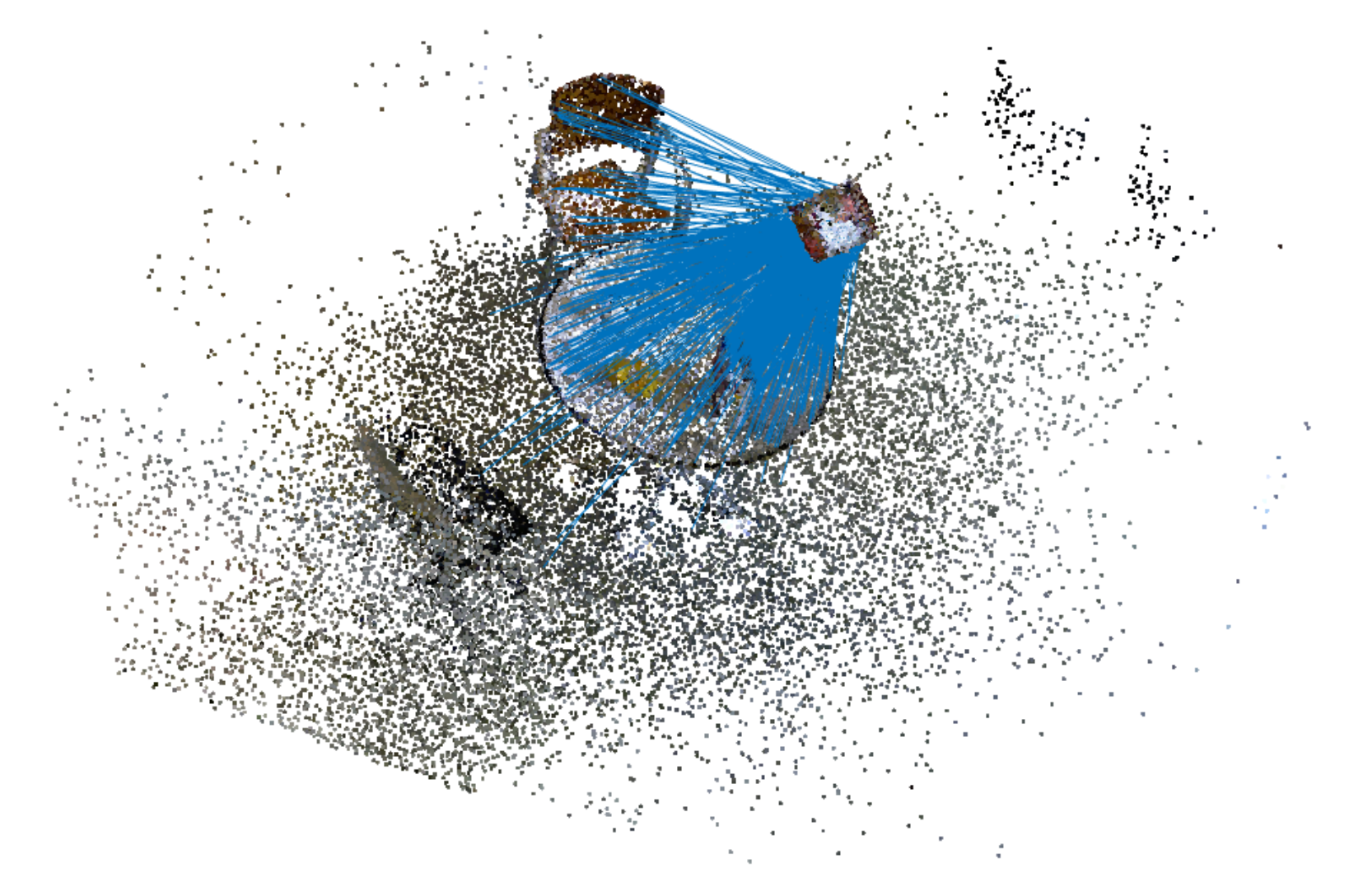} \\
			\end{minipage}
		& \myhspace
			\begin{minipage}{\mpw}%
			\centering%
			\includegraphics[width=1.0\columnwidth]{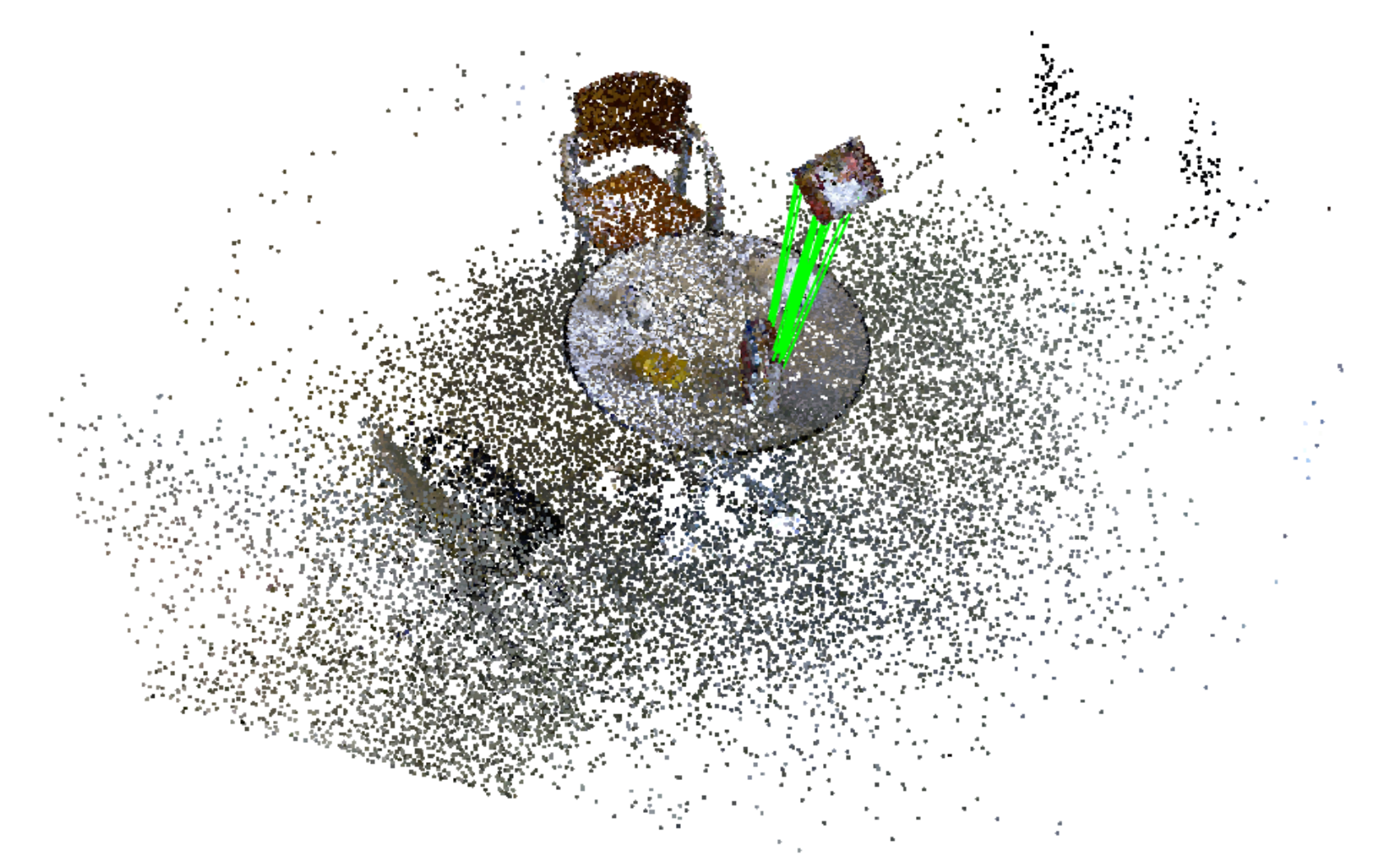} \\
			\end{minipage}
		& \myhspace
			\begin{minipage}{\mpw}%
			\centering%
			\includegraphics[width=1.0\columnwidth]{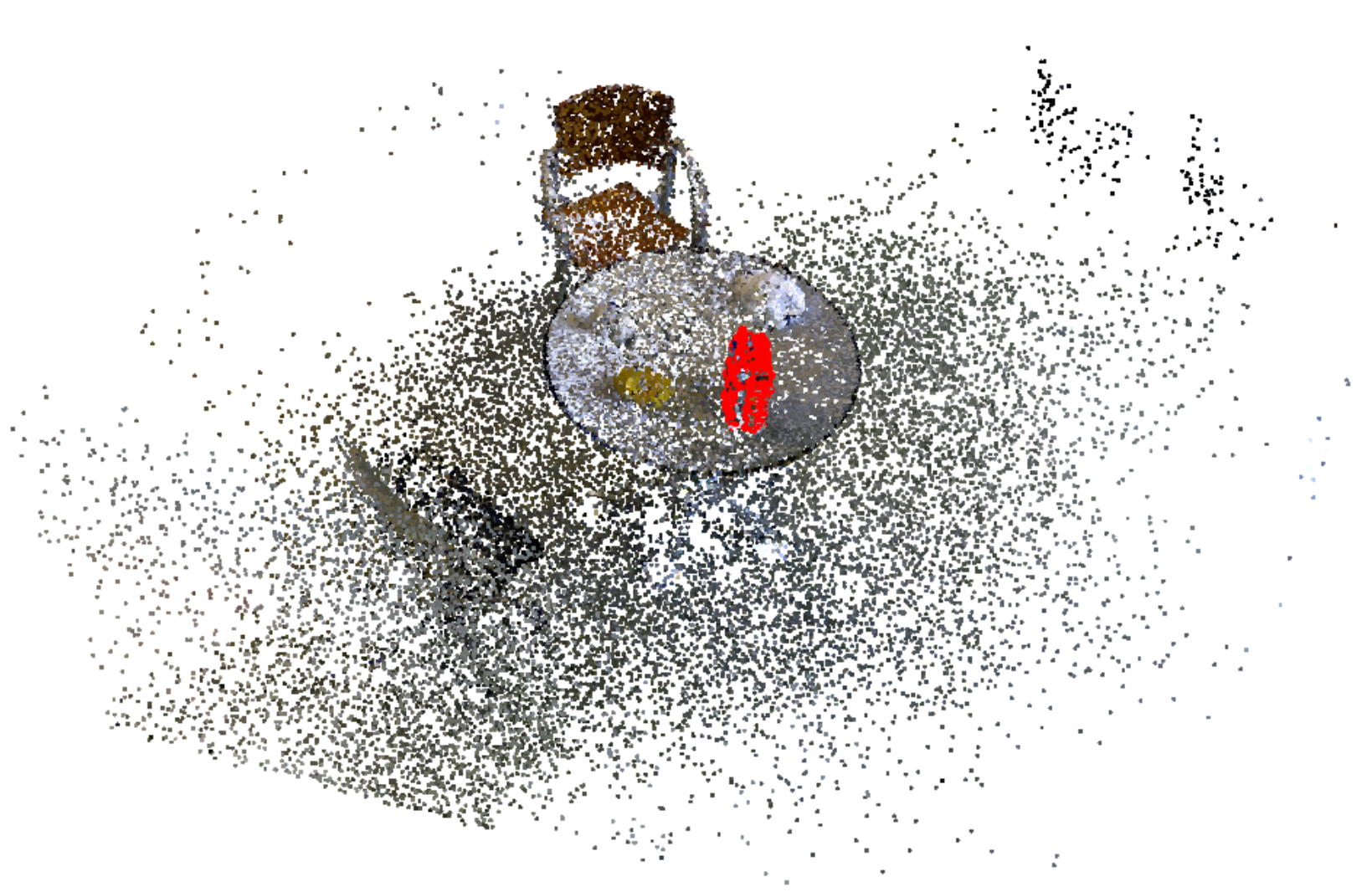} \\
			\end{minipage} \\
		\multicolumn{3}{c}{\emph{scene-5}, \# of FPFH correspondences: 685, Inlier ratio: 2.63\%, Rotation error: 0.146, Translation error: 0.176.}\\
		
		\begin{minipage}{\mpw}%
			\centering%
			\includegraphics[width=1.0\columnwidth]{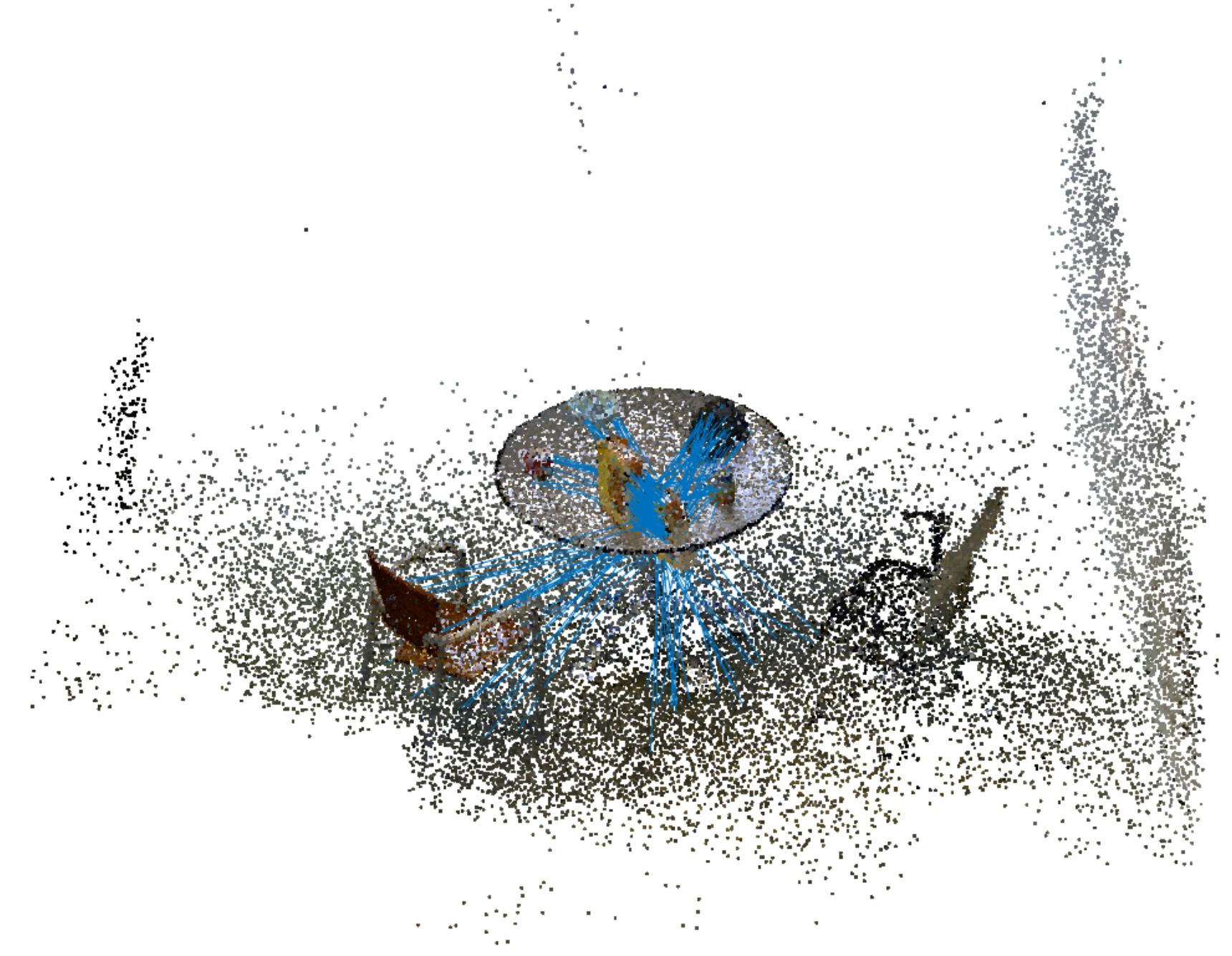} \\
			\end{minipage}
		& \myhspace
			\begin{minipage}{\mpw}%
			\centering%
			\includegraphics[width=1.0\columnwidth]{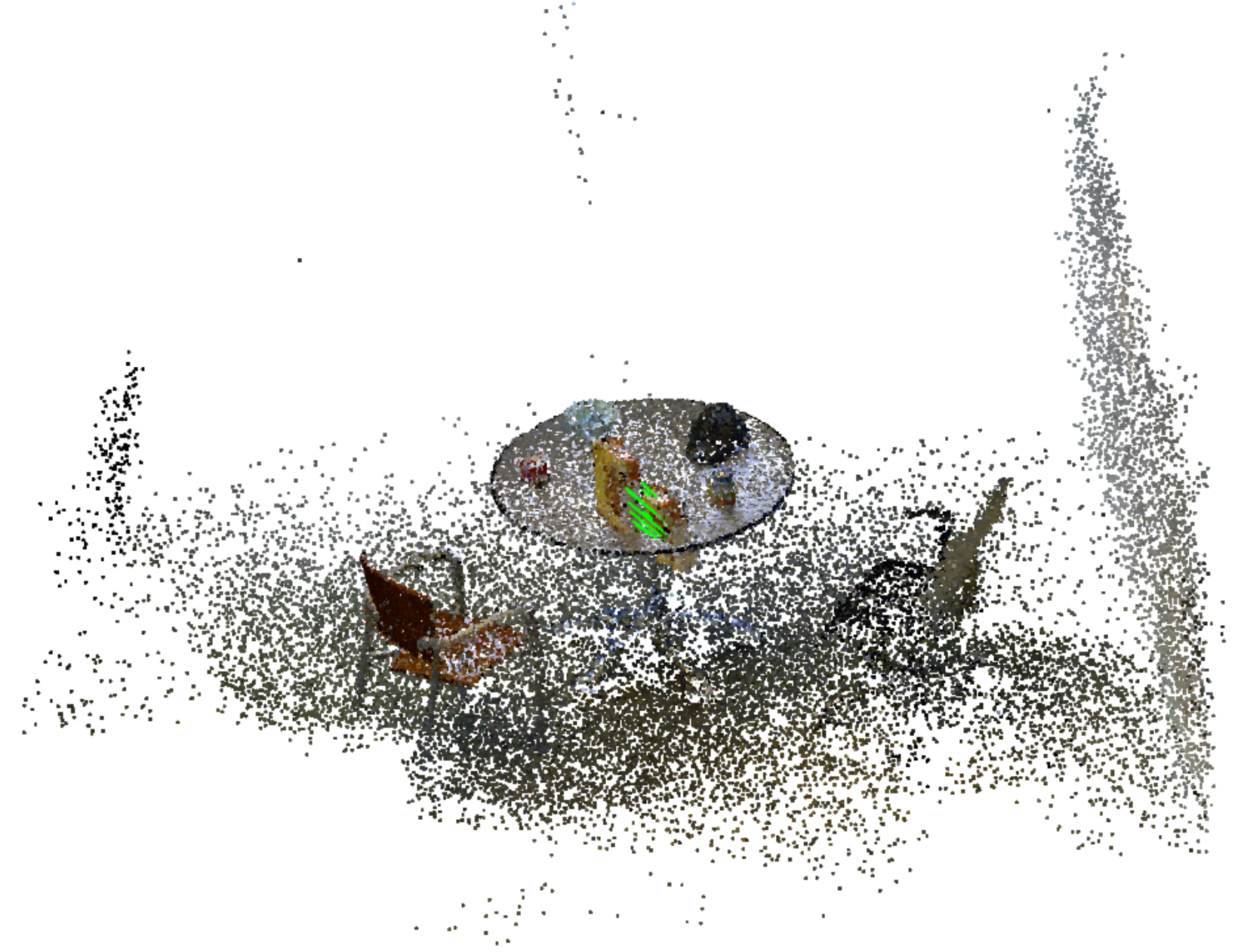} \\
			\end{minipage}
		& \myhspace
			\begin{minipage}{\mpw}%
			\centering%
			\includegraphics[width=1.0\columnwidth]{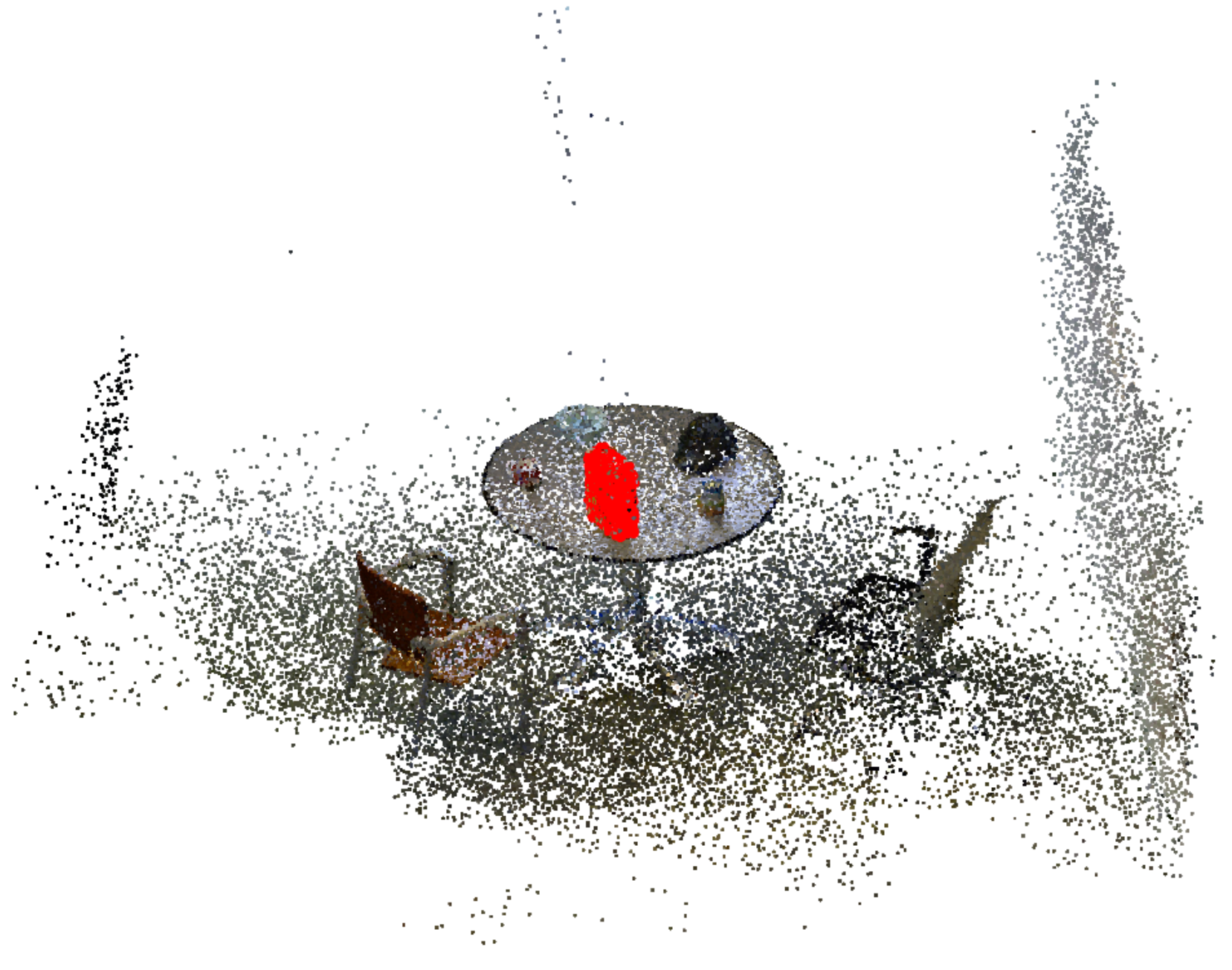} \\
			\end{minipage} \\
		\multicolumn{3}{c}{\emph{scene-7}, \# of FPFH correspondences: 416, Inlier ratio: 3.13\%, Rotation error: 0.058, Translation error: 0.097.} 
		
		\end{tabular}
	\end{minipage}
	\vspace{-3mm} 
	\caption{Object pose estimation on the large-scale RGB-D dataset~\cite{Lai11icra-largeRGBD}. First column: FPFH correspondences, second column: inlier correspondences after \name, third column: registration result with the registered object highlighted in red. Scene number and related registration information are listed below each scene.}
	\vspace{-8mm} 
	\end{center}
\end{figure*}

\begin{figure*}[h]\ContinuedFloat
	\begin{center}
	\begin{minipage}{\textwidth}
	\hspace{-0.2cm}
	\begin{tabular}{ccc}%
		\begin{minipage}{\mpw}%
			\centering%
			\includegraphics[width=1.0\columnwidth]{scene_5_a_9.pdf} \\
			\end{minipage}
		& \myhspace
			\begin{minipage}{\mpw}%
			\centering%
			\includegraphics[width=1.0\columnwidth]{scene_5_b_9.pdf} \\
			\end{minipage}
		& \myhspace
			\begin{minipage}{\mpw}%
			\centering%
			\includegraphics[width=1.0\columnwidth]{scene_5_c_9.pdf} \\
			\end{minipage} \\
	\multicolumn{3}{c}{\emph{scene-9}, \# of FPFH correspondences: 651, Inlier ratio: 8.29\%, Rotation error: 0.036, Translation error: 0.011.} \\
	
		\begin{minipage}{\mpw}%
			\centering%
			\includegraphics[width=1.0\columnwidth]{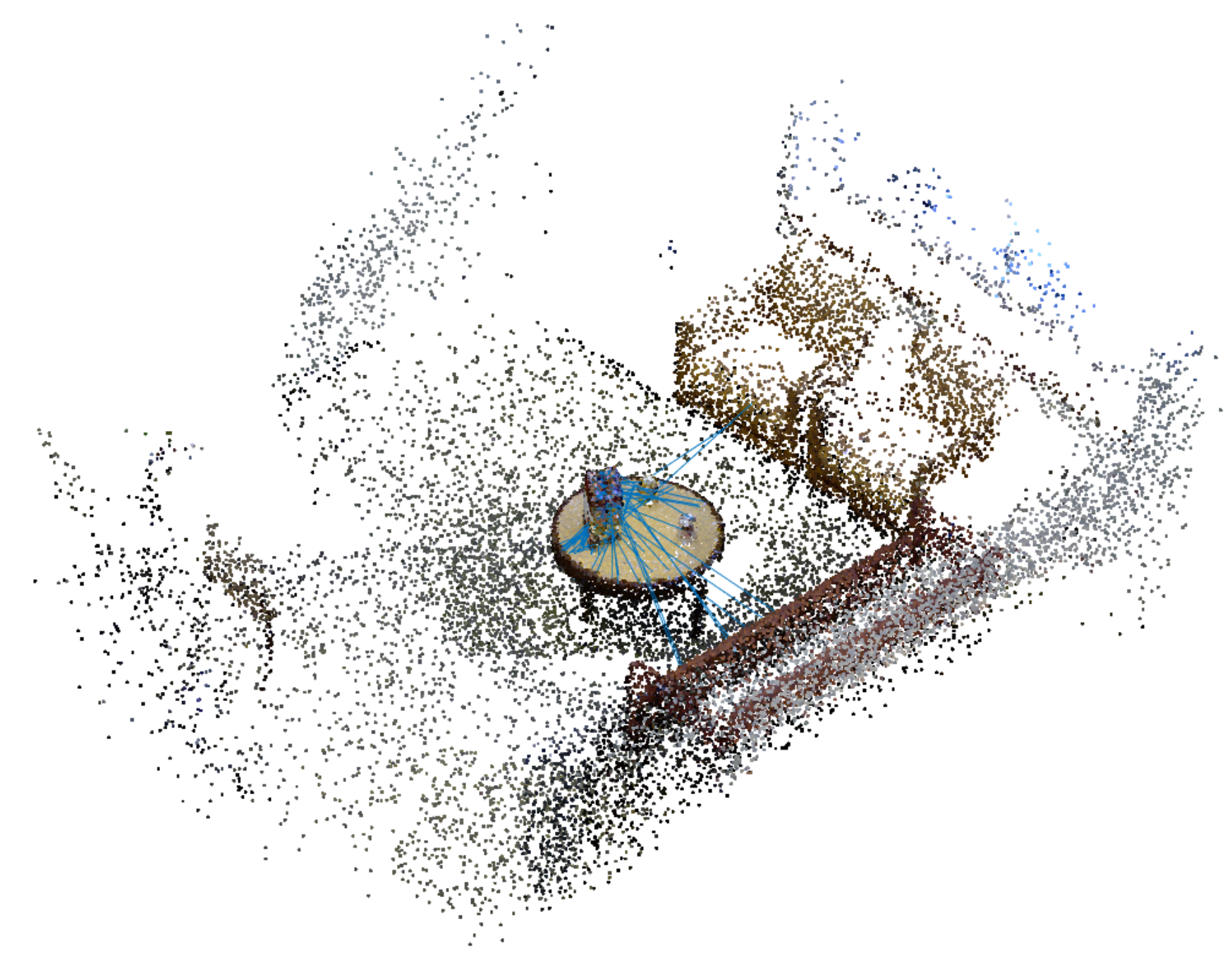} \\
			\end{minipage}
		& \myhspace
			\begin{minipage}{\mpw}%
			\centering%
			\includegraphics[width=1.0\columnwidth]{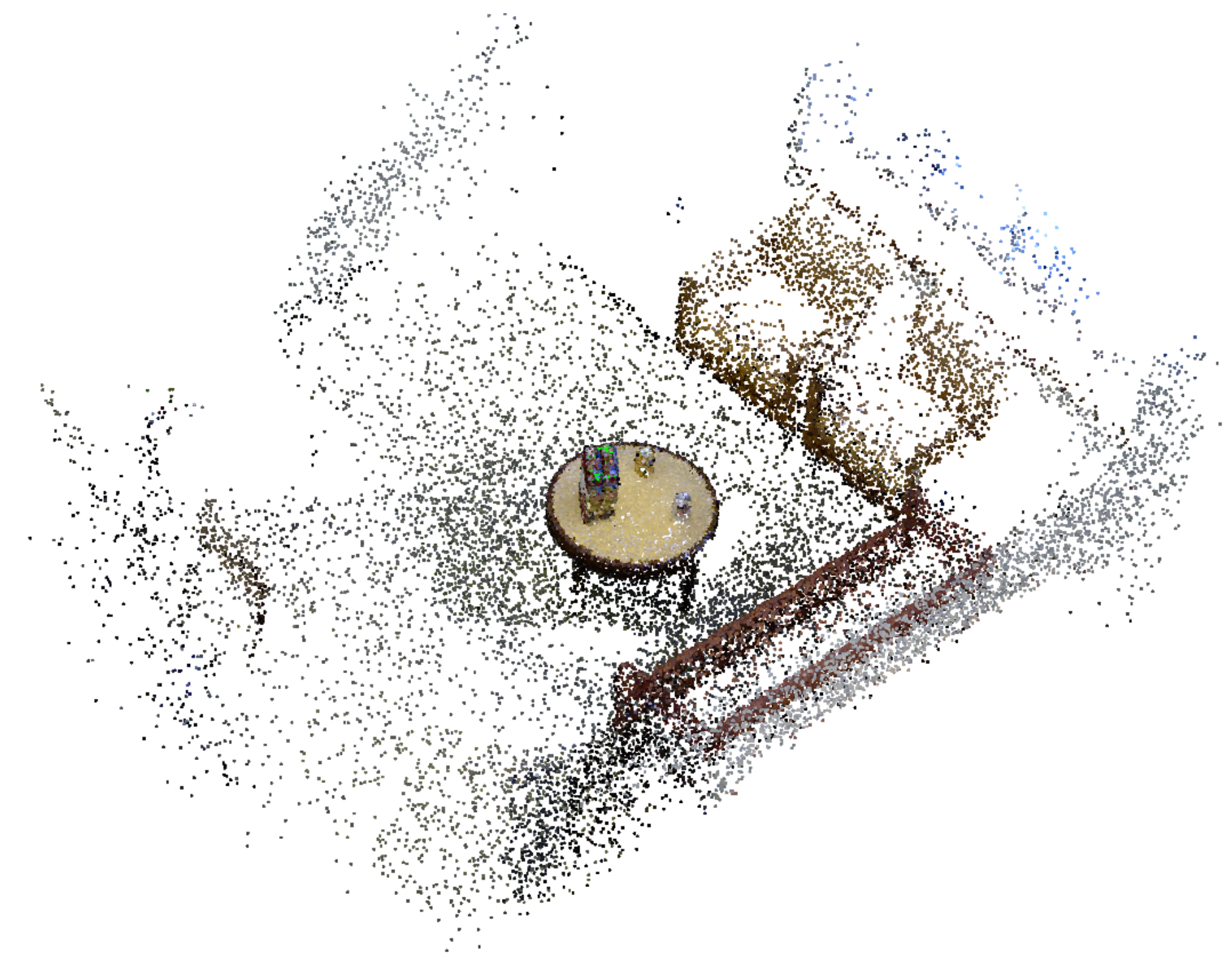} \\
			\end{minipage}
		& \myhspace
			\begin{minipage}{\mpw}%
			\centering%
			\includegraphics[width=1.0\columnwidth]{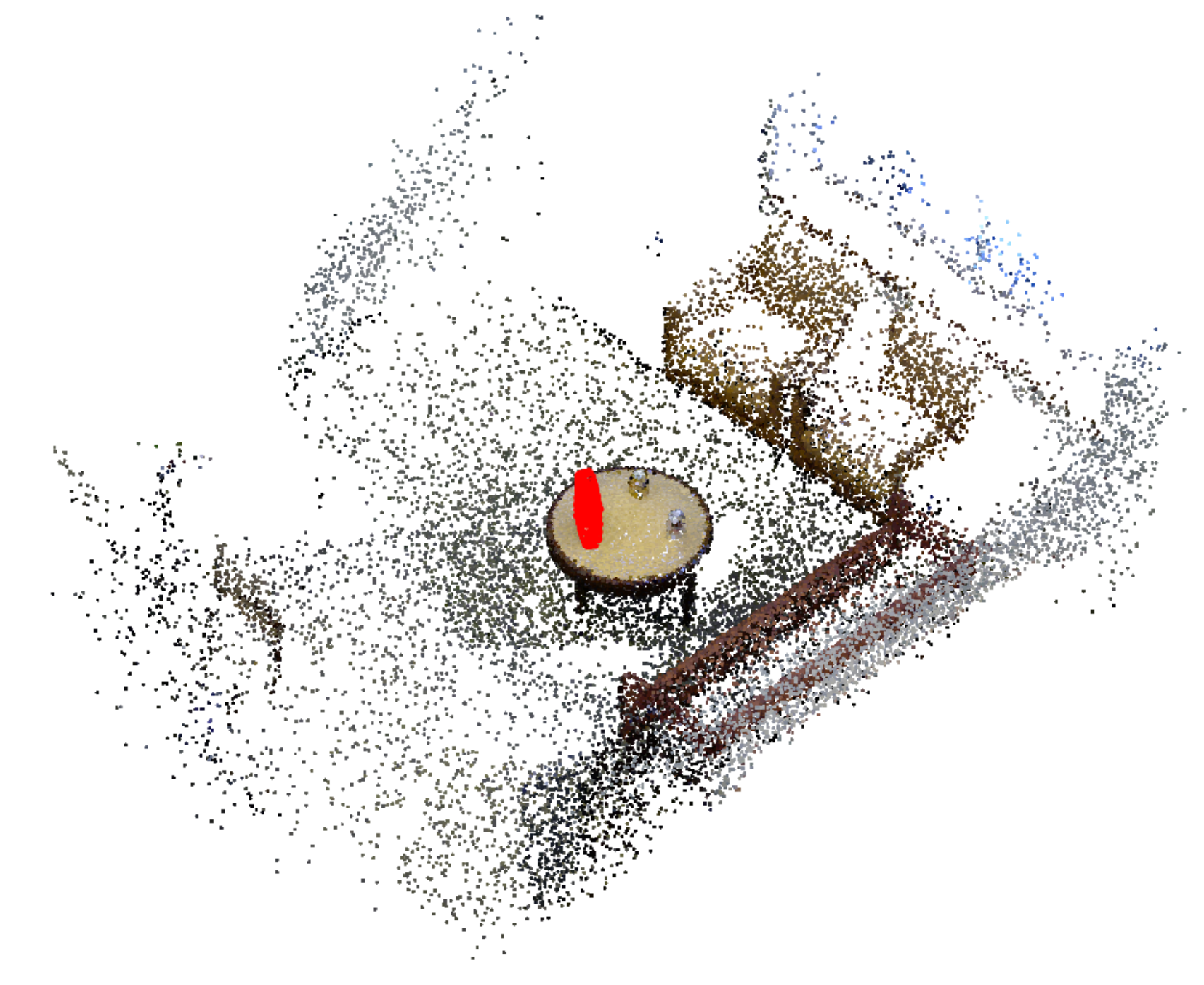} \\
			\end{minipage} \\
		\multicolumn{3}{c}{\emph{scene-11}, \# of FPFH correspondences: 445, Inlier ratio: 6.97\%, Rotation error: 0.028, Translation error: 0.016.} \\
		
		\begin{minipage}{\mpw}%
			\centering%
			\includegraphics[width=1.0\columnwidth]{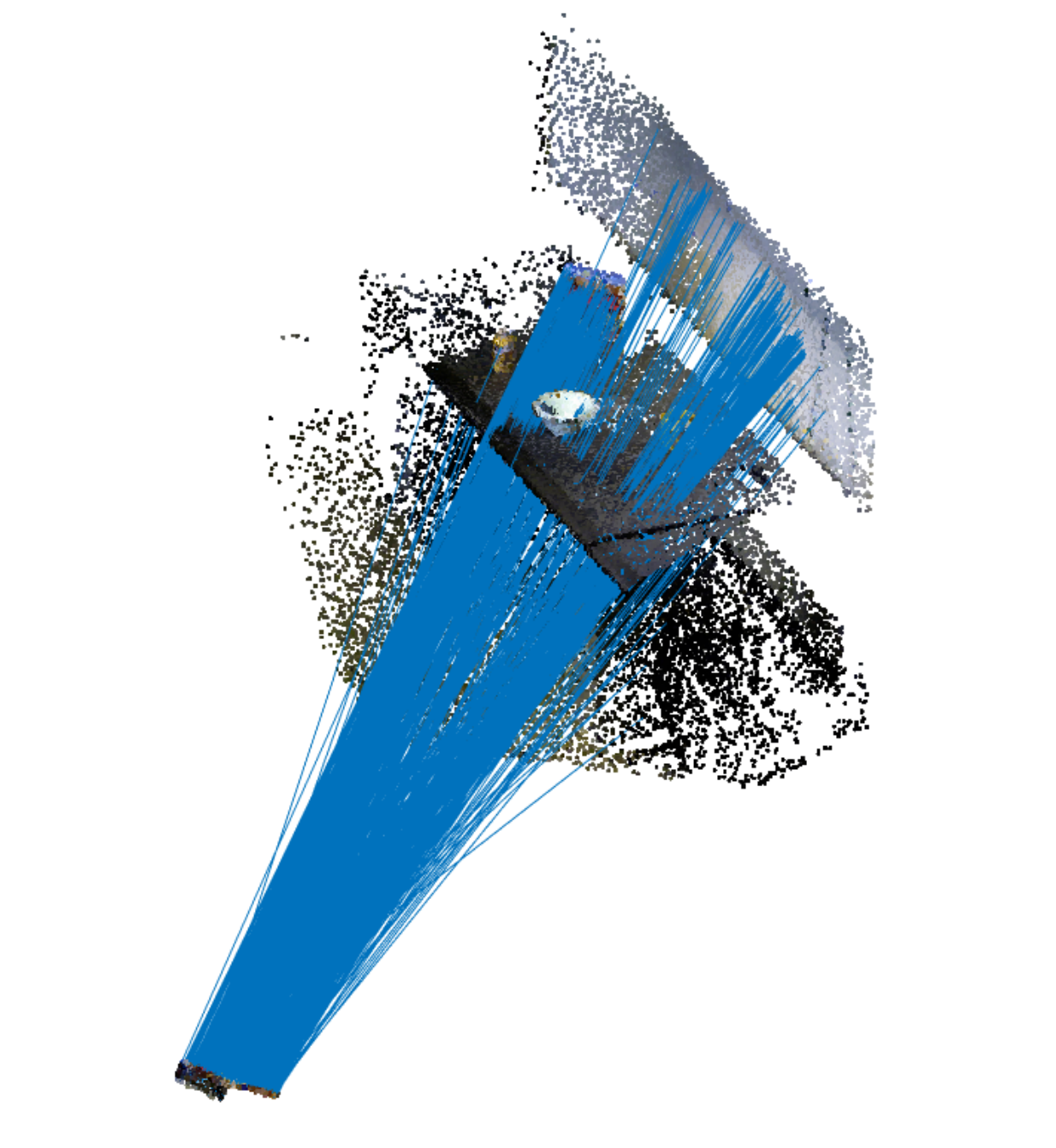} \\
			\end{minipage}
		& \myhspace
			\begin{minipage}{\mpw}%
			\centering%
			\includegraphics[width=1.0\columnwidth]{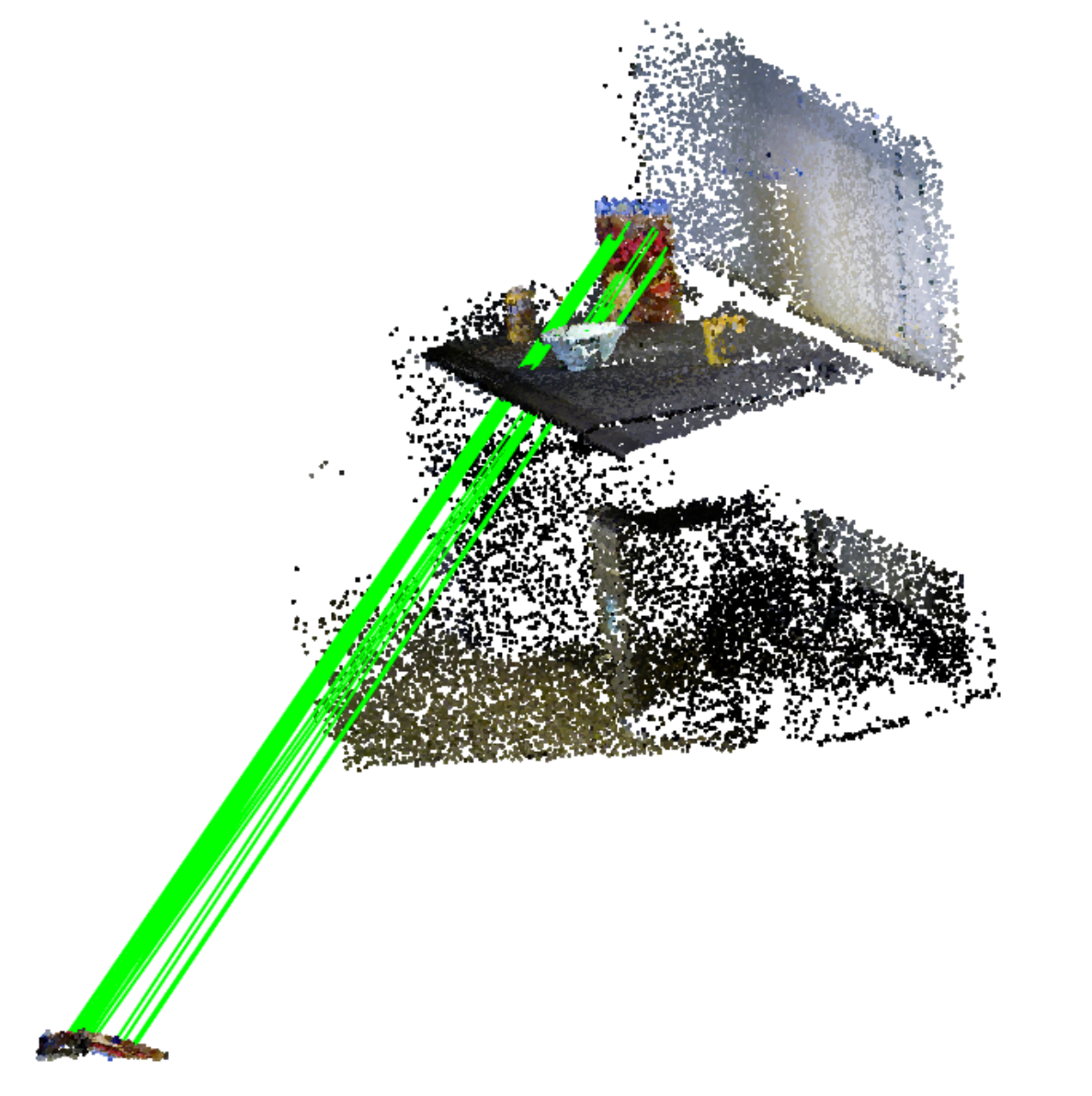} \\
			\end{minipage}
		& \myhspace
			\begin{minipage}{\mpw}%
			\centering%
			\includegraphics[width=1.0\columnwidth]{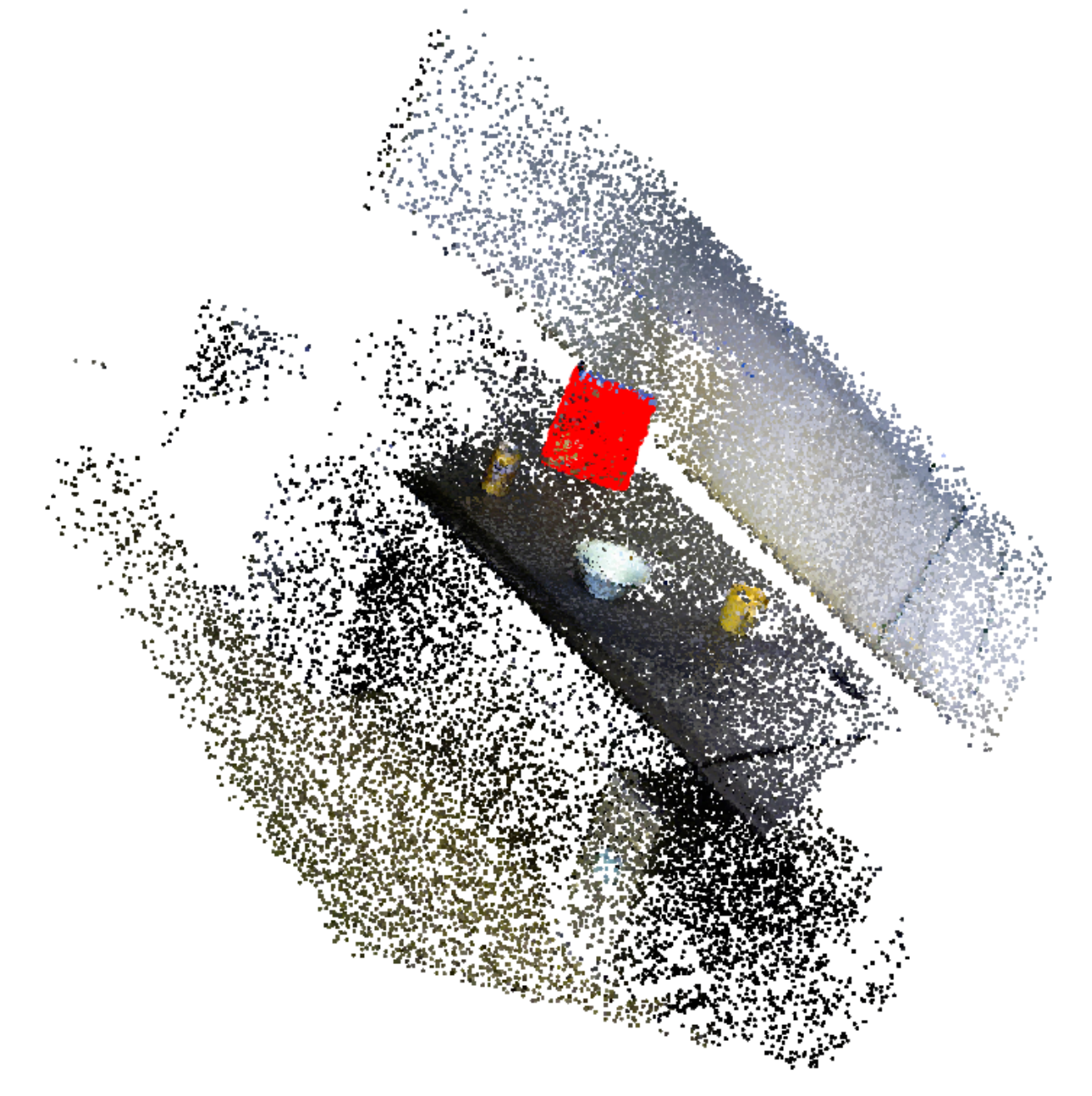} \\
			\end{minipage}\\
		\multicolumn{3}{c}{\emph{scene-13}, \# of FPFH correspondences: 612, Inlier ratio: 5.23\%, Rotation error: 0.036, Translation error: 0.064.} \\
		
		\begin{minipage}{\mpw}%
			\centering%
			\includegraphics[width=1.0\columnwidth]{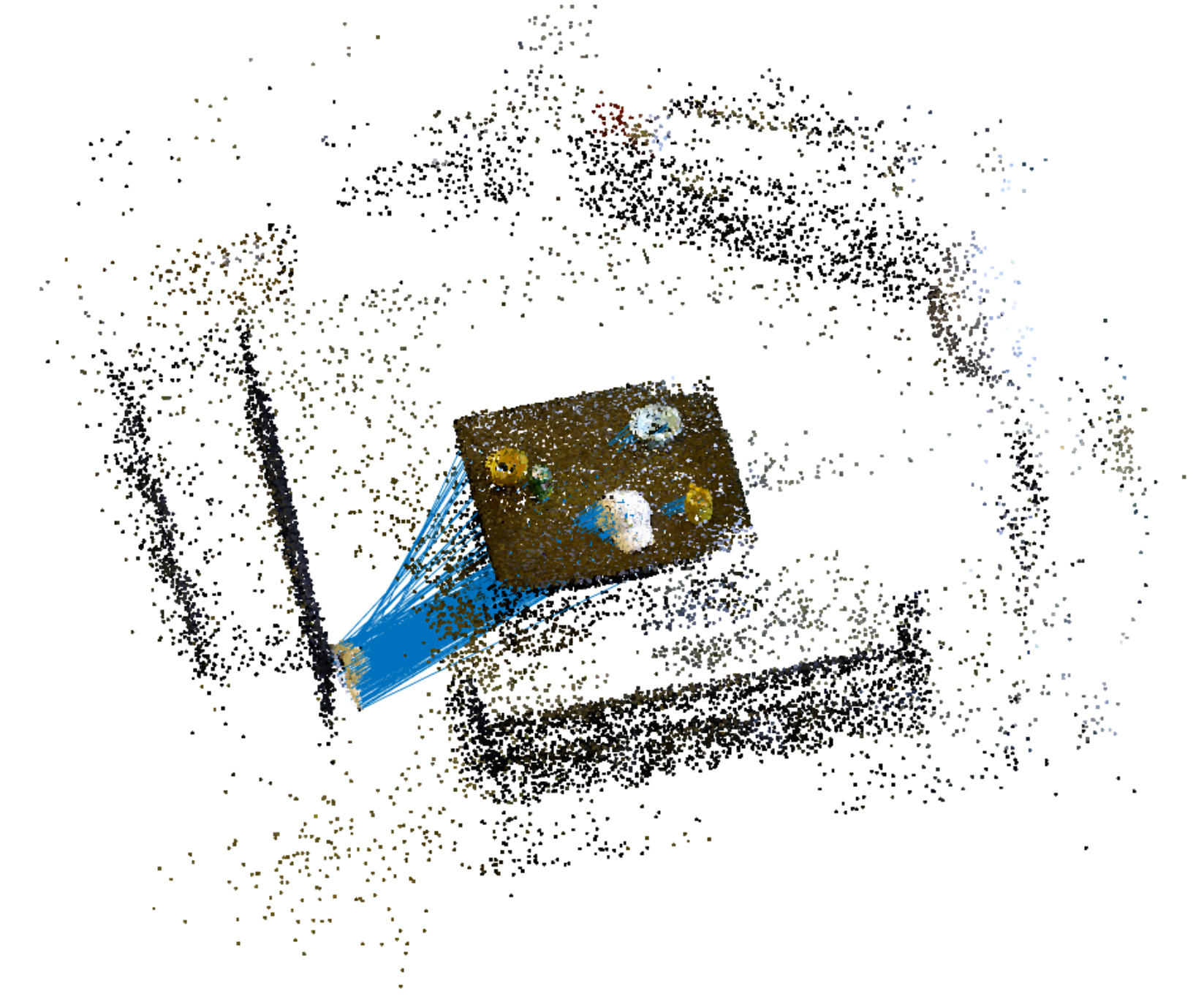} \\
			\end{minipage}
		& \myhspace
			\begin{minipage}{\mpw}%
			\centering%
			\includegraphics[width=1.0\columnwidth]{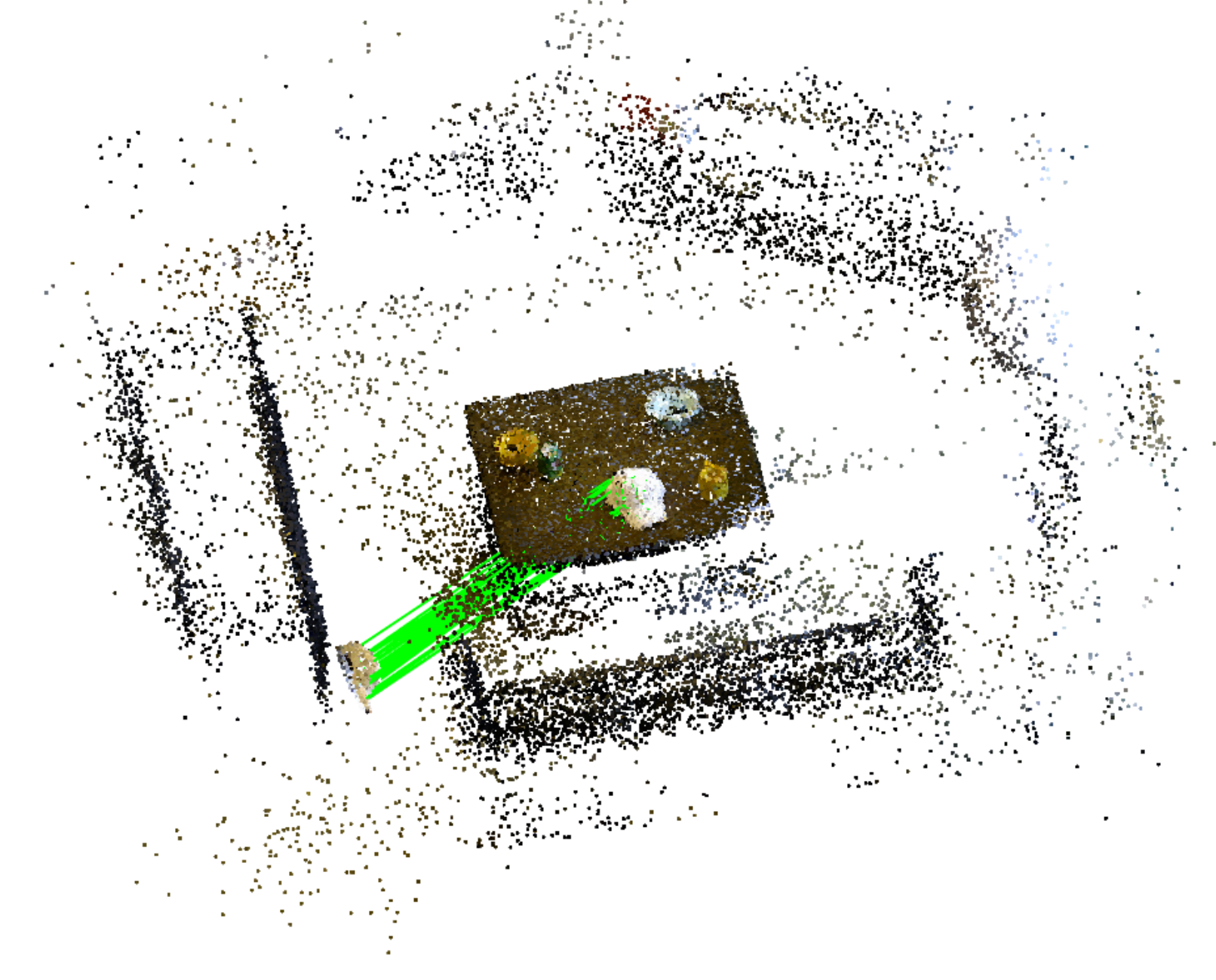} \\
			\end{minipage}
		& \myhspace
			\begin{minipage}{\mpw}%
			\centering%
			\includegraphics[width=1.0\columnwidth]{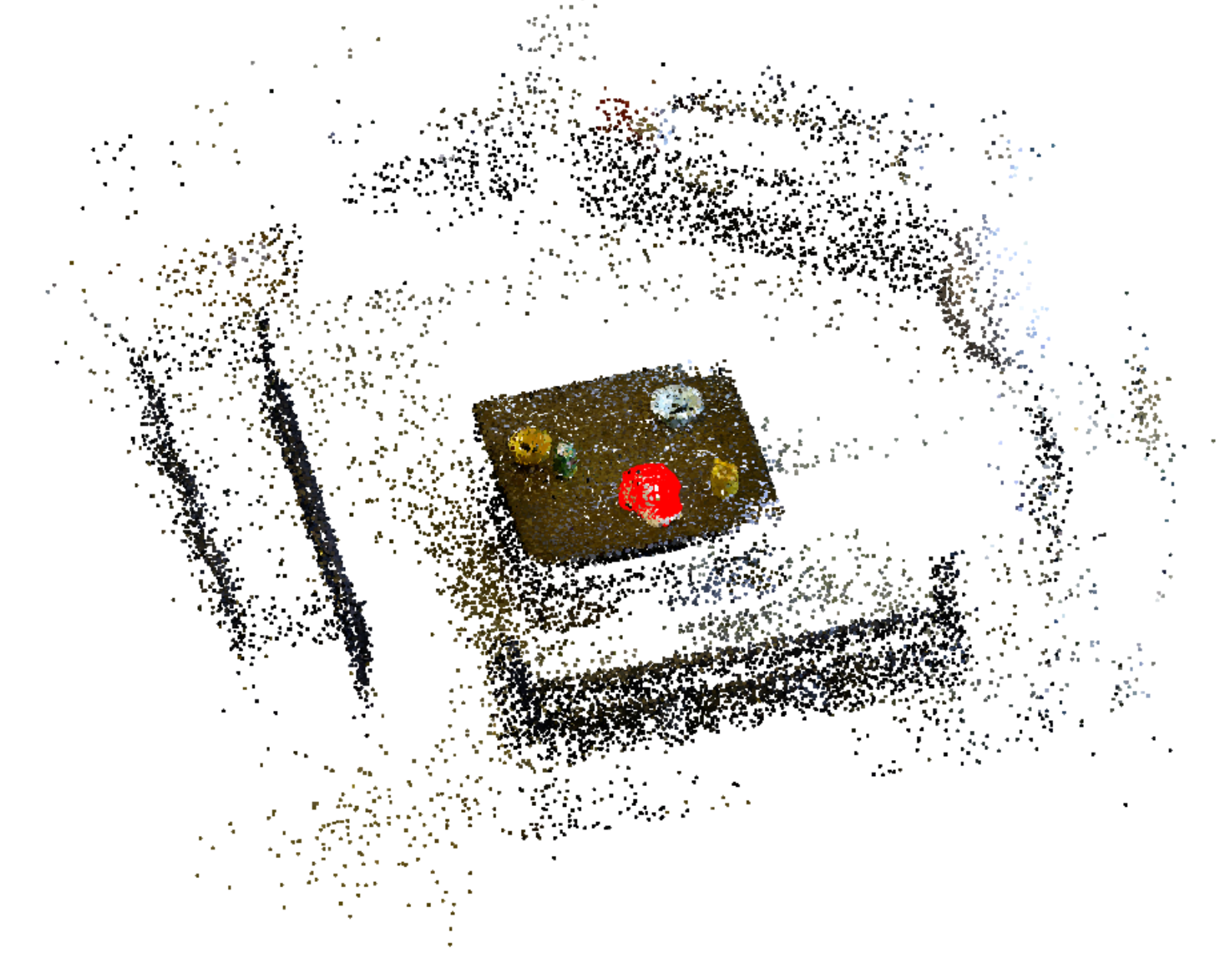} \\
			\end{minipage}\\
		\multicolumn{3}{c}{\emph{scene-1}, \# of FPFH correspondences: 207, Inlier ratio: 16.91\%, Rotation error: 0.066, Translation error: 0.090.} 
		
		\end{tabular}
	\end{minipage}
	\vspace{-3mm} 
	\caption{Object pose estimation on the large-scale RGB-D dataset~\cite{Lai11icra-largeRGBD} (cont.).
	 \label{fig:objectPoseEstimation}}
	\vspace{-8mm} 
	\end{center}
\end{figure*}


\bibliographystyle{plainnat}
\bibliography{../../references/refs.bib}

\begin{thebibliography}{62}
\providecommand{\natexlab}[1]{#1}
\providecommand{\url}[1]{\texttt{#1}}
\expandafter\ifx\csname urlstyle\endcsname\relax
  \providecommand{\doi}[1]{doi: #1}\else
  \providecommand{\doi}{doi: \begingroup \urlstyle{rm}\Url}\fi

\bibitem[Agarwal et~al.(2017)Agarwal, Shree, and
  Chakravorty]{agarwal2017icra-RFM-SLAM}
S.~Agarwal, V.~Shree, and S.~Chakravorty.
\newblock {RFM}-{SLAM}: Exploiting relative feature measurements to separate
  orientation and position estimation in slam.
\newblock In \emph{IEEE Intl. Conf. on Robotics and Automation (ICRA)}, pages
  6307--6314. IEEE, 2017.

\bibitem[Arun et~al.(1987)Arun, Huang, and Blostein]{Arun87pami}
K.~S. Arun, T.~S. Huang, and S.~D. Blostein.
\newblock Least-squares fitting of two 3-{D} point sets.
\newblock \emph{{IEEE} Trans. Pattern Anal. Machine Intell.}, 9\penalty0
  (5):\penalty0 698 --700, sept. 1987.

\bibitem[Audette et~al.(2000)Audette, Ferrie, and
  Peters]{Audette00mia-surveyMedical}
M.~A. Audette, F.~P. Ferrie, and T.~M. Peters.
\newblock An algorithmic overview of surface registration techniques for
  medical imaging.
\newblock \emph{Med. Image Anal.}, 4\penalty0 (3):\penalty0 201--217, 2000.

\bibitem[Bazin et~al.(2014)Bazin, Seo, Hartley, and
  Pollefeys]{Bazin14eccv-robustRelRot}
J.~C. Bazin, Y.~Seo, R.~I. Hartley, and M.~Pollefeys.
\newblock Globally optimal inlier set maximization with unknown rotation and
  focal length.
\newblock In \emph{European Conf. on Computer Vision (ECCV)}, pages 803--817,
  2014.

\bibitem[Besl and McKay(1992)]{Besl92pami}
P.~J. Besl and N.~D. McKay.
\newblock A method for registration of {3-D} shapes.
\newblock \emph{{IEEE} Trans. Pattern Anal. Machine Intell.}, 14\penalty0 (2),
  1992.

\bibitem[Black and Rangarajan(1996)]{Black96ijcv-unification}
M.~J. Black and A.~Rangarajan.
\newblock On the unification of line processes, outlier rejection, and robust
  statistics with applications in early vision.
\newblock \emph{Intl. J. of Computer Vision}, 19\penalty0 (1):\penalty0 57--91,
  1996.

\bibitem[Blais and Levine(1995)]{Blais95pami-registration}
G.~Blais and M.~D. Levine.
\newblock Registering multiview range data to create 3d computer objects.
\newblock \emph{{IEEE} Trans. Pattern Anal. Machine Intell.}, 17\penalty0
  (8):\penalty0 820--824, 1995.

\bibitem[Bouaziz et~al.(2013)Bouaziz, Tagliasacchi, and
  Pauly]{Bouaziz13acmsig-sparseICP}
S.~Bouaziz, A.~Tagliasacchi, and M.~Pauly.
\newblock Sparse iterative closest point.
\newblock In \emph{ACM Symp. Geom. Process.}, pages 113--123. Eurographics
  Association, 2013.

\bibitem[Breuel(2003)]{Breuel03cviu-BnBimplementation}
T.~M. Breuel.
\newblock Implementation techniques for geometric branch-and-bound matching
  methods.
\newblock \emph{Comput. Vis. Image Underst.}, 90\penalty0 (3):\penalty0
  258--294, 2003.

\bibitem[Briales and Gonzalez-Jimenez(2017)]{Briales17cvpr-registration}
J.~Briales and J.~Gonzalez-Jimenez.
\newblock {Convex Global 3D Registration with Lagrangian Duality}.
\newblock In \emph{IEEE Conf. on Computer Vision and Pattern Recognition
  (CVPR)}, 2017.

\bibitem[Bron and Kerbosch(1973)]{Bron73acm-allCliques}
C.~Bron and J.~Kerbosch.
\newblock Algorithm 457: finding all cliques of an undirected graph.
\newblock \emph{Communications of the ACM}, 16\penalty0 (9):\penalty0 575--577,
  1973.

\bibitem[Bustos and Chin(2018)]{Bustos18pami-GORE}
{\'A}.~P. Bustos and T.~J. Chin.
\newblock Guaranteed outlier removal for point cloud registration with
  correspondences.
\newblock \emph{{IEEE} Trans. Pattern Anal. Machine Intell.}, 40\penalty0
  (12):\penalty0 2868--2882, 2018.

\bibitem[Bustos et~al.(2014)Bustos, Chin, and
  Suter]{Parra14cvpr-fastRotationRegistration}
A.~Parra Bustos, T.~J. Chin, and D.~Suter.
\newblock Fast rotation search with stereographic projections for 3d
  registration.
\newblock In \emph{IEEE Conf. on Computer Vision and Pattern Recognition
  (CVPR)}, pages 3930--3937, 2014.

\bibitem[Campbell and Petersson(2015)]{Campbell15iccv-adaptiveRegistration}
D.~Campbell and L.~Petersson.
\newblock An adaptive data representation for robust point-set registration and
  merging.
\newblock In \emph{Intl. Conf. on Computer Vision (ICCV)}, pages 4292--4300,
  2015.

\bibitem[Campbell and Petersson(2016)]{Campbell16cvpr-gogma}
D.~Campbell and L.~Petersson.
\newblock Gogma: Globally-optimal gaussian mixture alignment.
\newblock In \emph{IEEE Conf. on Computer Vision and Pattern Recognition
  (CVPR)}, pages 5685--5694, 2016.

\bibitem[Chen et~al.(1999)Chen, Hung, and Cheng]{Chen99pami-ransac}
C.~S. Chen, Y.~P. Hung, and J.~B. Cheng.
\newblock {RANSAC}-based {DARCES}: A new approach to fast automatic
  registration of partially overlapping range images.
\newblock \emph{{IEEE} Trans. Pattern Anal. Machine Intell.}, 21\penalty0
  (11):\penalty0 1229--1234, 1999.

\bibitem[Chetverikov et~al.(2005)Chetverikov, Stepanov, and
  Krsek]{Chetverikov05ivc-trimmedICP}
D.~Chetverikov, D.~Stepanov, and P.~Krsek.
\newblock Robust euclidean alignment of 3d point sets: the trimmed iterative
  closest point algorithm.
\newblock \emph{Image and Vision Computing}, 23\penalty0 (3):\penalty0
  299--309, 2005.

\bibitem[Choi et~al.(2015)Choi, Zhou, and
  Koltun]{Choi15cvpr-robustReconstruction}
S.~Choi, Q.~Y. Zhou, and V.~Koltun.
\newblock Robust reconstruction of indoor scenes.
\newblock In \emph{IEEE Conf. on Computer Vision and Pattern Recognition
  (CVPR)}, pages 5556--5565, 2015.

\bibitem[Chung(1996)]{Chung96book}
F.R.K. Chung.
\newblock \emph{Spectral Graph Theory}.
\newblock American Mathematical Soc., CBMS Regional Conference Series in
  Mathematics, No. 92, 1996.

\bibitem[Curless and Levoy(1996)]{Curless96siggraph}
B.~Curless and M.~Levoy.
\newblock A volumetric method for building complex models from range images.
\newblock In \emph{SIGGRAPH}, pages 303--312, 1996.

\bibitem[Drost et~al.(2010)Drost, Ulrich, Navab, and Ilic]{Drost10cvpr}
B.~Drost, M.~Ulrich, N.~Navab, and S.~Ilic.
\newblock Model globally, match locally: Efficient and robust {3D} object
  recognition.
\newblock In \emph{IEEE Conf. on Computer Vision and Pattern Recognition
  (CVPR)}, pages 998--1005, 2010.

\bibitem[Eppstein et~al.(2010)Eppstein, L{\"o}ffler, and
  Strash]{Eppstein10isac-maxCliques}
D.~Eppstein, M.~L{\"o}ffler, and D.~Strash.
\newblock Listing all maximal cliques in sparse graphs in near-optimal time.
\newblock In \emph{International Symposium on Algorithms and Computation},
  pages 403--414. Springer, 2010.

\bibitem[Fischler and Bolles(1981)]{Fischler81}
M.~Fischler and R.~Bolles.
\newblock Random sample consensus: a paradigm for model fitting with
  application to image analysis and automated cartography.
\newblock \emph{Commun. ACM}, 24:\penalty0 381--395, 1981.

\bibitem[Granger and Pennec(2002)]{Granger02eccv}
S.~Granger and X.~Pennec.
\newblock Multi-scale {EM-ICP}: A fast and robust approach for surface
  registration.
\newblock In \emph{European Conf. on Computer Vision (ECCV)}, 2002.

\bibitem[Grant and Boyd()]{CVXwebsite}
M.~Grant and S.~Boyd.
\newblock {CVX}: Matlab software for disciplined convex programming.
\newblock URL \url{http://cvxr.com/cvx}.

\bibitem[Guo et~al.(2014)Guo, Bennamoun, Sohel, Lu, and
  Wan]{Guo14pami-3Dkeypoints}
Y.~Guo, M.~Bennamoun, F.~Sohel, M.~Lu, and J.~Wan.
\newblock {3D} object recognition in cluttered scenes with local surface
  features: a survey.
\newblock \emph{{IEEE} Trans. Pattern Anal. Machine Intell.}, 36\penalty0
  (11):\penalty0 2270--2287, 2014.

\bibitem[Hartley and Zisserman(2000)]{Hartley00}
R.~Hartley and A.~Zisserman.
\newblock \emph{Multiple View Geometry in Computer Vision}.
\newblock Cambridge University Press, 2000.

\bibitem[Hartley et~al.(2013)Hartley, Trumpf, Dai, and Li]{Hartley13ijcv}
R.~Hartley, J.~Trumpf, Y.~Dai, and H.~Li.
\newblock Rotation averaging.
\newblock \emph{IJCV}, 103\penalty0 (3):\penalty0 267--305, 2013.

\bibitem[Hartley and Kahl(2009)]{Hartley09ijcv-globalRotationRegistration}
R.~I. Hartley and F.~Kahl.
\newblock Global optimization through rotation space search.
\newblock \emph{Intl. J. of Computer Vision}, 82\penalty0 (1):\penalty0 64--79,
  2009.

\bibitem[Henry et~al.(2012)Henry, Krainin, Herbst, Ren, and
  Fox]{Henry12ijrr-rgbdMapping}
P.~Henry, M.~Krainin, E.~Herbst, X.~Ren, and D.~Fox.
\newblock Rgb-d mapping: Using kinect-style depth cameras for dense 3d modeling
  of indoor environments.
\newblock \emph{Intl. J. of Robotics Research}, 31\penalty0 (5):\penalty0
  647--663, 2012.

\bibitem[Horn(1987)]{Horn87josa}
B.~K.~P. Horn.
\newblock Closed-form solution of absolute orientation using unit quaternions.
\newblock \emph{J. Opt. Soc. Amer.}, 4\penalty0 (4):\penalty0 629--642, Apr
  1987.

\bibitem[Izatt et~al.(2017)Izatt, Dai, and
  Tedrake]{Izatt17isrr-MIPregistration}
G.~Izatt, H.~Dai, and R.~Tedrake.
\newblock Globally optimal object pose estimation in point clouds with
  mixed-integer programming.
\newblock In \emph{Proc. of the Intl. Symp. of Robotics Research (ISRR)}, 2017.

\bibitem[Jian and Vemuri(2005)]{Jian05iccv-registrationGMM}
B.~Jian and B.~C. Vemuri.
\newblock A robust algorithm for point set registration using mixture of
  gaussians.
\newblock In \emph{Intl. Conf. on Computer Vision (ICCV)}, volume~2, pages
  1246--1251. IEEE, 2005.

\bibitem[Jian and Vemuri(2011)]{Jian11pami-registrationGMM}
B.~Jian and B.~C. Vemuri.
\newblock Robust point set registration using gaussian mixture models.
\newblock \emph{{IEEE} Trans. Pattern Anal. Machine Intell.}, 33\penalty0
  (8):\penalty0 1633--1645, 2011.

\bibitem[Kaneko et~al.(2003)Kaneko, Kondo, and Miyamoto]{Kaneko03pr-robustICP}
S.~Kaneko, T.~Kondo, and A.~Miyamoto.
\newblock Robust matching of 3d contours using iterative closest point
  algorithm improved by m-estimation.
\newblock \emph{Pattern Recognition}, 36\penalty0 (9):\penalty0 2041--2047,
  2003.

\bibitem[Koppula et~al.(2011)Koppula, Anand, Joachims, and
  Saxena]{Koppula11nips}
H.S. Koppula, A.~Anand, T.~Joachims, and A.~Saxena.
\newblock Semantic labeling of 3d point clouds for indoor scenes.
\newblock In \emph{Advances in Neural Information Processing Systems (NIPS)},
  2011.

\bibitem[Lai et~al.(2011)Lai, Bo, Ren, and Fox]{Lai11icra-largeRGBD}
K.~Lai, L.~Bo, X.~Ren, and D.~Fox.
\newblock A large-scale hierarchical multi-view {RGB-D} object dataset.
\newblock In \emph{IEEE Intl. Conf. on Robotics and Automation (ICRA)}, pages
  1817--1824. IEEE, 2011.

\bibitem[Lajoie et~al.(2019)Lajoie, Hu, Beltrame, and
  Carlone]{Lajoie19ral-DCGM}
P.~Lajoie, S.~Hu, G.~Beltrame, and L.~Carlone.
\newblock Modeling perceptual aliasing in {SLAM} via discrete-continuous
  graphical models.
\newblock \emph{{IEEE} Robotics and Automation Letters}, 2019.

\bibitem[Li et~al.(2019)Li, Liu, Wang, Wang, Wang, and
  Song]{Li19arxiv-fastRegistration}
X.~Li, Y.~Liu, Y.~Wang, C.~Wang, M.~Wang, and Z.~Song.
\newblock Fast and globally optimal rigid registration of 3d point sets by
  transformation decomposition.
\newblock \emph{arXiv preprint arXiv:1812.11307}, 2019.

\bibitem[Liu et~al.(2018)Liu, Wang, Song, and Wang]{Liu18eccv-registration}
Y.~Liu, C.~Wang, Z.~Song, and M.~Wang.
\newblock Efficient global point cloud registration by matching rotation
  invariant features through translation search.
\newblock In \emph{European Conf. on Computer Vision (ECCV)}, September 2018.

\bibitem[Maier-Hein et~al.(2012)Maier-Hein, Franz, dos Santos, Schmidt,
  Fangerau, Meinzer, and Fitzpatrick]{Maier12pami-convergentICP}
L.~Maier-Hein, A.~M. Franz, T.~R. dos Santos, M.~Schmidt, M.~Fangerau, H.~P.
  Meinzer, and J.~M. Fitzpatrick.
\newblock Convergent iterative closest-point algorithm to accomodate
  anisotropic and inhomogenous localization error.
\newblock \emph{{IEEE} Trans. Pattern Anal. Machine Intell.}, 34\penalty0
  (8):\penalty0 1520--1532, 2012.

\bibitem[Makadia et~al.(2006)Makadia, Patterson, and
  Daniilidis]{Makadia06cvpr-registration}
A.~Makadia, A.~Patterson, and K.~Daniilidis.
\newblock Fully automatic registration of 3d point clouds.
\newblock In \emph{IEEE Conf. on Computer Vision and Pattern Recognition
  (CVPR)}, volume~1, pages 1297--1304, 2006.

\bibitem[Marion et~al.(2018)Marion, Florence, Manuelli, and
  Tedrake]{Marion18icra-labelFusion}
P.~Marion, P.~R. Florence, L.~Manuelli, and R.~Tedrake.
\newblock Label fusion: A pipeline for generating ground truth labels for real
  rgbd data of cluttered scenes.
\newblock In \emph{IEEE Intl. Conf. on Robotics and Automation (ICRA)}, pages
  1--8. IEEE, 2018.

\bibitem[Meer et~al.(1991)Meer, Mintz, Rosenfeld, and
  Kim]{Meer91ijcv-robustVision}
P.~Meer, D.~Mintz, A.~Rosenfeld, and D.~Y. Kim.
\newblock Robust regression methods for computer vision: A review.
\newblock \emph{Intl. J. of Computer Vision}, 6\penalty0 (1):\penalty0 59--70,
  Apr 1991.

\bibitem[Milanese(1989)]{Milanese89chapter-ubb}
M.~Milanese.
\newblock \emph{Estimation and Prediction in the Presence of Unknown but
  Bounded Uncertainty: A Survey}, pages 3--24.
\newblock Springer US, Boston, MA, 1989.

\bibitem[Myronenko and Song(2010)]{Myronenko10pami-coherentPointDrift}
A.~Myronenko and X.~Song.
\newblock Point set registration: Coherent point drift.
\newblock \emph{{IEEE} Trans. Pattern Anal. Machine Intell.}, 32\penalty0
  (12):\penalty0 2262--2275, 2010.

\bibitem[Olsson et~al.(2009)Olsson, Kahl, and Oskarsson]{olsson2009pami-branch}
C.~Olsson, F.~Kahl, and M.~Oskarsson.
\newblock Branch-and-bound methods for euclidean registration problems.
\newblock \emph{{IEEE} Trans. Pattern Anal. Machine Intell.}, 31\penalty0
  (5):\penalty0 783--794, 2009.

\bibitem[Pattabiraman et~al.(2015)Pattabiraman, Patwary, Gebremedhin, Liao, and
  Choudhary]{Pattabiraman15im-maxClique}
B.~Pattabiraman, M.~M.~A. Patwary, A.~H. Gebremedhin, W.~K. Liao, and
  A.~Choudhary.
\newblock Fast algorithms for the maximum clique problem on massive graphs with
  applications to overlapping community detection.
\newblock \emph{Internet Mathematics}, 11\penalty0 (4-5):\penalty0 421--448,
  2015.

\bibitem[Rusu et~al.(2008)Rusu, Blodow, Marton, and
  Beetz]{Rusu08iros-3Dkeypoints}
R.~B. Rusu, N.~Blodow, Z.~C. Marton, and M.~Beetz.
\newblock Aligning point cloud views using persistent feature histograms.
\newblock In \emph{IEEE/RSJ Intl. Conf. on Intelligent Robots and Systems
  (IROS)}, pages 3384--3391. IEEE, 2008.

\bibitem[Rusu et~al.(2009)Rusu, Blodow, and Beetz]{Rusu09icra-fast3Dkeypoints}
R.B. Rusu, N.~Blodow, and M.~Beetz.
\newblock Fast point feature histograms (fpfh) for 3d registration.
\newblock In \emph{IEEE Intl. Conf. on Robotics and Automation (ICRA)}, pages
  3212--3217. Citeseer, 2009.

\bibitem[Sandhu et~al.(2010)Sandhu, Dambreville, and
  Tannenbaum]{Sandhu10pami-PFregistration}
R.~Sandhu, S.~Dambreville, and A.~Tannenbaum.
\newblock Point set registration via particle filtering and stochastic
  dynamics.
\newblock \emph{{IEEE} Trans. Pattern Anal. Machine Intell.}, 32\penalty0
  (8):\penalty0 1459--1473, 2010.

\bibitem[Speciale et~al.(2017)Speciale, Paudel, Oswald, Kroeger, Gool, and
  Pollefeys]{Speciale17cvpr-consensusMaximization}
P.~Speciale, D.~P. Paudel, M.~R. Oswald, T.~Kroeger, L.~V. Gool, and
  M.~Pollefeys.
\newblock Consensus maximization with linear matrix inequality constraints.
\newblock In \emph{IEEE Conf. on Computer Vision and Pattern Recognition
  (CVPR)}, pages 5048--5056, July 2017.
\newblock \doi{10.1109/CVPR.2017.536}.

\bibitem[Tam et~al.(2013)Tam, Cheng, Lai, Langbein, Liu, Marshall, Martin, Sun,
  and Rosin]{Tam13tvcg-registrationSurvey}
G.~K.~L. Tam, Z.~Q. Cheng, Y.~K. Lai, F.~C. Langbein, Y.~Liu, D.~Marshall,
  R.~R. Martin, X.~F. Sun, and P.~L. Rosin.
\newblock Registration of 3d point clouds and meshes: a survey from rigid to
  nonrigid.
\newblock \emph{IEEE Trans. Vis. Comput. Graph.}, 19\penalty0 (7):\penalty0
  1199--1217, 2013.

\bibitem[Tavish and Barfoot(2015)]{MacTavish15crv-robustEstimation}
K.~Mac Tavish and T.~D. Barfoot.
\newblock At all costs: A comparison of robust cost functions for camera
  correspondence outliers.
\newblock In \emph{Computer and Robot Vision (CRV), 2015 12th Conference on},
  pages 62--69. IEEE, 2015.

\bibitem[Tombari et~al.(2013)Tombari, Salti, and
  Stefano]{Tombari13ijcv-3DkeypointEvaluation}
F.~Tombari, S.~Salti, and L.~Di Stefano.
\newblock Performance evaluation of 3d keypoint detectors.
\newblock \emph{Intl. J. of Computer Vision}, 102\penalty0 (1-3):\penalty0
  198--220, 2013.

\bibitem[Wong et~al.(2017)Wong, Kee, Le, Wagner, Mariottini, Schneider,
  Hamilton, Chipalkatty, Hebert, Johnson, et~al.]{Wong17iros-segicp}
J.~M. Wong, V.~Kee, T.~Le, S.~Wagner, G.~L. Mariottini, A.~Schneider,
  L.~Hamilton, R.~Chipalkatty, M.~Hebert, D.~M.~S. Johnson, et~al.
\newblock Segicp: Integrated deep semantic segmentation and pose estimation.
\newblock In \emph{IEEE/RSJ Intl. Conf. on Intelligent Robots and Systems
  (IROS)}, pages 5784--5789. IEEE, 2017.

\bibitem[Wu and Hao(2015)]{wu2015ejor-reviewMCPAlgs}
Q.~Wu and J.K. Hao.
\newblock A review on algorithms for maximum clique problems.
\newblock \emph{European Journal of Operational Research}, 242\penalty0
  (3):\penalty0 693--709, 2015.

\bibitem[Yang et~al.(2016)Yang, Li, Campbell, and Jia]{Yang16pami-goicp}
J.~Yang, H.~Li, D.~Campbell, and Y.~Jia.
\newblock {Go-ICP}: A globally optimal solution to {3D ICP} point-set
  registration.
\newblock \emph{{IEEE} Trans. Pattern Anal. Machine Intell.}, 38\penalty0
  (11):\penalty0 2241--2254, November 2016.
\newblock ISSN 0162-8828.

\bibitem[Zeng et~al.(2017)Zeng, Yu, Song, Suo, Walker, Rodriguez, and
  Xiao]{Zeng17icra-amazonChallenge}
A.~Zeng, K.~T. Yu, S.~Song, D.~Suo, E.~Walker, A.~Rodriguez, and J.~Xiao.
\newblock Multi-view self-supervised deep learning for 6d pose estimation in
  the amazon picking challenge.
\newblock In \emph{IEEE Intl. Conf. on Robotics and Automation (ICRA)}, pages
  1386--1383. IEEE, 2017.

\bibitem[Zhang et~al.(2016)Zhang, Du, Liu, and Xue]{Zhang16sp-robustICP}
C.~Zhang, S.~Du, J.~Liu, and J.~Xue.
\newblock Robust 3d point set registration using iterative closest point
  algorithm with bounded rotation angle.
\newblock \emph{Signal Processing}, 120:\penalty0 777--788, 2016.

\bibitem[Zhang and Singh(2015)]{Zhang15icra-vloam}
J.~Zhang and S.~Singh.
\newblock Visual-lidar odometry and mapping: Low-drift, robust, and fast.
\newblock In \emph{IEEE Intl. Conf. on Robotics and Automation (ICRA)}, pages
  2174--2181. IEEE, 2015.

\bibitem[Zhou et~al.(2016)Zhou, Park, and
  Koltun]{Zhou16eccv-fastGlobalRegistration}
Q.~Y. Zhou, J.~Park, and V.~Koltun.
\newblock Fast global registration.
\newblock In \emph{European Conf. on Computer Vision (ECCV)}, pages 766--782.
  Springer, 2016.

\end{thebibliography}

\end{document}